\documentclass[11pt, a4paper, logo, copyright, nonumbering]{map}

\definecolor{hidden-draw}{RGB}{20,68,106}
\definecolor{hidden-pink}{RGB}{255,245,247}
\usepackage[edges]{forest}

\usepackage{natbib}
\usepackage{CJKutf8}

\usepackage{xargs}  

\usepackage{todonotes}  

\usepackage{multirow}

\usepackage{cleveref}

\usepackage{amsmath}
\usepackage{dsfont}

\usepackage{subcaption}

\usepackage{svg}
\usepackage[breakable]{tcolorbox}
\usepackage{fontawesome}
\usepackage{xcolor}

\usepackage[utf8]{inputenc}
\usepackage{booktabs}
\usepackage{graphicx}
\usepackage{array}
\usepackage{tabularx}
\usepackage{xcolor,colortbl}  
\definecolor{boxcolor}{HTML}{d92523} 
\definecolor{bulbcolor}{HTML}{e3b87f} 
\usepackage{url}

\usepackage{hyperref}       
\usepackage{url}            
\usepackage{booktabs}       
\usepackage{amsfonts}       
\usepackage{nicefrac}       
\usepackage{microtype}      
\usepackage{xcolor}         
\usepackage{graphicx}
\usepackage{longtable}
\usepackage{array}
\usepackage{pbox}
\usepackage{amsmath}
\usepackage{amssymb}
\usepackage{tabularx}
\usepackage{multirow}
\usepackage{wrapfig}
\usepackage{pifont}
\newcommand{\Checkmark}{\ding{51}} 
\newcommand{\XSolidBrush}{\ding{55}} 
\usepackage{algorithm}
\usepackage{algorithmic}
\usepackage{lipsum}
\usepackage{subcaption}
\usepackage{tcolorbox}
\usepackage{enumitem}
\usepackage{tikz}
\usepackage{xstring}

\usepackage{color-edits}
\addauthor[Zhejian]{zzj}{blue}

\title{A Comprehensive Survey on Long Context Language Modeling}

\hypersetup{
    colorlinks=true,            
    linkcolor=blue,             
    filecolor=magenta,          
    urlcolor=cyan,              
    citecolor=purple,             
    pdftitle={Overleaf Example},
    pdfpagemode=FullScreen,
    }

\renewcommand{\today}{}
\newcommandx{\info}[2][1=]{\todo[linecolor=red,backgroundcolor=red!25,bordercolor=red,#1]{#2}}

\title{\vspace{-0.2in}
\centering A Comprehensive Survey on Long Context \\ Language Modeling
    \vspace{-0.2in}
}

\author{
\small
Jiaheng Liu$^{\ast,\dagger}$, 
Dawei Zhu$^{\ast, \dagger}$, 
Zhiqi Bai$^{\star}$,
Yancheng He$^{\star}$, 
Huanxuan Liao$^{\star}$,
Haoran Que$^{\star}$, 
Zekun Wang$^{\star}$,
Chenchen Zhang$^{\star}$,
Ge Zhang$^{\star}$,
Jiebin Zhang$^{\star}$,
Yuanxing Zhang$^{\star}$,
Zhuo Chen,
Hangyu Guo,
Shilong Li,
Ziqiang Liu,
Yong Shan,
Yifan Song,
Jiayi Tian,
Wenhao Wu,
Zhejian Zhou,
Ruijie Zhu,
Junlan Feng,
Yang Gao,
Shizhu He,
Zhoujun Li,
Tianyu Liu,
Fanyu Meng,
Wenbo Su,
Yingshui Tan,
Zili Wang,
Jian Yang,
Wei Ye,
Bo Zheng,
Wangchunshu Zhou,
Wenhao Huang$^\dagger$,
Sujian Li$^\dagger$,
Zhaoxiang Zhang$^\dagger$
\\
        \vspace{0.1in}
NJU, PKU, CASIA, Alibaba, ByteDance, Tencent, Kuaishou, M-A-P
    \vspace{-0.35in}

}

\begin{abstract}

\vspace{-0.2in}

Efficient processing of long contexts has been a persistent pursuit in Natural Language Processing. With the growing number of long documents, dialogues, and other textual data, it is important to develop Long Context Language Models (LCLMs) that can process and analyze extensive inputs in an effective and efficient way.
In this paper, we present a comprehensive survey on recent advances in long-context modeling for large language models. Our survey is structured around three key aspects: how to obtain effective and efficient LCLMs, how to train and deploy LCLMs efficiently, and how to evaluate and analyze LCLMs comprehensively. For the first aspect, we discuss data strategies, architectural designs, and workflow approaches oriented with long context processing. For the second aspect, we provide a detailed examination of the infrastructure required for LCLM training and inference. For the third aspect, we present evaluation paradigms for long-context comprehension and long-form generation, as well as behavioral analysis and mechanism interpretability of LCLMs. Beyond these three key aspects, we thoroughly explore the diverse application scenarios where existing LCLMs have been deployed and outline promising future development directions. This survey provides an up-to-date review of the literature on long-context LLMs, which we wish to serve as a valuable resource for both researchers and engineers. An associated GitHub repository collecting the latest papers and repos is available at: \href{https://github.com/LCLM-Horizon/A-Comprehensive-Survey-For-Long-Context-Language-Modeling}{\color[RGB]{175,36,67}{LCLM-Horizon}}.

\end{abstract}

\begin{document}

\maketitle

\let\oldthefootnote\thefootnote

\let\thefootnote\relax\footnotetext{$^\ast$~Project Leaders (Equal Contribution). 
~~$^{\star}$ Core Contributors (Equal Contribution). ~~$^\dagger$~Corresponding Authors. }
\let\thefootnote\oldthefootnote

\begin{figure}[H]
    \vspace{-5mm}
    \centering
    \includegraphics[width=0.99\textwidth]{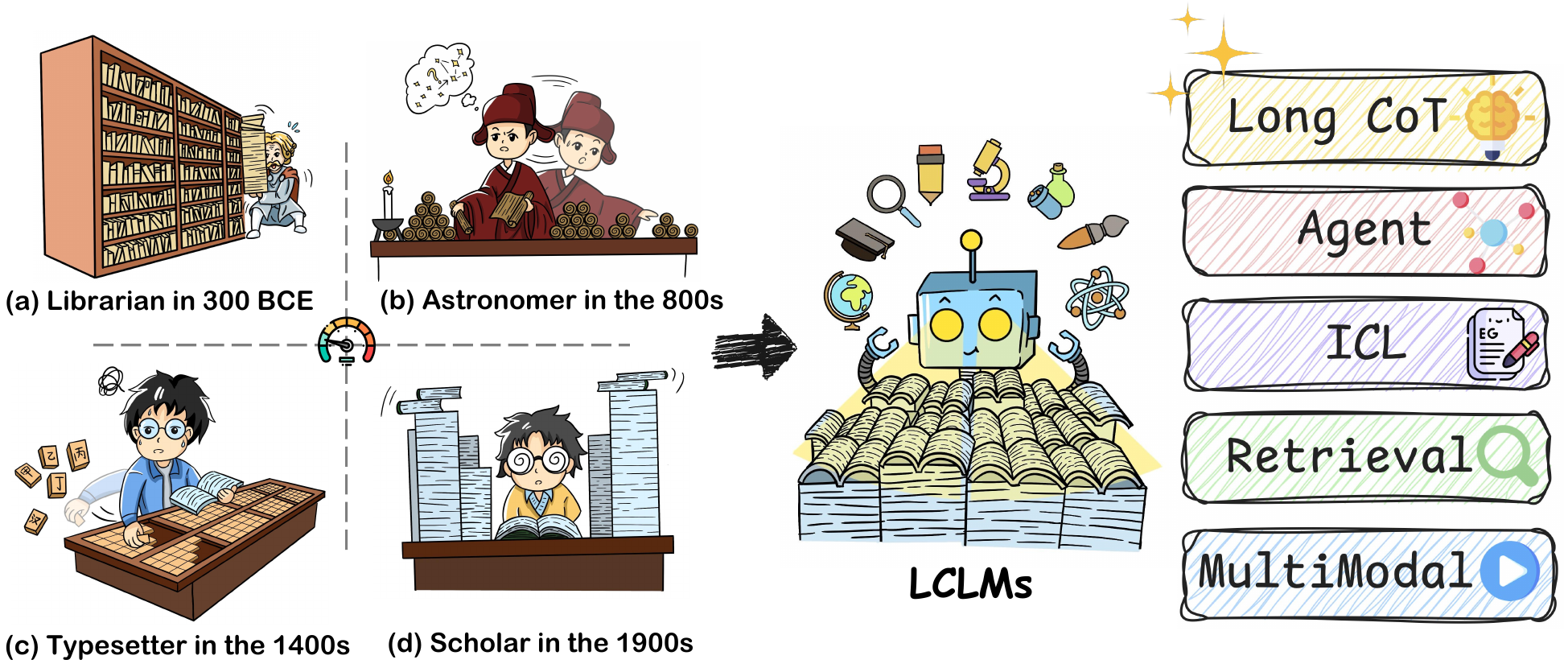}
    \vspace{-2mm}
    \caption{Evolution of Long Context Processing over time. The advent of LCLMs efficiently processes millions of data in minutes and unlocks intriguing applications.}
    \label{fig:intro}
\end{figure}

\newpage

\tableofcontents

\newpage

\section{Introduction}

Efficient data processing has long been humanity's aspiration, as our biological constraints limit us to localized, serial reading, making the manual processing of long-context of data painfully inefficient.
As shown in Figure~\ref{fig:intro},
considering these historical moments: In 300 BCE, librarians at the Library of Alexandria had to manually copy, proofread manuscripts, and compile catalogs to manage hundreds of thousands of ancient texts~\citep{alex_library}. In the 800s, astronomical officials of the Tang Dynasty needed to manually process vast astronomical measurement data to calculate seasonal divisions~\citep{tang_astronomer}. In the 1400s, typesetters had to manually arrange enormous printing plates before newspapers could be printed~\citep{print_press}. Even into the 1900s, scholars still needed to meticulously examine dozens or even hundreds of relevant documents before achieving a comprehensive understanding of a subject. 

A revolutionary leap finally arrived with the emergence of language models~\citep{wang2023rolellm, Zhang2025CodeCriticBenchAH,mceval,mdeval,tablebench,supergpqa,sailor2report,zhang2024mapneo,Huang2024OpenCoderTO,liu-etal-2024-e2,codearena,execrepobench}, which offer the promise of automatically processing text data in minutes. These language models operate within a fixed context window, probabilistically modeling input sequences and enabling next-token prediction. Early language models were limited to processing only a few or dozens of tokens~\citep{shannon1948mathematical,brants2007large,chen1999empirical,mikolov2010recurrent}. As context lengths expanded to several hundred or thousands of tokens, represented by BERT~\citep{devlin-etal-2019-bert} and GPT-3~\citep{10.5555/3495724.3495883} respectively, automatic processing of paragraphs, documents, and multi-turn conversations became achievable for the first time. Building on these advances, recent years have witnessed the advent of Long Context Language Models (LCLMs) with their context length growing exponentially from 4K to 128K~\citep{grattafiori2024llama}, 1M~\citep{yang2025qwen2} and even 10M~\citep{team2024gemini} tokens, enabling single-pass ingestion of Tolstoyan-scale narratives (560K words) and effectively condensing 60 hours of human reading into minutes of computational processing. More importantly, these extensive context lengths provide sufficient space for test-time scaling~\citep{openai_o1_2024,guo2025deepseek}, where models can explore, reflect, backtrack, and summarize within a single context, which fundamentally transforms our interaction with generative AI and unlocks a series of intriguing abilities, including: o1-like long reasoning~\citep{guo2025deepseek,openai_o1_2024,muennighoff2025s1}, complex agent workflows~\citep{oai2025deepresearch}, superior in-context learning~\citep{minimax2025minimax01scalingfoundationmodels,team2024gemini}, efficient information retrieval \& comprehension~\citep{wang2024large,lee2024can}, and advanced multimodal intelligence~\citep{Qwen2.5-VL,weng2024longvlm}.

In this paper, 
we present a comprehensive survey on long context language modeling. As shown in Figure~\ref{fig:taxo_of_lclms}, our survey is oriented with the following three pivotal dimensions: \textbf{RQ1}: How to obtain effective and efficient LCLMs? \textbf{RQ2}: How to train and deploy LCLMs efficiently? \textbf{RQ3}: How to evaluate and analyze LCLMs comprehensively? Beyond these three key
aspects, we also thoroughly explore the diverse application scenarios of existing  LCLMs.
\tikzstyle{my-box}=[
    rectangle,
    draw=hidden-draw,
    rounded corners,
    text opacity=1,
    minimum height=1.5em,
    minimum width=5em,
    inner sep=2pt,
    align=center,
    fill opacity=.5,
    line width=0.8pt,
]
\tikzstyle{leaf}=[my-box, minimum height=1.5em,
    fill=hidden-pink!80, text=black, align=left,font=\normalsize,
    inner xsep=2pt,
    inner ysep=4pt,
    line width=0.8pt,
]

    
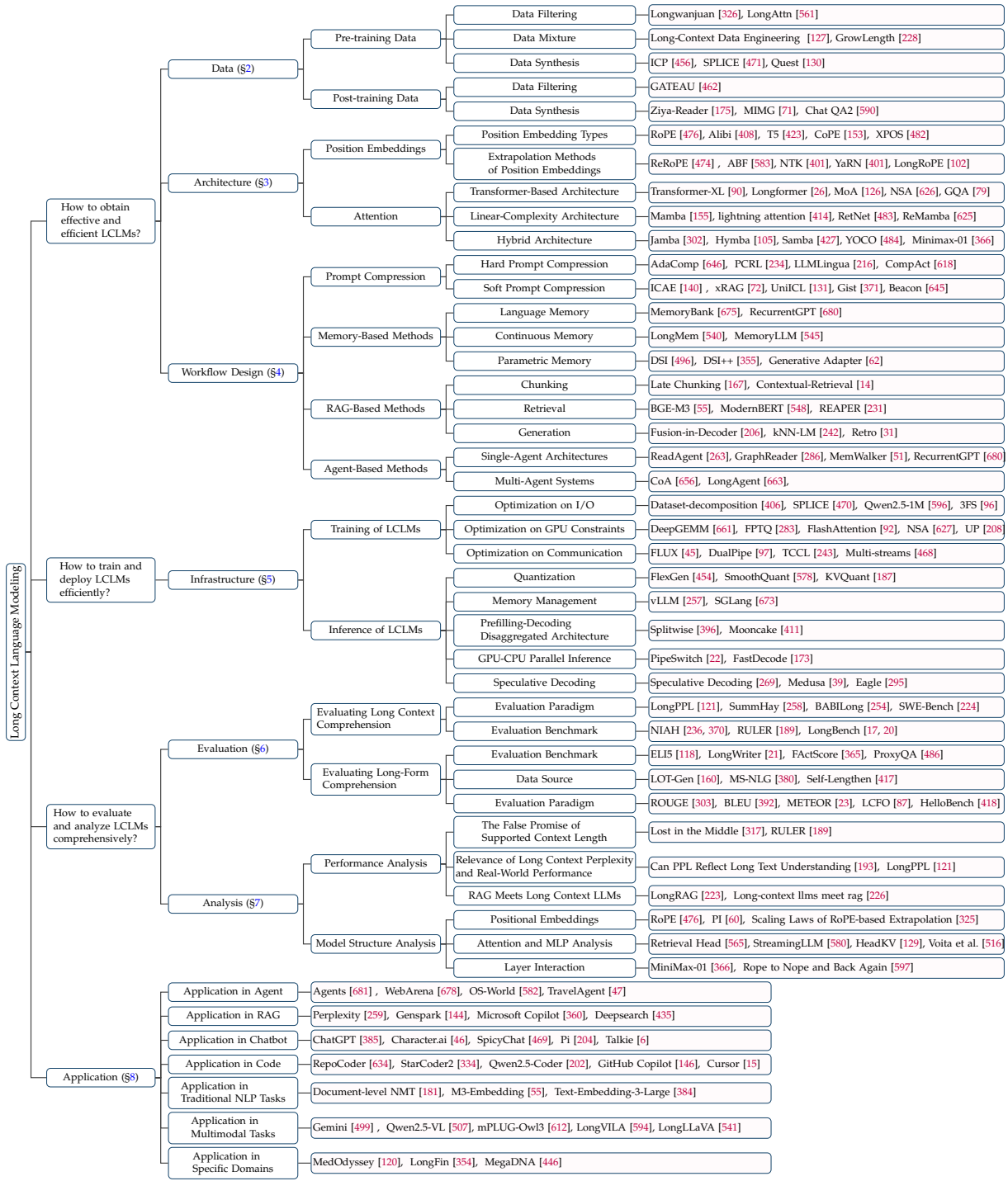
\begin{figure}[t!]
    \centering
    \resizebox{\textwidth}{!}{
        \begin{forest}
        forked edges,
            for tree={
                grow=east,
                reversed=true,
                anchor=base west,
                parent anchor=east,
                child anchor=west,
                base=center,
                font=\large,
                rectangle,
                draw=hidden-draw,
                rounded corners,
                align=left,
                text centered,
                minimum width=4em,
                edge+={darkgray, line width=1pt},
                s sep=3pt,
                inner xsep=2pt,
                inner ysep=3pt,
                line width=0.8pt,
                ver/.style={rotate=90, child anchor=north, parent anchor=south, anchor=center}
            },
            where level=1{text width=10em,font=\normalsize,}{},
            where level=2{text width=12em,font=\normalsize,}{},
            where level=3{text width=12em,font=\normalsize,}{},
            where level=4{text width=17em,font=\normalsize,}{},
            where level=5{text width=6em,font=\normalsize,}{},
            [
                Long Context Language Modeling, ver
                [
                    How to obtain\\
                    effective and \\efficient LCLMs?
                    [
                        Data
                        (\S\ref{sec:data})
                        [
                            Pre-training Data
                            [
                                Data Filtering
                                [
                                    Longwanjuan~\citep{liu-etal-2024-longwanjuan}{, }LongAttn~\citep{wu2025longattn}
                                    , leaf, text width=33.6em
                                ]
                            ]
                            [
                                Data Mixture
                                [
                                   Long-Context Data Engineering
~\cite{Fu2024DataEF}{, }GrowLength~\citep{growlength}
                                    , leaf, text width=33.6em
                                ]
                            ]
                            [
                                Data Synthesis
                                [
                                    ICP~\citep{Shi2023InContextPL}{, } SPLICE~\citep{spacking}{, }Quest~\citep{quest}
                                    , leaf, text width=33.6em
                                ]
                            ]
                        ]
                        [
                            Post-training Data
                            [
                                Data Filtering
                                [
                                    GATEAU~\citep{geteau}
                                    , leaf, text width=33.6em
                                ]
                            ]
                            [
                                Data Synthesis
                                [
                                    Ziya-Reader~\citep{pamqa}{, }
                                    MIMG~\citep{Chen2024WhatAT}{, }
                                    Chat QA2~\citep{Xu2024ChatQA2B}
                                    , leaf, text width=33.6em
                                ]
                            ]
                        ]
                    ]
                    [
                        Architecture (\S\ref{sec:model})
                        [
                            Position Embeddings
                            [
                                Position Embedding Types
                                [
                                    RoPE~\citep{su2024roformer}{, }Alibi~\citep{press2021train}{, }
                                    T5~\citep{raffel2020exploring}{, }
                                CoPE~\citep{golovneva2024contextual}{, }
                                XPOS~\citep{sun2022length}
                                    , leaf, text width=33.6em
                                ]
                            ]
                            [
                                 Extrapolation Methods \\
                                 of Position Embeddings
                                [
                                   ReRoPE~\citep{kexuefm-9708} {, }
                                    ABF~\citep{xiong2023effective}{, }NTK~\citep{peng2023yarn}{, }YaRN~\citep{peng2023yarn}{, }LongRoPE~\cite{ding2024longrope}
                                    , leaf, text width=33.6em
                                ]
                            ]
                        ]
                        [
                            Attention
                            [
                                Transformer-Based Architecture
                                [Transformer-XL~\cite{TransformerXL}{, }Longformer~\citep{beltagy2020longformerlongdocumenttransformer}{, }MoA~\citep{fu2024moamixturesparseattention}{, }NSA~\citep{yuan2025nativesparseattentionhardwarealigned}{, }GQA~\citep{chinnakonduru2024weightedgroupedqueryattention}
                                    , leaf, text width=33.6em
                                ]
                            ]
                            [
                                Linear-Complexity Architecture
                                [
                                    Mamba~\cite{gu2023mamba}{, }lightning attention~\cite{qin2024lightningattention2freelunch}{, }RetNet~\cite{sun2023retentivenetworksuccessortransformer}{, }ReMamba~\cite{yuan2025remambaequipmambaeffective}
                                    , leaf, text width=33.6em
                                ]
                            ]
                            [
                                Hybrid Architecture
                                [
                                    Jamba~\cite{lieber2024jamba}{, }  Hymba~\cite{dong2024hymba}{, }Samba~\cite{ren2024samba}{, }YOCO~\cite{sun2025you}{, }
                                    Minimax-01~\citep{minimax2025minimax01scalingfoundationmodels}
                                    , leaf, text width=33.6em
                                ]
                            ]
                        ]
                    ]
                    [
                        Workflow Design (\S\ref{sec:workflow})
                        [
                            Prompt Compression
                            [
                                Hard Prompt Compression
                                [
                                    AdaComp~\citep{zhang2024adacompextractivecontextcompression}{, }
                                    PCRL~\citep{Jung_2024}{, }LLMLingua~\citep{jiang-etal-2023-llmlingua}{, }
                                    CompAct~\citep{yoon-etal-2024-compact} 
                                    , leaf, text width=33.6em
                                ]
                            ]
                            [
                                Soft Prompt Compression
                                [
                                    ICAE~\citep{ge2024incontext} {, } xRAG~\citep{cheng2024xrag}{, }UniICL~\citep{gao2024unifyingdemonstrationselectioncompression}{, }Gist~\citep{NEURIPS2023_3d77c6dc}{, }Beacon~\citep{zhang2025long}
                                    , leaf, text width=33.6em
                                ]
                            ]
                        ]
                        [
                            Memory-Based Methods
                            [
                                Language Memory
                                [
                                    MemoryBank~\citep{DBLP:conf/aaai/ZhongGGYW24}{, }
                                    RecurrentGPT~\citep{recurrentgpt}
                                    , leaf, text width=33.6em
                                ]
                            ]
                            [
                                Continuous Memory
                                [
                                    LongMem~\citep{longmem}{, }
                                    MemoryLLM~\citep{memoryllm}
                                    , leaf, text width=33.6em
                                ]
                            ]
                            [
                                Parametric  Memory
                                [
                                    DSI~\citep{dsi}{, }
                                    DSI++~\citep{dsiplusplus}{, }
                                    Generative Adapter~\citep{chen2024generative}
                                    , leaf, text width=33.6em
                                ]
                            ]
                        ]
                        [
                            RAG-Based Methods
                            [
                                Chunking
                                [
                                    Late Chunking~\citep{günther2024late}{, }
                                    Contextual-Retrieval~\citep{contextual-retrieval}
                                    , leaf, text width=33.6em
                                ]
                            ]
                            [
                                Retrieval
                                [
                                    BGE-M3~\citep{chen2024bge}{, }
                                    ModernBERT~\citep{warner2024smarter0}{, }
                                    REAPER~\citep{joshi2024reaper}
                                    , leaf, text width=33.6em
                                ]
                            ]
                            [
                                Generation
                                [
                                    Fusion-in-Decoder~\citep{fusion-in-decoder}{, }
                                    kNN-LM~\citep{knn-lm}{, }
                                    Retro~\citep{retro}
                                    , leaf, text width=33.6em
                                ]
                            ]
                        ]
                        [
                            Agent-Based Methods
                            [
                                Single-Agent Architectures
                                [
                                    ReadAgent~\citep{readagent}{,}
                                    GraphReader~\citep{graphreader}{,}
                                    MemWalker~\citep{memwalker}{,}
                                    RecurrentGPT~\citep{recurrentgpt}
                                    , leaf, text width=33.6em
                                ]
                            ]
                            [
                                Multi-Agent Systems
                                [
                                    CoA~\citep{chain-of-agent}{, }
                                    LongAgent~\citep{longagent}{, }
                                    , leaf, text width=33.6em
                                ]
                            ]
                        ]
                    ]
                ]
                [
                    How to train and 
                    \\deploy LCLMs \\
                    efficiently?
                    [
                        Infrastructure
                        (\S\ref{sec:infra})
                        [
                            Training of LCLMs
                            [
                                Optimization on I/O
                                [
                                     Dataset-decomposition~\citep{pouransari2024dataset}{, }
                                     SPLICE~\citep{staniszewski2023structured}{, }
                                     Qwen2.5-1M~\citep{yang2025qwen2}{, }
                                     3FS~\citep{deepseek3fs}
                                    , leaf, text width=33.6em
                                ]
                            ]
                            [
                               Optimization on GPU Constraints
                                [
                                    DeepGEMM~\citep{deepgemm2025}{, }
                                     FPTQ~\citep{li2023fptq}{, }
                                     FlashAttention~\citep{dao2024flashattention}{, }
                                     NSA~\citep{yuan2025native}{, }
                                     UP~\citep{jacobs2023deepspeed}
                                    , leaf, text width=33.6em
                                ]
                            ]
                            [
                                Optimization on Communication
                                [
                                    FLUX~\citep{chang2024flux}{, }
                                     DualPipe~\citep{deepseekai2024deepseekv3technicalreport}{, }
                                     TCCL~\citep{kim2024tccl}{, }
                                     Multi-streams~\citep{sourouri2014effective}
                                    , leaf, text width=33.6em
                                ]
                            ]
                        ]
                        [
                            Inference of LCLMs
                            [
                                Quantization
                                [   FlexGen~\citep{sheng2023flexgen}{, }
                                    SmoothQuant~\citep{xiao2023smoothquant}{, }
                                    KVQuant~\citep{hooper2025kvquant}
                                    , leaf, text width=33.6em
                                ]
                            ]
                            [
                                Memory Management
                                [   vLLM~\citep{kwon2023efficient}{, } 
                                    SGLang~\citep{zheng2023efficiently}
                                    , leaf, text width=33.6em
                                ]
                            ]
                            [
                                Prefilling-Decoding \\
                                Disaggregated Architecture
                                [
                                    Splitwise~\citep{patel2024splitwise}{, }
                                    Mooncake~\citep{qin2024mooncake}
                                    , leaf, text width=33.6em
                                ]
                            ]
                            [
                                GPU-CPU Parallel Inference
                                [
                                    PipeSwitch~\citep{bai2020pipeswitch}{, } 
                                    FastDecode~\citep{he2024fastdecode}
                                    , leaf, text width=33.6em
                                ]
                            ]
                            [
                                Speculative Decoding
                                [
                                    Speculative Decoding~\citep{leviathan2023fast}{, }
                                    Medusa~\citep{cai2024medusa}{, }
                                    Eagle~\citep{li2024eagle}
                                    , leaf, text width=33.6em
                                ]
                            ]
                        ]
                    ]
                ]
                [
                    How to  evaluate \\
                    and analyze LCLMs \\
                    comprehensively?
                     [
                        Evaluation
                        (\S\ref{sec:evaluation})
                        [
                            Evaluating Long Context\\
                            Comprehension
                            [
                                Evaluation Paradigm
                                [
                                   LongPPL~\cite{fang2024wrong}{, } 
                                   SummHay~\citep{laban2024SummHay}{, }  
                                   BABILong~\citep{kuratov2024babilong}{, } 
                                   SWE-Bench~\citep{jimenez2024swebench}
                                    , leaf, text width=33.6em
                                ]
                            ]
                            [
                                Evaluation Benchmark
                                [
                                   NIAH~\citep{mohtashami2023landmark,needleinhaystack}{, }
                                   RULER~\citep{hsieh2024ruler}{, }
                                   LongBench~\citep{bai2023longbench,bai2024longbench2}
                                    , leaf, text width=33.6em
                                ]   
                            ]
                        ]
                        [
                            Evaluating Long-Form\\
                            Comprehension
                            [
                                Evaluation Benchmark
                                [
                                   ELI5~\citep{fan2019eli5}{, }
                                   LongWriter~\citep{bai2024longwriter}{, }
                                   FActScore~\citep{min2023factscore}{, }
                                   ProxyQA~\citep{tan2024proxyqa}
                                    , leaf, text width=33.6em
                                ]
                            ]
                            [
                                Data Source
                                [
                                   LOT-Gen~\citep{guan2022lot}{, }
                                   MS-NLG~\citep{nguyen2016ms}{, }
                                   Self-Lengthen~\citep{quan2024language}
                                    , leaf, text width=33.6em
                                ]
                            ]
                            [
                                Evaluation Paradigm
                                [
                                   ROUGE~\citep{lin2004rouge}{, }
                                   BLEU~\citep{papineni2002bleu}{, }
                                   METEOR~\citep{banerjee2005meteor}{, }
                                   LCFO~\citep{costa2024lcfo}{, }
                                   HelloBench~\citep{que2024hellobench}
                                    , leaf, text width=33.6em
                                ]
                            ]
                        ]
                    ]
                    [
                        Analysis
                        (\S\ref{sec:analysis})
                        [
                            Performance Analysis
                            [
                                The False Promise of \\
                                Supported Context Length
                                [
                                    Lost in the Middle~\cite{liu2024lost}{, }RULER~\cite{hsieh2024ruler}
                                    , leaf, text width=33.6em
                                ]
                            ]
                            [
                                Relevance of Long Context Perplexity \\
                                and Real-World Performance
                                [
                                    Can PPL Reflect Long Text Understanding~\cite{hu2024can}{, }
                                    LongPPL~\cite{fang2024wrong}
                                    , leaf, text width=33.6em
                                ]
                            ]
                            [
                                RAG Meets Long Context LLMs
                                [
                                    LongRAG~\citep{jiang2024longrag}{, }
                                    Long-context llms meet rag~\citep{jin2024long}
                                    , leaf, text width=33.6em
                                ]
                            ]
                        ]
                        [
                            Model Structure Analysis
                            [
                                Positional Embeddings
                                [
                                    RoPE~\citep{su2024roformer}{, }
                                    PI~\citep{chen2023extending}{, }
                                    Scaling Laws of RoPE-based Extrapolation~\citep{liu2023scaling}
                                    , leaf, text width=33.6em
                                ]
                            ]
                            [
                                Attention and MLP Analysis
                                [
                                    Retrieval Head~\citep{wu2025retrieval_head}{,}
                                    StreamingLLM~\citep{attn_sink}{,}
                                    HeadKV~\citep{fu2025not}{,}
                                    \citet{voita-etal-2024-neurons_func}
                                    , leaf, text width=33.6em
                                ]
                            ]
                            [
                                Layer Interaction
                                [
                                    MiniMax-01~\citep{minimax2025minimax01scalingfoundationmodels}{, }
                                    Rope to Nope and Back Again~\citep{yang_hybrid_attn_rope_nope}
                                    , leaf, text width=33.6em
                                ]
                            ]
                        ]
                    ]
                ]
                [
                    Application (\S\ref{sec:application})
                    [
                        Application in Agent
                        [
                            Agents~\citep{zhou2023agents} {, } WebArena~\citep{zhou2023webarena}{, } OS-World~\citep{xie2024osworld}{, }TravelAgent~\citep{chen2024travelagent}
                                , leaf, text width=43.6em
                        ]
                    ]
                    [
                        Application in RAG
                        [
                            Perplexity~\citep{perplexity_pages}{, } Genspark~\citep{genspark}{, } Microsoft Copilot~\citep{bing_copilot}{, } Deepsearch~\citep{deepsearch}
                                , leaf, text width=43.6em
                        ]
                    ]
                    [
                        Application in Chatbot
                        [
                            ChatGPT~\cite{openai2024memory}{, }
                            Character.ai~\cite{character_ai}{, }
                            SpicyChat~\cite{spicychat}{, }
                            Pi~\cite{inflection2023impi}{, }
                            Talkie~\cite{talkie}
                                , leaf, text width=43.6em
                        ]
                    ]
                    [
                        Application in Code
                        [
                            RepoCoder~\cite{zhang2023repocoder}{, }
                            StarCoder2~\cite{lozhkov2024starcoder2stackv2}{, }
                            Qwen2.5-Coder~\cite{hui2024qwen2}{, }
                            GitHub Copilot~\cite{github_copilot}{, }
                            Cursor~\cite{cursor_ai_2025}
                                , leaf, text width=43.6em
                        ]
                    ]
                    [
                        Application in \\
                        Traditional NLP Tasks
                        [
                            Document-level NMT~\cite{herold2023improving}{, }
                            M3-Embedding~\cite{chen2024bge}{, }
                            Text-Embedding-3-Large~\cite{openai2024embedding}
                                , leaf, text width=43.6em
                        ]
                    ]
                    [
                        Application in \\
                        Multimodal Tasks
                        [
                                    Gemini~\citep{team2024gemini} {, } Qwen2.5-VL~\citep{Qwen2.5-VL}{, }mPLUG-Owl3~\citep{ye-2024-arxiv-mPLUG-Owl3}{, }LongVILA~\citep{xue-2024-arxiv-LongVILA}{, }LongLLaVA~\citep{wang-2024-arxiv-LongLLaVA}
                                , leaf, text width=43.6em
                        ]
                    ]
                    [
                        Application in \\
                        Specific Domains
                        [
                            MedOdyssey~\cite{fan2024medodyssey}{, }
                            LongFin~\cite{masry2024longfin}{, }
                            MegaDNA~\cite{shao2024long}
                                , leaf, text width=43.6em
                        ]
                    ]
                ]
            ]
        \end{forest}
    }
    \caption{Taxonomy of Long Context Language Modeling.}
    \label{fig:taxo_of_lclms}
\end{figure}

First, to obtain effective and efficient LCLMs (\textbf{RQ1}), we review data strategies~(\S~\ref{sec:data}), architecture designs~(\S~\ref{sec:model}), and workflow design~(\S~\ref{sec:workflow}). For data strategies, we provide a detailed discussion on data engineering works for building LCLMs across both pre-training and post-training phase, including data selection, data filtering, data synthesis, data mixture, etc. For architecture design, we first examine position embeddings commonly adopted by LCLMs, and their extrapolation strategies, then we systematically discuss three major
types of architectural designs: transformer-based modifications, linear-complexity architectures, and hybrid approaches that integrate both paradigms. On top of this, we further introduce workflow designs that expand the scope of a single LCLM, including prompt compression, memory-based workflow, RAG-based workflow, and agent-based workflow. 

Second, in order to efficiently train and deploy LCLMs (\textbf{RQ2}), we thoroughly summarize AI infrastructure optimization strategies~(\S~\ref{sec:infra}). For training infrastructure, we examine I/O optimization, optimizations on GPU constraints \& memory access, and optimizations on communication-computation overlapping. For inference infrastructure, we review five types of efficient inference strategies, including quantization, memory management, prefilling-decoding disaggregated architecture, GPU-CPU parallel inference, and speculative decoding.

Third, building on top of effective LCLMs and efficient infrastructure, we proceed to discuss their performance evaluation~(\S~\ref{sec:evaluation}) and analysis~(\S~\ref{sec:analysis}) (\textbf{RQ3}). The evaluation section divides long context capabilities into two broad categories: long context comprehension and long-form generation. For each category, we discuss the evaluation paradigm, and present an overview of existing benchmarks. For the analysis section, we include both the external performance analysis (e.g.,  the effective context length, PPL metric, lost-in-the-middle) and the internal model structure analysis (e.g., position embedding, attention head, mlp layers).

Finally, in Section~\ref{sec:application}, we summarize the main applications for long context large language models (LLMs), including the agent, RAG, Code, multimodal tasks and etc,
and in Section~\ref{sec:directions}, we propose five potential \textbf{future directions} for LCLMs including the long Chain-of-Thought reasoning, effective long context extension, efficient architecture and infrastructure, robust evaluation, and mechanistic interpretability.

As shown in Table~\ref{tab: benchmark_compare},
we also compare with this survey paper with existing surveys~\citep{Liu2025ThusSL, DBLP:journals/corr/abs-2302-14502, DBLP:journals/corr/abs-2401-07872} on long context modeling,
and we observe that existing surveys usually focus on several specific topics in long context modeling.
In contrast, 
this comprehensive survey provides an in-depth exploration of the rapidly evolving landscape of LCLMs,
and address the aforementioned RQs for long context modeling by covering a wide range of topics.

In summary, by providing a comprehensive overview of the current state of long context language models, this survey aims to serve as a valuable resource for researchers, practitioners, and enthusiasts in the field of NLP. We hope to shed light on the progress made thus far, highlight the remaining challenges, and inspire future innovations in this exciting research area.

\begin{table}[t]
    \centering
    \footnotesize
    \renewcommand{\arraystretch}{1.2}
        \setlength\tabcolsep{3.5pt}

    \resizebox{\textwidth}{!}
    {
    \begin{tabular}{lccccccc}
    \toprule
    \textbf{Survey} & \textbf{Data} & \textbf{Architecture} & \textbf{Workflow Design} & \textbf{Infrastructure} & \textbf{Evaluation} & \textbf{Analysis}  \\
    \midrule
    \citet{huang2024advancing}&$\texttimes$ & $\checkmark $ &  $\texttimes$  &  $\texttimes$ & $\texttimes $ & $\texttimes $ \\
    \citet{zhao-etal-2024-length}& $\texttimes$ & $\checkmark $ &  $\texttimes$  &  $\texttimes$ & $\texttimes $ & $\texttimes $ \\
    \citet{Li2024PromptCF} & $\texttimes$ & $\texttimes $ &  $\checkmark$  &  $\texttimes$ & $\texttimes $ & $\texttimes $ \\
    \citet{DBLP:journals/corr/abs-2302-14502}& $\texttimes$ & $\checkmark $ &  $\texttimes$  &  $\texttimes$ & $\texttimes $ & $\texttimes $ \\
    \textbf{Ours} &$\checkmark$ & $\checkmark$ &  $\checkmark$ &  $\checkmark$ & $\checkmark$ & $\checkmark$  \\ 
    \bottomrule
    \end{tabular}
    }
    \caption{Comparisons between our survey and other related long context modeling surveys.}
    \label{tab: benchmark_compare}
\end{table}

\section{Data} 
\label{sec:data}

\begin{figure}[!htp]
    \centering
    \includegraphics[width=1.0\textwidth]{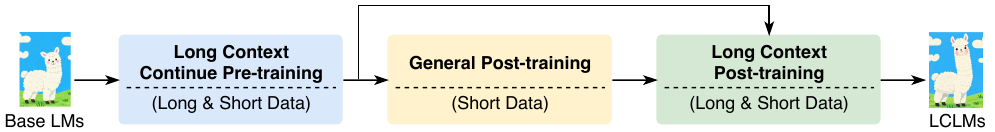}
    \caption{Illustration of training pipeline of LCLMs.}
    \label{fig:data_arg}
\end{figure}

In this section, we discuss the data-related topics for long context modeling. Specifically, 
as shown in Figure~\ref{fig:data_arg},
the long context data is used both in the pre-training and post-training stages.
To illustrate the data processing pipeline for LCLMs,
we first provide the data strategies of long context pre-training data~(\S~\ref{sec:pt_data}), including data filtering, data mixture and data synthesis.
Then, we illustrate the data strategies of long context post-training data~(\S~\ref{sec:sft_data}), including data filtering and data synthesis.

\subsection{Pre-training}
\label{sec:pt_data}

\subsubsection{Data Filtering}
Foundation models' performance is significantly influenced by the quality of pre-training data as shown in many studies~\citep{gopher, phi, glam, zhang2024mapneo},
and various approaches have been employed to enhance data quality. 
Specifically, some works~\cite{raffel2020exploring,slimpajama,zhang2024mapneo} have implemented many heuristic rules (e.g., removing short entries) and duplication methods (e.g., MinHashLSH deduplication~\citep{leskovec2020mining}) to improve the data quality. 
Besides, SemDeDup~\citep{SemDeDup} is proposed to utilize the embeddings from existing pre-trained models to remove semantic duplicates.
In addition, several methods have been proposed to improve the diversity and complexity of the data. For example, ~\citet{D2} viewed diversity and difficulty as complementary factors, employing a dataset graph with bidirectional message passing for data selection,
and 
~\citet{D4} applied clustering techniques to broaden the data diversity. 
Nevertheless, the aforementioned approaches primarily address standard pre-training datasets, typically limited to 4,000 or 8,000 tokens. Recently, researchers have developed specialized criteria to evaluate the quality of extended context from various perspectives. For example, the work~\citep{liu-etal-2024-longwanjuan} introduced a comprehensive framework for assessing the quality of long texts,
which mainly includes three core linguistic aspects: coherence, cohesion, and complexity.
To quantify these aspects, various metrics including statistical measures and evaluations based on pre-trained language models have been proposed.
Meanwhile, apart from Longwanjuan, 
another representative work~\citep{prolong} is proposed to evaluate training samples based on their long-range dependency characteristics. This approach also utilizes three key metrics (i.e., dependency strength, dependency distance, dependency specificity) to assess and prioritize samples that are particularly beneficial for enhancing models' long context comprehension.
Recently, some works have begun to use the attention pattern to select high-quality long context data.
For example,
the LongAttn~\citep{wu2025longattn} uses the self-attention mechanism to quantify the long
long-range dependencies for
accurate and efficient data selection.

\subsubsection{Data Mixture}
Language model training typically utilizes datasets drawn from diverse sources~\citep{10.5555/3495724.3495883,10.5555/3648699.3648939,Du2021GLaMES,Gao2020ThePA,gao2024prolong}. For instance, The Pile~\citep{Gao2020ThePA}, a widely accessible dataset, comprises various components including 24\% web content, 9\% Wikipedia, and 4\% GitHub, etc.
Besides, many works have observed that 
the composition of pre-training data significantly impacts an LM's performance~\citep{Du2021GLaMES,hoffmann2022an}, 
and current approaches to determining domain weights (sampling probabilities for each source) often rely on intuition or specific downstream tasks. For example, The Pile employs heuristically chosen weights, which may not be optimal. 
PaLM~\citep{10.5555/3648699.3648939} and GLaM~\citep{Du2021GLaMES} fine-tune these weights based on selected downstream tasks, which need to training numerous LMs with different weight configurations.
To address this limitation, 
Data Mixing Laws~\citep{ye2024data} and RegMix~\citep{liu2024regmix} investigate the data-mixing scaling law to find the optimal domain mixture for improving the pre-training efficiency with minimal training costs.
However, the above-mentioned methods usually focus on pre-training data within limited context windows (e.g., 4k/8k),
and several works  have begun to focus on the data mixture on the long context pre-training.
For example,
~\citep{Fu2024DataEF} explore five types of long context data composition including ``Cutting documents at 4K'',  ``{Cutting documents at 128K}'', ``Global upsampling'', ``Per-source upsampling'' and ``Upsampling Arxiv / Book / Github'' and have following observations:
(1) Continual pre-training on a small amount of long context data can significantly improve a 7B model's capability 
 to accurately retrieve information over long context input. 
(2) When scaling up context length, it is crucial to oversample lengthy sequences while maintaining the original domain diversity of the pre-training datasets.
ProLong~\citep{gao2024prolong}
shows that incorporating code repositories and long books as long context sources, and mixing them with high-quality short-context sources is vital for improving performance on extended inputs while preserving the model's proficiency with short contexts.
The GrowLength approach~\citep{growlength} progressively expands the input size throughout the pre-training process, optimizing computational resources and boosting efficiency.

\subsubsection{Data Synthesis}
As the long context data is usually rare in the real-world corpus (e.g., web documents),
many methods have been proposed to synthesize or construct the long context pre-training data.
For example,
one approach clusters semantically related texts within a single context window~\citep{realm}, while another work~\citep{levine2022the} demonstrates improved sentence representations by incorporating related, non-adjacent sentences in pre-training examples. 
Besides, to address document redundancy, a traveling salesman algorithm has been employed in ICP~\citep{Shi2023InContextPL}. 
Moreover, SPLICE~\citep{spacking} involves structured packing, where training samples are created by combining multiple similar retrieved documents. Additionally, a query-centric synthesis method Quest~\citep{quest} has been proposed to aggregate diverse yet semantically linked documents. This approach utilizes a generative model to predict potential queries for each document, subsequently grouping documents with similar queries and keywords. Another method~\citep{tian2024utk} creates complex "knotted" text structures by shuffling document chunks, training models to untangle and locate relevant segments, thereby improving both contextual attention and training efficiency.

\subsection{Post-training}
\label{sec:sft_data}

\subsubsection{Data Filtering}
Unlike pre-training,
the data filtering strategy in post-training usually aims to select influential samples to empower the LLMs’ instruction-following capabilities.
\citet{chen2024alpagasus, liu2024what, Chen2025LADMLT, wu2025longattn} explore the use of feedback from proprietary language models to curate training samples.
Meanwhile,
\citet{cao2024instruction, Li2023FromQT, ge-etal-2024-clustering, xia2024less} 
have developed sophisticated metrics based on open-source LLMs to evaluate the importance of samples.
However, these methods only focus on selecting short-from SFT data, overlooking the specific challenges posed by long context alignment.
Recently,
the GATEAU~\citep{geteau} introduces two components (i.e., Homologous Models' Guidance and Contextual Awareness Measurement) to identify the influential samples with long-range dependency relations.

\begin{table*}[t]
    \centering
    \footnotesize
    \renewcommand{\arraystretch}{1.25}
    
    \begin{tabular}{llc}
    \toprule
    {\textbf{Training Data}}  &{\textbf{Characteristics}} & \textbf{Stage}  \\ 
    \midrule
    Longwanjuan~\citep{liu-etal-2024-longwanjuan} &   Bilingual, 
     filtered from SlimPajama~\citep{slimpajama} and Wanjuan~\citep{He2023WanJuanAC} & Pre-training  \\
    Long-Data-Collections~\citep{longdata} &  A wide variety of data sources &Pre-training\\
    LongAttn~\citep{wu2025longattn} & Long-range dependency selected using attention patterns & Pre-training \\
    \midrule
LongAlign~\citep{longalign}& Diverse tasks and various sources, Self-Instruct & Post-training\\
FILM~\citep{an2024make} & Information-Intensive, context length balance, multi-hop reasoning& Post-training\\
PAM QA~\citep{pamqa}&Position-agnostic,  multi-hop reasoning& Post-training\\
LongAlpaca~\citep{longlora}&Self-collected, instruction following&Post-training\\
ChatQA2~\citep{Xu2024ChatQA2B}& Synthesized from NarrativeQA~\citep{narrativeqa}&Post-training\\
LongMIT~\citep{Chen2024WhatAT} &Multi-hop, diverse, automatic&Post-training \\ 
LongWriter-6k~\citep{bai2024longwriter}&Long-form generation, output lengths ranging from 2k to 32k word&Post-training\\
Long Reward~\citep{zhang2024longreward}&Bilingual, preference optimization&Post-training\\
LOGO~\citep{tang2024logo}&Preference optimization&Post-training\\
LongDPO~\citep{ping2025longdpounlockbetterlongform}&Long-form Generation, preference optimization, step-level&Post-training\\
LongFaith~\citep{yang2025longfaith}&Faithfulness, reasoning, preference optimization&Post-training\\

     \bottomrule
    \end{tabular}%
    \caption{Overview of training datasets for long context modeling.}
    \label{tab:dataset_compare}
\end{table*}

\subsubsection{Data Synthesis}
For the data synthesis of post-training, 
we need to construct the long context queries effectively,
and many works have been proposed.
For example,
Ziya-Reader~\citep{pamqa} constructs a tailored Multi-doc QA task that requires concentration on different positions in contexts to address the ``lost in the middle'' problem~\citep{liu2024lost}. 
~\citet{xiong-etal-2024-effective} proposes to select a document from pre-training corpus and prompt the language model to write question-answer pairs based on information in the selected chunk.
Recently, ~\citet{an2024make} presents information-intensive training to overcome lost-in-the-middle,
which leverages a synthesized long context question-answer dataset including two types of questions (i.e., fine-grained information awareness on exactly one short segment, and the integration and reasoning of information from two or more segments).
~\citet{Xu2024ChatQA2B} introduce to assemble all the related
paragraphs and randomly insert the ground truth summary to simulate a real long document for obtaining long context instructions.
~\citet{Chen2024WhatAT} propose the
 Multi-agent Interactive Multihop Generation (MIMG) framework, incorporating a Quality Verification Agent,
a Single-hop Question Generation Agent, a Multiple Question Sampling Strategy, and a Multi-hop Question Merger Agent, to improve the quality of long context instruction data.

Recently, several works~\citep{tang2024logo,zhang2024longreward} have begun to focus the preference optimization on long context models~\citep{Rafailov2023DirectPO,Ouyang2022TrainingLM} to align models with human preferences. 
A representative work DPO~\citep{Rafailov2023DirectPO} can eliminate the need for a separate reward model,
and teach the model to ``reject'' misaligned responses and ``accept'' preferred responses with differently assigned prediction scores.
Meanwhile, many efforts have been made to enhance the effectiveness and efficiency of DPO, such as SimPO~\citep{Meng2024SimPOSP}, ORPO~\citep{orpo}, TPO~\citep{saeidi2024triple}, 2D-DPO~\citep{Li20242DDPOSD}, and Constitutional DPO~\citep{wang2024weaver}.
However, these diverse approaches mainly focus on short-context scenarios, and the long context preference optimization is ignored.
Recently, several works for long context preference optimization have been proposed.
For example, 
LongReward~\citep{zhang2024longreward} utilizes an off-the-shelf LLM to provide rewards for long context model responses from four human-valued dimensions: helpfulness, logicality, faithfulness, and completeness,
and then obtain the preference pairs based on the synthesized queries.
LOGO~\citep{Tang2024LOGOL} introduces the importance of scoring with an automatic evaluator to synthesize preference (aligned) and dispreference (misaligned) data in long context comprehension scenarios.
More recently, long context generation has drawn great attentions, and ~\citet{ping2025longdpounlockbetterlongform} propose the LongDPO method, which uses the Monte Carlo Tree Search to gather stepwise preference pairs and adopt the step-level DPO  to improve the capabilities of existing  LLMs.

\subsection{Training Data}
Summary on the long context datasets for pre-training and post-training are as shown in Table~\ref{tab:dataset_compare}.

\section{Architecture}

\label{sec:model}

\tikzstyle{my-box}=[
    rectangle,
    draw=hidden-draw,
    rounded corners,
    text opacity=1,
    minimum height=1.5em,
    minimum width=5em,
    inner sep=2pt,
    align=center,
    fill opacity=.5,
    line width=0.8pt,
]
\tikzstyle{leaf}=[my-box, minimum height=1.5em,
    fill=hidden-pink!80, text=black, align=left,font=\normalsize,
    inner xsep=2pt,
    inner ysep=4pt,
    line width=0.8pt,
]
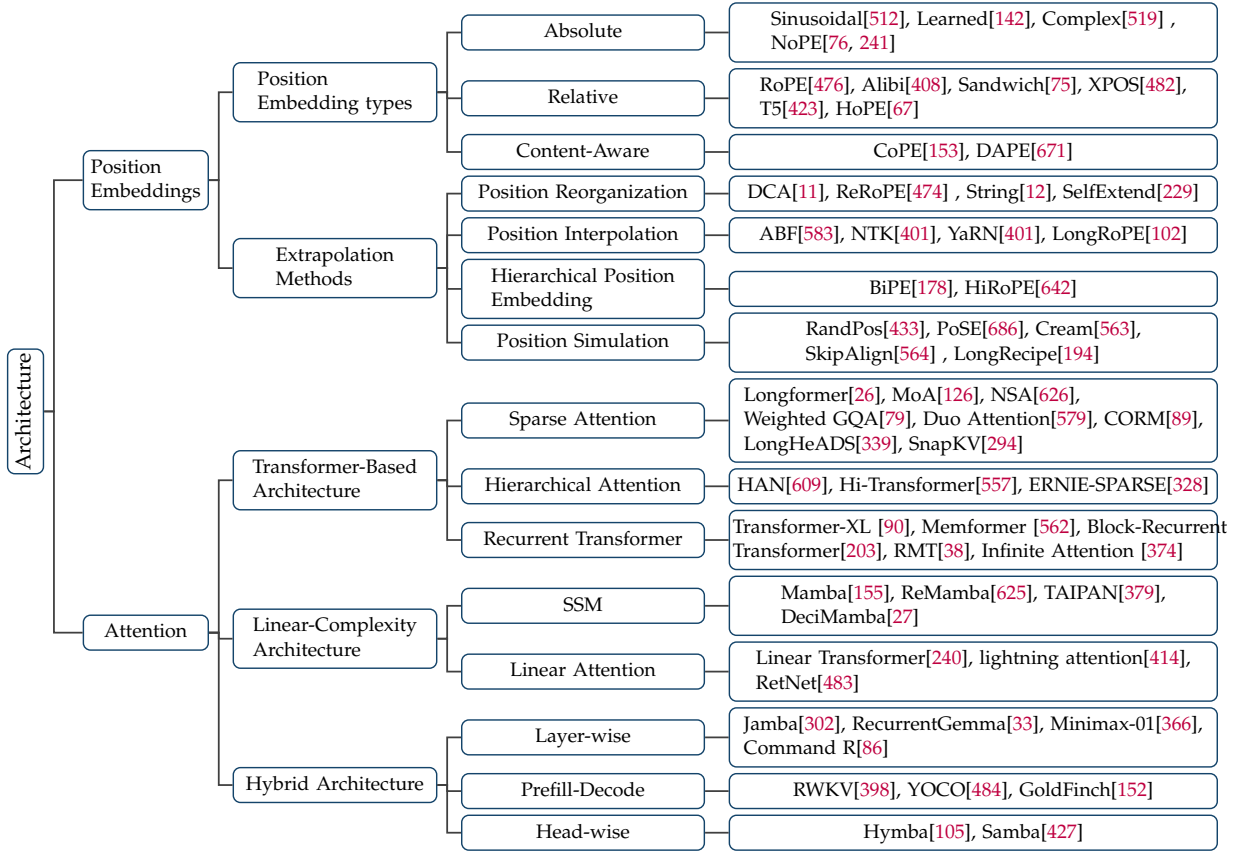
\begin{figure*}[t]
	\centering
    \resizebox{\textwidth}{!}{
	\begin{forest}
        forked edges,
		for tree={
                grow=east,
                reversed=true,
                anchor=base west,
                parent anchor=east,
                child anchor=west,
                base=center,
                font=\large,
                rectangle,
                draw=hidden-draw,
                rounded corners,
                align=left,
                text centered,
                minimum width=4em,
                edge+={darkgray, line width=1pt},
                s sep=3pt,
                inner xsep=2pt,
                inner ysep=3pt,
                line width=0.8pt,
                ver/.style={rotate=90, child anchor=north, parent anchor=south, anchor=center},
            },
            where level=1{text width=6em,font=\normalsize, }{},
            where level=2{text width=10em,font=\normalsize}{},
            where level=3{text width=12em,font=\normalsize,}{},
            where level=4{text width=20em,font=\normalsize,}{},
	    [Architecture, ver
			[Position \\Embeddings
            [
                Position \\Embedding types
                [Absolute
    				    [
                            Sinusoidal\cite{vaswani2017attention}{, }Learned\cite{gehring2017convolutional}{, }Complex\cite{wang2019encoding} {, }\\NoPE\cite{chi2023latent,kazemnejad2024impact}
                           , text width=24.5em
                            ]
    			     ]
    			  [Relative
    					[
        					RoPE\cite{su2024roformer}{, }Alibi\cite{press2021train}{, }Sandwich\cite{chi2022dissecting}{, }XPOS\cite{sun2022length}{, }\\T5\cite{raffel2020exploring}{, }HoPE\cite{chen2024hope}{, }BAM \cite{bianchessi2025bayesian}
                           , text width=24.5em
    					]
    			  ]
                    [Content-Aware
                        [
                            CoPE\cite{golovneva2024contextual}{, }DAPE\cite{zheng2025dape}
                           , text width=24.5em
                        ]
                    ]
            ]
            [
                Extrapolation\\ Methods
                [
                    Position Reorganization
                    [
                    DCA\cite{an2024training}{, }ReRoPE\cite{kexuefm-9708} {, }String\cite{an2024does}{, }SelfExtend\cite{jin2024llm}
                   , text width=24.5em
                    ]
                ]
                [
                    Position Interpolation
                    [
                    ABF\cite{xiong2023effective}{, }NTK\cite{peng2023yarn}{, }YaRN\cite{peng2023yarn}{, }LongRoPE\cite{ding2024longrope}
                   , text width=24.5em
                    ]
                ]
                [
                    Hierarchical Position \\Embedding
                    [
                    BiPE\cite{he2024two}{, }HiRoPE\citep{zhang2024hirope}
                  , text width=24.5em
                    ]
                ]
                [
                    Position Simulation
                    [
                    RandPos\cite{ruoss2023randomized}{, }PoSE\cite{zhu2023pose}{, }Cream\cite{wuefficient}{, }\\SkipAlign\cite{wu2024long} {, }LongRecipe\cite{hu2024longrecipe}
                   , text width=24.5em
                    ]
                ]
            ]
			]
			[Attention, 
				    [Transformer-Based \\Architecture
    				[
                    Sparse Attention
                            [
                            Longformer\cite{beltagy2020longformerlongdocumenttransformer}{, }MoA\cite{fu2024moamixturesparseattention}{, }NSA\cite{yuan2025nativesparseattentionhardwarealigned}{, }\\Weighted GQA\cite{chinnakonduru2024weightedgroupedqueryattention}{, }Duo Attention\cite{xiao2024duoattentionefficientlongcontextllm}{, }CORM\cite{dai2024cormcacheoptimizationrecent}{, }\\LongHeADS\cite{lu2024longheadsmultiheadattentionsecretly4}{, }SnapKV\cite{li2024snapkvllmknowslooking}
                           , text width=24.5em
                            ]
                    ]
                    [Hierarchical Attention
                            [
                            HAN\cite{han}{, }Hi-Transformer\cite{hi-transformer}{, }ERNIE-SPARSE\cite{ERNIE-SPARSE}
                           , text width=24.5em
                            ]
                    ]
                    [Recurrent Transformer
                            [
                            Transformer-XL~\cite{TransformerXL}{, }Memformer~\cite{Memformer}{, }Block-Recurrent\\Transformer\cite{BlockRecurrentTransformers}{, }RMT\cite{rmt}{, }Infinite Attention~\cite{infinitransformer}
                           , text width=24.5em
                            ]
                    ]
			]
                    [Linear-Complexity \\Architecture
                        [
                        SSM
                        [
                            Mamba\cite{gu2023mamba}{, }ReMamba\cite{yuan2025remambaequipmambaeffective}{, }TAIPAN\cite{vannguyen2024taipanefficientexpressivestate}{, }\\DeciMamba\cite{benkish2024decimambaexploringlengthextrapolation}
                           , text width=24.5em
                            ]
                        ]
                        [
                        Linear Attention
                        [
                            Linear Transformer\cite{katharopoulos2020transformersrnnsfastautoregressive}{, }lightning attention\cite{qin2024lightningattention2freelunch}{, }\\RetNet\cite{sun2023retentivenetworksuccessortransformer}
                           , text width=24.5em
                            ]
                        ]
                        [
                        Test-Time Training
                        [
                            Titans\cite{behrouz2024titans}{, }Atlas\cite{behrouz2025atlas}{, }TTT Done Right\cite{zhang2025test}
                           , text width=24.5em
                            ]
                        ]
                    ]
				[Hybrid Architecture
					[
                        Layer-wise
                        [
                            Jamba\cite{lieber2024jamba}{, }RecurrentGemma\cite{recurrentgemma}{, }Minimax-01\cite{minimax2025minimax01scalingfoundationmodels}{, }\\Command R\cite{c4ai-command-r7b-12-2024}
                           , text width=24.5em
                            ]
					]
                    [
                    Prefill-Decode
                    [
                            RWKV\cite{peng2023rwkvreinventingrnnstransformer}{, }YOCO\cite{sun2025you}{, }GoldFinch\cite{goldstein2024goldfinch}
                           , text width=24.5em
                            ]
                    ]
                    [
                    Head-wise
                    [
                            Hymba\cite{dong2024hymba}{, }Samba\cite{ren2024samba}
                           , text width=24.5em
                            ]
                    ]
				]	
                ]
                ]
	\end{forest}}
	\caption{Taxonomy of Long Context Model Architectures.}
    \label{fig:taxonomy-model}
\end{figure*}

For long context language models, architectural design faces dual challenges: effectively processing extremely long texts while maintaining computational efficiency in both training and inference phases. This section presents a comprehensive review of architectural designs tailored for long context modeling. First, we examine position embeddings commonly adopted by LCLMs, and their extrapolation strategies~(\S~\ref{sec:pos_emb}). Then, we systematically discuss three major types of architectural designs: transformer-based modifications~(\S~\ref{ssec:transformer_based_arch}), linear-complexity architectures~(\S~\ref{ssec:linear_comp_arch}), and hybrid approaches that integrate both paradigms~(\S~\ref{ssec:hybrid_arch}).

\subsection{Position Embeddings}
\label{sec:pos_emb}
Most long context LLMs are based on the popular Transformer architecture. Since the Transformer computes the representations of all tokens in parallel, it is necessary to integrate positional information within the sequence with the aid of positional encoding. In this subsection, we will first introduce different types of positional embeddings used in the Transformer, and then demonstrate the extrapolation methods for adapting the positional encoding to larger sequence lengths. 

\subsubsection{Position Embedding Types}

As shown in Table \ref{tab:pe}, positional embeddings in LLMs can be classified into absolute positional embeddings, relative positional embeddings, and content-aware positional embeddings according to the form of their positional information.
\paragraph{Absolute Position Embedding} Absolute Position Embedding provides each token with information about its absolute position in the sequence. This approach can be categorized into three types: (1) directly specifying the positional embedding (PE) for each position through a functional approach; (2) setting the PE for each position as trainable vectors that are updated during training; and (3) omitting positional embeddings entirely and relying on unidirectional attention in decoder-only architectures to learn positions implicitly. Below are the details:
\begin{itemize}
\item \textit{Functional Positional Embedding~\citep{vaswani2017attention}}. The most representative example of functional position embedding is Sinusoidal Positional Embedding, which was proposed by the vanilla Transformer~\citep{vaswani2017attention} and was widely used in variant models based on Transformers. It encodes positional information through periodic sine and cosine functions. In Transformer, it is usually added to word embeddings to inject absolute positional information into the network. 
\item \textit{Learned Positional Embedding \cite{gehring2017convolutional}}. It was firstly proposed by Meta to introduce positional information into the convolutional sequence-to-sequence learning. The embedding representations of different positions are treated as learnable parameters and trained along with the network. This is widely used in models such as BERT~\citep{devlin-etal-2019-bert}, GPT~\citep{10.5555/3495724.3495883}, and OPT~\citep{zhang2022opt_model_pretrained}. There are also some variants~\cite{wang2019encoding} that represents the position embedding in the complex plane and combines it with the continuous word embedding which is also represented in complex vectors.

\item \textit{No Position Embedding (NoPE)~\cite{chi2023latent,kazemnejad2024impact}.} NoPE proposes to incorporate no explicit position encoding into Transformer. It founds that the model still learns the permutation order of tokens. NoPE is also effective in language modeling and several probing tasks. \citep{chi2023latent} prove that NoPE in causal language models can incorporate implicit absolute positional information into the models. Note that NoPE also conveys relative positional information \cite{kazemnejad2024impact}. 
\end{itemize}
\paragraph{Relative Position Embedding} Instead of focusing on absolute positional information, Relative Position Embedding~\citep{shaw2018self} captures the relative distances between tokens, as it assumes that relative positional information is more crucial for language understanding. For LCLMs, commonly used Relative Position Embeddings can be categorized into following three types:
\begin{table}[t!]
\centering
\footnotesize
\renewcommand{\arraystretch}{1.1}
\begin{tabular}{lcccc}
\toprule
\bf Category                    & \bf Position Embedding                          & \bf Parametric & \bf Representation & \bf Injection Method \\ \midrule
\multirow{4}{*}{Absolute}      & Sinusoidal \cite{vaswani2017attention}         & \XSolidBrush      & Embedding         & Add                 \\
                               & Learned \cite{gehring2017convolutional}        & \Checkmark      & Embedding         & Add                 \\
                               & Complex \cite{wang2019encoding}                & \Checkmark      & Embedding         & Multiply            \\
                               & NoPE \cite{chi2023latent,kazemnejad2024impact} & \XSolidBrush      & NA                & NA                  \\ \midrule
\multirow{9}{*}{Relative}      & Relative Position \cite{shaw2018self}          & \Checkmark      & Embedding         & Add                 \\
                               & T5 \cite{raffel2020exploring}                  & \Checkmark      & Bias              & Add                 \\
                               & Alibi \cite{press2021train}                    & \XSolidBrush      & Bias              & Add                 \\
                               & Kerple \cite{chi2022kerple}                    & \Checkmark      & Bias              & Add                 \\
                               & Sandwich \cite{chi2022dissecting}              & \XSolidBrush      & Embedding         & Add                 \\
                               & FIRE \cite{li2023functional}                   & \Checkmark      & Bias              & Add                 \\
                               & RoPE \cite{su2024roformer}                     & \XSolidBrush      & Embedding         & Multiply            \\
                               & XPOS \cite{sun2022length}                      & \XSolidBrush      & Embedding         & Multiply            \\
                               & HoPE \cite{chen2024hope}                       & \XSolidBrush      & Embedding         & Multiply            \\
                               & BAM \cite{bianchessi2025bayesian}                       & \XSolidBrush      & Bias         & Add            \\
                               \midrule
\multirow{2}{*}{Content-Aware} & CoPE \cite{golovneva2024contextual}            & \Checkmark      & Embedding         & Add                 \\
                               & DAPE \cite{zheng2025dape,zheng2024dape}                      & \Checkmark      & Bias              & Add                 \\ \bottomrule
\end{tabular}
\caption{Overview of position embeddings for the Transformer architecture.}
\label{tab:pe}
\end{table}
\begin{itemize}
\item \textit{T5-Style~\cite{raffel2020exploring}}. T5 initially maps the relative distance $(i - j)$ between tokens at positions $i$ and $j$ to a scalar bias value $b = f(i - j)$, where $f$ represents a lookup table. Subsequently, the relative bias $b$ (learned during the training process) is incorporated into the dot product of the query and key within the self-attention mechanism. The lookup table assigns the same parameter to distances exceeding a certain threshold, thereby facilitating generalization to unobserved distances. Base on T5's position embedding, FIRE~\cite{li2023functional} proposes to map the positional representations as a learnable function. To solve the length generalization issue when the test length varies, FIRE proposes progressive interpolation by normalizing the distance by query position index. In this way, FIRE can map any test length into the training range.
\item \textit{ALiBi~\cite{press2021train} and its Variants}. ALiBi, which is adopted in BLOOM~\citep{Scao2022BLOOMA1} and Baichuan~\citep{Yang2023Baichuan2O}, is similar to T5's position encoding. However, it differs in that it subtracts a scalar bias from the attention score. This bias increases linearly in proportion to the distance between the query and key tokens. As a result, this actually generates a preference for tokens that are closer in position, which is known as a recency bias. Alibi can be formulated as the following:
\begin{equation}
    \mathbf{q}_i \mathbf{k}_j^T = (\mathbf{x}_i \mathbf{W}_q)(\mathbf{x}_j \mathbf{W}_k)^T + m(j-i)
\end{equation}
where $m$ is a head-specific scalar hyper-parameter. $i$ and $j$ mean the position indices of $\mathbf{q}$ and $\mathbf{k}$. $\mathbf{W}_q$ and $\mathbf{W}_k$ represent the projection layer in the attention. Building on top of ALiBi, Kerple~\cite{chi2022kerple} introduces two trainable parameters for better length extrapolation, while Sandwich \cite{chi2022dissecting} simplifies the form of sinusoidal position embedding by only considering the cross term of position embeddings to alleviate the overfitting issue of sinusoidal position embedding. BAM \cite{bianchessi2025bayesian} frames attention as a Bayesian mechanism and adopts a Generalized Gaussian Distribution as prior, achieving improved length extrapolation capability.
\item \textit{Rotary Position Embedding (RoPE) \cite{su2024roformer} and its variants}. RoPE rotates the query and key representations by an angle that is proportional to their absolute positions prior to dot product attention. Owing to this rotation, the attention dot product will rely solely on the relative distance between tokens, thereby effectively transforming it into a relative positional encoding. To elucidate, given a hidden vector $\mathbf{h} =[h_0,h_1,...,h_{d-1}]$, where $d$ is the hidden dimension, and a position index $m$, RoPE operates as follows:
\begin{equation}
	f(\mathbf{h},m) = 
	\begin{pmatrix}
		h_0\\
		h_1\\
		h_2\\
		h_3\\
		\vdots\\
		h_{d-2}\\
		h_{d-1}
	\end{pmatrix}
	\otimes
	\begin{pmatrix}
		\cos{m\theta_0} \\
		\cos{m\theta_0} \\
		\cos{m\theta_1} \\
		\cos{m\theta_1} \\
		\vdots \\
		\cos{m\theta_{d/2-1}} \\
		\cos{m\theta_{d/2-1}} 
	\end{pmatrix}
	+
	\begin{pmatrix}
		-h_1\\
		h_0\\
		-h_3\\
		h_2\\
		\vdots\\
		-h_{d-1}\\
		h_{d-2}
	\end{pmatrix}
	\otimes
	\begin{pmatrix}
		\sin{m\theta_0}\\
		\sin{m\theta_0}\\
		\sin{m\theta_1}\\
		\sin{m\theta_1}\\
		\vdots\\
		\sin{m\theta_{d/2-1}}\\
		\sin{m\theta_{d/2-1}}
	\end{pmatrix}
\end{equation}
where $\theta_j=10000^{-2j/d},j\in\{0,1,...,d/2-1\}$. 
Given a query $\mathbf{q}$ at position $m$ and a key $\mathbf{k}$ at position $n$, attention score $a(\mathbf{q},\mathbf{k})$ is defined as:
\begin{align} 
a(\mathbf{q},\mathbf{k})&=<f(\mathbf{q},m), f(\mathbf{k},n)> \nonumber \\
&=\sum_{j=0}^{d/2-1}[(q_{2j}k_{2j}+q_{2j+1}k_{2j+1})\cos{(m-n)\theta_{j}}+(q_{2j}k_{2j+1}-q_{2j+1}k_{2j})\sin{(m-n)\theta_{j}}] \nonumber \\
&:=g(\mathbf{q},\mathbf{k},(m-n)\boldsymbol{\theta}) 
\label{eqn:attention_score}
\end{align}
where g(·) is an abstract mapping function exclusively dependent on $\mathbf{q},\mathbf{k}$ and $(m-n)\boldsymbol{\theta}$. Benefiting from this good nature, RoPE becomes the most pervasive position embedding strategy in the era of LLMs, including LLaMA~\citep{touvron2023llama},  Qwen~\citep{yang2025qwen2}, etc. \citet{sun2022length} ascribes that the weak extrapolation ability of RoPE is related to the significant oscillation in their attention expectations. To address this problem, they propose XPOS that incorporates a balancing term to penalize the oscillation of unstable dimensions while maintaining the distribution of stable dimensions. Also building upon RoPE's foundation, HoPE \cite{chen2024hope} enhances length extrapolation by replacing specific components with position-independent ones while retaining only high-frequency signals.
\end{itemize}
\paragraph{Content-Aware Position Embedding} More recently, there has been a growing interest in Content-Aware Position Embedding~\cite{golovneva2024contextual,zheng2025dape}, which argues that position measurement should take into account more semantically meaningful units such as words or sentences, as illustrated below:
\begin{itemize}
    \item CoPE \cite{golovneva2024contextual}. CoPE considers the joint modeling of content and position information by leveraging the content of the tokens and their positions in the sequence. Specifically, CoPE first calculates context-dependent gate values, and then employs these values to determine token positions through a cumulative summation process. In this way, positions embeddings are equipped with contextualized information.
    \item DAPE \cite{zheng2025dape,zheng2024dape}. It proposes to model the positional information dynamically with the attention. Specifically, DAPE determines the position bias by not only the position indices but also the semantics information. In this way, DAPE overcomes the inflexibility and achieves relatively optimal performance for each individual instance by dynamically adjusting on each specific input data.
\end{itemize}

\begin{table}[t!]
\centering
\footnotesize
\renewcommand{\arraystretch}{1.1}
\setlength\tabcolsep{30pt}
\begin{tabular}{llc}
\toprule
\textbf{Intuition} & \textbf{Method}  & \textbf{Training-free} \\
\midrule
\multirow{4}{*}{Position Reorganization} & SelfExtend \cite{jin2024llm} & \Checkmark           \\
& DCA \cite{an2024training}             & \Checkmark                     \\
& ReRoPE \cite{kexuefm-9708}          & \Checkmark                         \\
& String \cite{an2024does}          & \Checkmark              \\ 
\midrule
\multirow{9}{*}{Position Interpolation} & PI \cite{chen2023extending}               & \Checkmark          \\
& NTK \cite{peng2023yarn}             & \Checkmark             \\
& ABF \cite{xiong2023effective}             & \Checkmark         \\
& YaRN \cite{peng2023yarn}            & \Checkmark                         \\
& Truncated Basis \cite{pal2023giraffe}  & \Checkmark                         \\
& CLEX  \cite{chen2023clex}           & \XSolidBrush                       \\
& Resonance RoPE \cite{wang2024resonance}            & \XSolidBrush                         \\
& LongRoPE \cite{ding2024longrope}        & \XSolidBrush                       \\
& MsPoE \cite{zhang2024found}          & \Checkmark                        \\
\midrule
\multirow{2}{*}{Hierarchical Position} & BiPE \cite{he2024two}            & \XSolidBrush                       \\
& HiRoPE~\citep{zhang2024hirope} & \Checkmark                      \\   
\midrule
\multirow{5}{*}{Position Simulation} & RandomPE \cite{ruoss2023randomized}         & \XSolidBrush              \\
& PoSE \cite{zhu2023pose}            & \XSolidBrush                    \\
& SkipAlign \cite{wu2024long}            & \XSolidBrush                     \\
& Cream \cite{wuefficient}            & \XSolidBrush                      \\
& LongRecipe~\citep{hu2024longrecipe} & \XSolidBrush                      \\
\bottomrule
\end{tabular}
\caption{RoPE variants of length generalization.}
\label{tab:extrapolation}
\end{table}
\subsubsection{Extrapolation Methods of Position Embeddings}

For a language model with an original context window size of $L_o$, when processing a target sequence of length $L_t$ (with scaling factor $\alpha=L_t/L_o$) that exceeds this range, the first challenge is the position encoding OOD problem, as $L_o$ position encodings cannot cover a larger range. As mentioned in LM-Infinite~\citep{han-etal-2024-lm}, position encoding OOD is a significant factor hindering length extrapolation. From the perspective of avoiding out-of-distribution positions, position encoding-based length extrapolation strategies can be mainly divided into two categories: \textbf{(1)} Mapping the target sequence to the position range supported by the model, and \textbf{(2)} Enabling the model to support position ranges larger than the context window size. Notably, most methods in the first category can perform well without additional tuning, while designs in the second category often rely on training to be effective.

Mapping target sequences to the model's supported position range can be further divided into two lines:
\begin{itemize}
\item \textbf{Position Reorganization}: This approach reorganizes, particularly reuses, position indices that appeared during training to handle inputs exceeding training length. This idea of reorganizing position indices was evident in the T5 model~\citep{raffel2020exploring}, which supports 512 tokens input but only contains 32 types of relative positions. For mainstream large language models using RoPE encoding, similar approaches can be applied. In SelfExtend~\citep{zhou2023dynaicl}, for each token, normal relative positions are maintained for the nearest $w$ tokens, while distant tokens are grouped. DCA~\citep{an2024training} follows a similar approach. ReRoPE~\citep{kexuefm-9708} makes relative positions beyond window $w$ increase at smaller intervals. Furthermore, String~\citep{an2024does} discovered that even within the model's supported position range, it performs better at shorter relative positions, thus utilizing well-trained positions more extensively.
\item \textbf{Position Interpolation}: Unlike reusing existing positions, position interpolation chooses to monotonically reduce all input token position indices to not exceed the maximum training value. The first proposed interpolation strategy was linear interpolation, which directly scales down each token's position index $m$ to $m/\alpha$ \citep{chen2023extending}. Under RoPE position encoding, this is equivalent to uniformly reducing the angle $\theta$. According to neural tangent kernel theory~\citep{jacot2018neural}, this approach might prevent the model from learning high-frequency features. To address this, NTK interpolation~\citep{peng2023yarn}, also known as ABF~\cite{xiong2023effective}, reduces the scaling ratio for high-frequency parts while increasing it for low-frequency parts. In practice, NTK interpolation directly adjusts the original $\theta_j=10000^{-2j/d}$ to $\theta'j=(10000\lambda)^{-2j/d}$, where $\lambda$ is typically chosen to be slightly larger than $s$. YaRN~\citep{peng2023yarn} discovered that using a ramp function to perform NTK interpolation at different ratios across dimensions could achieve better results. Building on YaRN, Resonance RoPE~\citep{wang2024resonance} further optimized RoPE's features using integer wavelengths. LongRoPE~\citep{ding2024longrope} directly uses evolution search to find optimal frequency scaling parameters for each dimension. Some work has explored assigning different scaling factors to different attention heads~\citep{zhang2024found} or layers~\citep{chen2023fortify}, or dynamically deciding the optimal scaling factor via extra modules~\citep{zhu2025psc}, to improve model performance on long context tasks. In addition to the extrapolation method based on RoPE, some researchers have also applied linear and NTK interpolation techniques to ALiBi \citep{al2023position,NtkAlibi2023}.
\end{itemize}

On the other hand, enabling models to support position ranges larger than the context window size is mainly achieved through two approaches:
\begin{itemize}
\item \textbf{Hierarchical Position Embedding}: Similar to a number system, this approach greatly increases the representable range by introducing hierarchy in position encoding. BiPE~\citep{he2024two} introduces a two-layer position encoding system responsible for modeling positions within and between segments. HiRoPE~\citep{zhang2024hirope} focuses on code scenarios, utilizing code's natural hierarchy by using RoPE's lower $d/2$ dimensions and higher $d/2$ dimensions to handle token-level and function-level distances respectively.
\item \textbf{Position Simulation}: Interestingly, a line of research explores the use of short training data to simulate long training data, effectively decoupling training length from inference length. RandPos~\citep{ruoss2023randomized} randomly samples a set of positions from a longer position sequence, sorts them in ascending order, and uses them as position indices for shorter input data. Encoders trained using this approach demonstrated superior length extrapolation capabilities. For LLMs, \citet{zhu2023pose} proposed PoSE, which divides the original context window into several blocks, ensuring continuous position indices within blocks while allowing jumps between blocks, thereby covering longer relative positions with a shorter context window. CREAM~\citep{wuefficient}, LongRecipe~\citep{hu2024longrecipe}, and SkipAlign~\citep{wu2024long} have made further improvements upon PoSE. Specifically, CREAM employs strategies such as Gaussian Sampling to optimize text block partitioning, enhancing both continuity and relativity of position indices. LongRecipe similarly optimizes PoSE's text block partitioning and introduces "Impactful Token Analysis" to select text content for padding within each block. SkipAlign determines block sizes and position index skip rates based on specific instruction-tuning requirements, achieving performance comparable to GPT-3.5-Turbo-16k on LongBench.
\end{itemize}

\subsection{Attention}
\begin{figure}[!htp]
    \centering
    \includegraphics[width=1.0\linewidth]{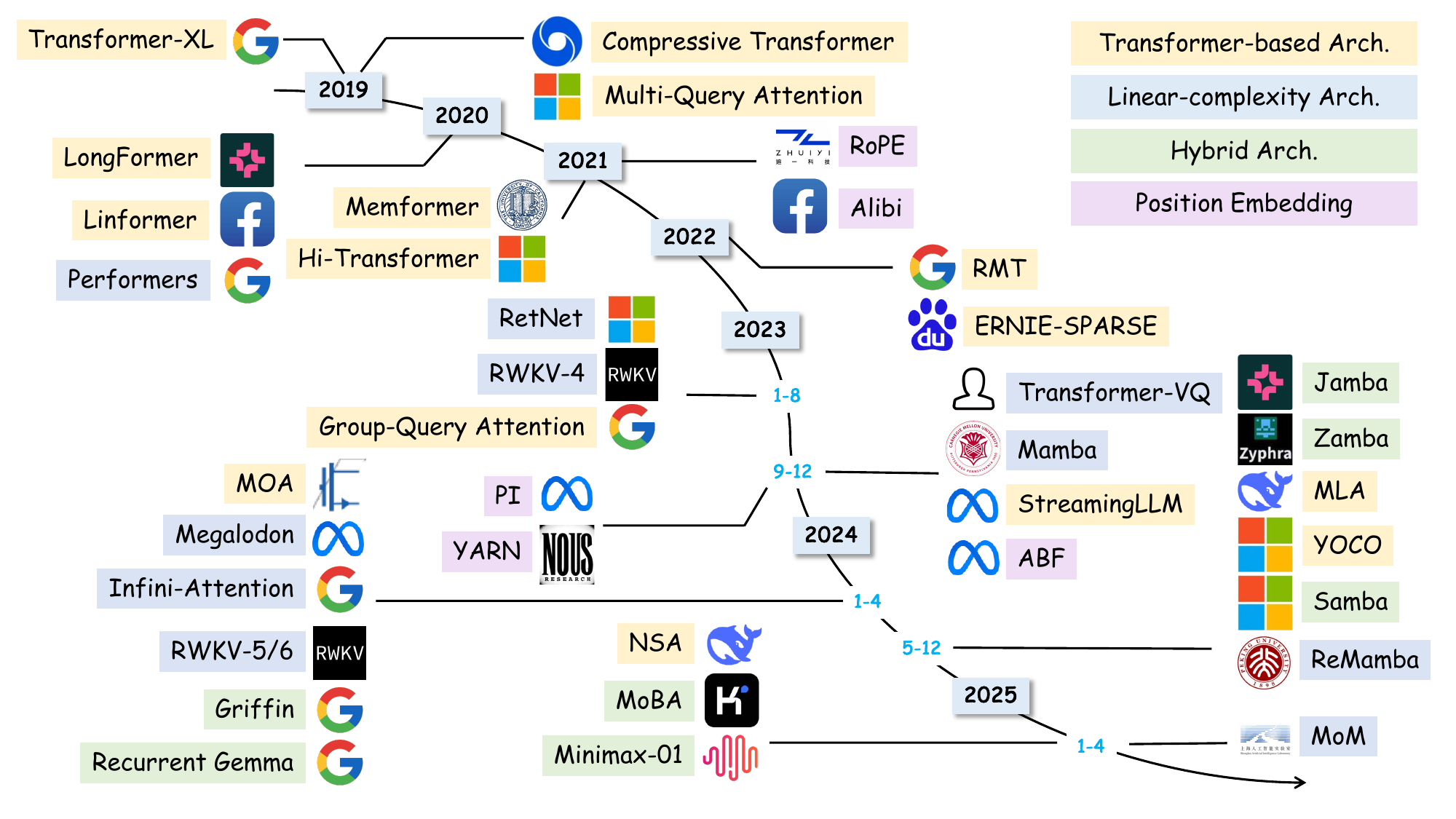}
    \caption{Illustration of different model architectures.}
    \label{fig:overview_model}
\end{figure}
\subsubsection{Transformer-Based Architecture}
\label{ssec:transformer_based_arch}

The predominant Large Language Models are based on the Transformer architecture~\cite{vaswani2017attention}. Specifically, this architecture consists of multiple stacked Transformer layers, where each layer is composed of Multi-Head Self-Attention (MHSA), Feed-Forward Networks (FFNs), and Layer Normalization (LN). Each layer takes the output from the preceding layer as its input and passes its own output to the subsequent layer. Classic configurations encompass encoder-only architectures such as BERT~\citep{Devlin2019BERTPO} and GLM~\citep{du-etal-2022-glm} for encoding models, decoder-only structures like GPT~\citep{10.5555/3495724.3495883} and LLaMA~\citep{touvron2023llama} for generative models, and encoder-decoder frameworks like T5 in sequence-to-sequence modeling. The core Self-attention~\citep{vaswani2017attention} mechanism  is as follows:

\begin{equation}
    \begin{aligned}
         Attention(Q, K, V)=softmax(\frac{QK^{\mathrm{T}}}{\sqrt{d_{k}}})V,
    \end{aligned}
    \label{eq:attention}
\end{equation}
where $Q \in \mathbb{R}^{n\times d_k}$, $K \in \mathbb{R}^{n\times d_k}$, $V \in \mathbb{R}^{n\times d_v}$. the time complexity of $QK^{T}$ is $O(n^{2})$.

Due to the inherent quadratic computational complexity associated with the MHSA component within the Transformer architecture, most Transformer-based models encounter limitations concerning length and efficiency, posing challenges for effective training and inference on long texts. Consequently, various approaches have emerged to enhance the Transformer structure, primarily focusing on improvements within the Attention mechanism and modifications to the overall architecture. The following sections of this chapter will delineate these methods including sparse attention, hierarchical attention, recurrent transformer, and efficiency-driven modifications.

\paragraph{Sparse Attention}
To address the quadratic complexity of attention in Transformer-based models and the challenge posed by the extremely large KV cache in long context scenarios, numerous prior works have explored sparse attention mechanisms, which reduce both computational and memory overhead, enabling faster processing and improved scalability for real-world applications. Sparse attention can be broadly categorized into training-based and training-free approaches, depending on whether it is applied during training or inference.

For training-based methods, the sparsity is mainly reflected from two perspectives,  head dimension (e.g. group query attention) and context window dimension (e.g. sliding window attention).
The most common function in head dimension sparse attention is group query attention(GQA)~\citep{ainslie2023gqatraininggeneralizedmultiquery}. 
Grouped-query attention organizes query heads into $G$ distinct groups, where each group utilizes a shared key head and value head. When $G=1$, GQA has only one group, one key and one value head, which is comparable to Multi-Query Attention (MQA).
When $G=H$ (the number of groups $G$ is equal to the number of heads $H$), the MQA is analogous to standard Multi-Head Attention (MHA). 
Weighted Grouped Query Attention~\citep{chinnakonduru2024weightedgroupedqueryattention} adds novel learnable parameters for every key and value head in the attention blocks to achieve aggregation among these key and value heads. Mixture of Attention (MoA)~\cite{fu2024moamixturesparseattention} automatically tailors distinct sparse attention configurations for different heads and layers, where the MoA constructs and navigates a search space of various attention patterns and their scaling rules relative to input sequence lengths.

As the context length grows, 
sparsity in the context window dimension can effectively reduce the extra computation caused by increasing the length and capturing valid information~\citep{lucas2024extraglobalattentiondesignation, wang2020linformerselfattentionlinearcomplexity,leviathan2024selectiveattentionimprovestransformer}. \citet{DBLP:journals/corr/abs-2004-05150} introduce the Longformer with an attention
mechanism that scales linearly with sequence
length to process documents of thousands of tokens.
Zebra~\citep{song2023zebraextendingcontextwindow} groups local-global attention layers into blocks during the training and inference phases. 
\citet{xiao2024efficientstreaminglanguagemodels} have found that keeping the KV of initial tokens will largely recover the performance of window attention, where these initial tokens lack meaningful semantics, but attract substantial attention scores. 
\citet{gao2024seerattentionlearningintrinsicsparse} adopts a fully learning-based approach to adaptively identify attention sparsity in LLMs and leverage the learned sparsity for efficient inference.
Recently, following the principles of Mixture of Experts (MoE),  the MoBA~\cite{lu2025mobamixtureblockattention} has been proposed, where the trainable block sparse attention and parameter-less gating mechanism are introduced.

However, most previous methods require training a model from scratch, limiting their usability as direct plugins for ready-to-use large language models. Some training-free sparse attention approaches are as follows:

\begin{itemize}
    \item \textbf{Static Strategy.} A variety of methods have been proposed to enhance efficiency while maintaining performance during inference. Window attention~\cite{DBLP:journals/corr/abs-2004-05150, child2019generatinglongsequencessparse, jiang2023mistral7b} maintains a fixed-size sliding window on the KV states of most recent tokens, ensuring efficiency but suffering from performance degradation once the initial tokens are evicted. StreamingLLM~\cite{xiao2024efficientstreaminglanguagemodels} mitigates this issue by identifying ``\textit{attention sinks}" where the initial tokens often receive disproportionately high attention scores, allowing LLMs trained on finite attention windows to handle infinite-length sequences without fine-tuning. LM-Infinite~\cite{han-etal-2024-lm} designs a $\Lambda$-shaped attention mask and a ceiling on attention distances and enables LLMs to generalize to extreme sequence lengths without any parameter updates.
    \item \textbf{Dynamic Strategy.} Unlike previous methods that rely on static, fixed-window strategies, later methods employ dynamic strategy to select and retain the most important tokens. H$_2$O~\cite{NEURIPS2023_6ceefa7b} retains only the ``\textit{Heavy Hitters}" tokens in the KV cache using accumulated attention scores. Similarly, Scissorhands~\cite{NEURIPS2023_a452a7c6} prioritizes important tokens based on the ``\textit{Persistence of Importance Hypothesis}". CORM~\cite{dai2024cormcacheoptimizationrecent} is a KV cache eviction policy to dynamically retains important key-value pairs for inference without model fine-tuning. SnapKV~\cite{li2024snapkvllmknowslooking} improves efficiency by utilizing attention scores to identify and cluster significant tokens. FastGen~\cite{ge2024model} recognizes several fundamental attention patterns and introduces an adaptive KV cache eviction policy. MInference~\cite{NEURIPS2024_5dfbe6f5} accelerates the prefilling stage using dynamic sparse attention with spatial aggregation patterns, allowing seamless integration into existing LLMs. Quest~\cite{pmlr-v235-tang24l} is a query-aware KV cache selection algorithm which tracks the minimum and maximum keys and values in KV cache pages.
    \item \textbf{Layer-Level Optimization.} Most of the aforementioned methods adopt the same strategy across layers to reduce the KV cache, ignoring the varying importance of information processing at different layers. PyramidKV~\cite{cai2024pyramidkvdynamickvcache} and PyramidInfer~\cite{yang-etal-2024-pyramidinfer} have discovered that LLMs aggregate information through a pyramid-shaped information funnel and dynamically adjust the KV cache size across different layers, allocating more KV cache to lower layers and less to higher layers. LazyLLM~\cite{fu2024lazyllmdynamictokenpruning} apply layer-wise token pruning in each generation step. DynamicKV~\cite{zhou2025dynamickvtaskawareadaptivekv} dynamically optimizes token retention by adjusting the number of tokens retained at each layer to better suit specific tasks. LightTransfer~\cite{zhang2025lighttransfer} minimizes inter-layer KV cache redundancies by selectively removing cache from identified lazy layers. TidalDecode~\cite{yang2024tidaldecodefastaccuratellm} accelerates LLM decoding through position persistent sparse attention.
    \item \textbf{Head-Level Optimization.} Previous works have not focused on the differences in attention patterns across different heads. RazorAttention~\cite{tang2025razorattention} and DuoAttention~\cite{xiao2024duoattentionefficientlongcontextllm} significantly improves the long context efficiency by dividing the attention heads into \textit{Retrieval Heads}~(which require full KV cache) and \textit{Non-retrieval Heads}~(which only need a fixed-size KV cache), which substantially reduces memory consumption and maintains accuracy while increasing both decoding and prefilling speeds. AdaKV~\cite{feng2025adakvoptimizingkvcache} adaptively allocates the KV cache budget across attention heads to optimize cache eviction strategies. HeadKV~\cite{fu2024headsmatterheadlevelkv} incorporates \textit{Retrieval Heads} identification to introduce a new adaptive budget allocation strategy, which globally allocates budgets individually for each attention head. LONGHEADS~\cite{lu2024longheadsmultiheadattentionsecretly4} splits the sequence context into chunks and each attention head focus on parts of chunks. When generating a particular token, LONGHEADS selects the $k$ blocks that are most relevant to the current token based on its query vector and chunk representation. 
    \item \textbf{Other Methods.} Despite these advancements, many approaches suffer from permanent information loss due to aggressive token pruning. Loki~\cite{NEURIPS2024_1e027da6} is a PCA-based sparse attention mechanism by leveraging the low-dimensionality of key vectors in the attention block, yielding inference speedup and maintaining efficacy.
\end{itemize}


\paragraph{Hierarchical Attention}
Various hierarchical mechanisms have been proposed to introduce a structured hierarchy into self-attention, 
which leverages high-level global information and low-level local attention for multi-scaled contextual receptive fields.
HAN~\citep{han} pioneers the use of a two-level attention mechanism. It first applies self-attention to word features to obtain a sentence representation, then employs self-attention on sentence-level features to generate document-level features. 
Hi-Transformer~\citep{hi-transformer} models documents in a hierarchical way,
which learns both the sentence representation and the document representation.
ERNIE-SPARSE\cite{ERNIE-SPARSE} leverages a Hierarchical Sparse Transformer to sequentially unify local and global information. HOMER~\citep{song2024hierarchical} processes extensive text by first dividing the input into smaller chunks and subsequently merging them in stages as they pass through the transformer's layers.
\paragraph{Recurrent Transformer}
Apart from the relatively high computational complexity, the self-attention mechanism integrates global and local information in the encoding of each token, rendering it less suitable for tasks such as long-text understanding. Based on this insight, some works strive to combine Transformer architecture with recurrent structures, enhancing the capture of long-term dependencies and improving the ability to understand long contexts~\citep{memorizingtransformer,trams,extensibleembedding}.

In the earlier attempts, Transformer-XL~\citep{TransformerXL} extends the context window beyond fixed lengths by employing segment-level recurrence and relative positional encoding. Memformer~\citep{Memformer} utilizes an external dynamic memory to encode and retrieve past information, attaining linear time complexity and constant memory space complexity when processing long sequences. Compressive Transformer~\citep{CompressiveTransformers} maps past hidden activations (memories) to a smaller set of compressed representations (compressed memories) for long-range sequence learning. Block-Recurrent Transformer~\cite{BlockRecurrentTransformers} applies a transformer layer in a recurrent manner along a sequence. During training, the recurrent cell operates on blocks of tokens rather than individual tokens and exploits parallel computation within a block to efficiently utilize accelerator hardware, exhibiting linear complexity with respect to the sequence length.
Unlike existing recurrent-based models that use special global tokens to store representations and place them at the beginning of the input sequence, RMT\cite{rmt} incorporates a memory mechanism through special memory tokens added to the input sequence, enabling the model to store and process both local and global information efficiently. SRformer\cite{segmentrecurrent} divides the input sequence into segments to compute segmented attention and adds recurrent attention to aggregate global information over segments. Infinite Attention\cite{infinitransformer} incorporates a compressive memory into the attention mechanism and incorporates both masked local attention and long-term linear attention mechanisms within a single Transformer block. ARMT\cite{armt} is based on transformer self-attention for local context and segment-level recurrence for storing task-specific information distributed over a long context.

\paragraph{Efficiency-Driven Modifications} 
By caching the previously computed key/value vectors, the decoders can reuse them for the current generation step. The key-value (KV) cache avoids encoding the history again for each token. Although this can significantly improve the inference efficiency, it imposes expensive memory overhead as the sequence length grows.
To alleviate the KV cache memory consumption, some efficiency-driven modifications have been proposed, including reducing KV-heads, such as Grouped-Query Attention (GQA), Multi-Query Attention (MQA) and  Multi-head Latent Attention (MLA) mechanism~\citep{liu2024deepseek}.
Note that the GQA and MQA has been discussed in Sparse Attention.
For MLA,
instead of directly reducing KV-heads, which is primarily considered a trade-off between performance and memory consumption, the MLA compresses the key and value into a latent vector to reduce the caching consumption and decompresses the key and value head for each query head when generating.
\subsubsection{Linear-Complexity Architecture}
\label{ssec:linear_comp_arch}

Linear complexity methods fall into two broad categories, including Mamba methods based on SSM architecture, and improvement methods based on Linear Attention. We also briefly discuss Test-Time Training at the end.

First, we focus on the State Space Model (SSM)~\cite{kalman1960new} and its variations. With the context length increasing rapidly, higher training and inference costs become a bottleneck. To speed up reasoning and capture long-term dependencies, the SSM models are gradually coming into people's view. SSM derived from modern control system theory, is a mathematical model that describes the behavior of a dynamic system using a set of first-order differential equations (continuous-time systems) or difference equations (discrete-time systems) to represent the evolution of the internal state of the system, while using another set of equations to describe the relationship between the state and the output of the system~\cite{voelker2018improving}.

\paragraph{SSM}
SSM is proposed based on classical Kalman filtering~\citet{Bang18, Gu2020HiPPORM}. It describes the state representation of a sequence at each time and predicts its next state based on the input. The SSM consists of two main parts - state and  observation. The state equation expresses the trend of the future state, while the observation equation predicts the current output based on the current state and input.

\begin{equation}
\begin{aligned}
x'(t)=\mathbf{A}x(t)+\mathbf{B}u(t) \\
y(t)=\mathbf{C}x(t) + \mathbf{D}u(t)
\label{eq:origin_ssm}
\end{aligned}
\end{equation}

$x(t) \in \mathbb{R}^{n}$ represents the state vector, $u(t) \in \mathbb{R}^{d_i}$ represents the input signal and $y(t) \in \mathbb{R}^{d_o}$ represents the output signal. $\mathbf{A} \in \mathbb{R}^{n \times n}$, $\mathbf{B} \in \mathbb{R}^{n \times d_i}$, $\mathbf{C} \in \mathbb{R}^{d_o \times n}$ and $\mathbf{D} \in \mathbb{R}^{d_o \times d_i}$ is the state matrix, input matrix, output matrix, and feed-forward matrix, which are parameters learned by gradient descent.  because the term $ \mathbf{D}u(t)$ can be viewed as a skip connection and is easy to compute. We can hypothesis $\mathbf{D}$ is a zero matrix. The rewritten formula is as follows:

\begin{equation}
\begin{aligned}
x'(t) &=\mathbf{A}x(t)+\mathbf{B}u(t) \\
y(t) &=\mathbf{C}x(t)
\label{eq:simple_ssm}
\end{aligned}
\end{equation}

The classical SSM model is used to process continuous signals, while discrete inputs are common in the field of NLP~\cite{gu2021combining}. Linear State Space Layer (LSSL) introduces continuous SSM  to obtain two discretized representations including the Recurrent Representation and the Convolutional Representation.
Mamba~\citep{gu2023mamba} leveraged the zero-order hold (ZOH) to discretize SSM. The Recurrent Representation of discretization is as follows:
\begin{equation}
\begin{aligned}
x_t &=\overline{\mathbf{A}}x_{t-1}+\overline{\mathbf{B}}u_t\\
y_t &=\mathbf{C}x_t
\label{eq:ssm}
\end{aligned}
\end{equation}

$\overline{\mathbf{A}}=\exp(\Delta \mathbf{A})$, $\overline{\mathbf{B}}=(\Delta \mathbf{A})^{-1}(\exp(\Delta \mathbf{A}) - \mathbf{I})\cdot \Delta \mathbf{B}$, $\Delta$ is step size that represents the resolution of the input. if we use $x$ instead of $u$  and $h$ instead of $x$, We can see that it has a similar formula as recurrent neural networks (RNNs),
\begin{equation}
\begin{aligned}
h_t&=\overline{\mathbf{A}}h_{t-1}+\overline{\mathbf{B}}x_t
y_t&=\mathbf{C}h_t
\label{eq:ssmaaa}
\end{aligned}
\end{equation}
Similar to RNN, the recurrent representation can be performed for efficient inference, but it cannot be used for parallel training. Therefore, LSSL proposes the convolutional representation for training.

Recently, the Structured State Space Sequence model (S4)~\citep{gu2022efficiently} is proposed based on SSM to modeling the  long sequences,
which can be computed more efficiently than previous methods while retaining their theoretical advantages.

\paragraph{Mamba and its Variants}
Prior work found that the SSM shows poor performance as it is a Linear Time-Invariant (LTI) system that is deeply connected to recurrence and convolutions. In other words, ($\Delta$, $\mathbf{A}$, $\mathbf{B}$, $\mathbf{C}$, $\overline{\mathbf{A}}$, $\overline{\mathbf{B}}$) in Equation \ref{eq:ssm} are fixed for all time-steps. However, LTI has fundamental limitations in modeling certain types of data. It loses the ability to efficiently select data in an input-dependent manner. ~\citet{gu2024mamba} propose Mamba to design a simple selection mechanism by parameterizing the SSM parameters. The details are as follows:

\begin{equation}
\begin{aligned}
\mathbf{B} \in \mathbb{R}^{n \times d} & \rightarrow \mathbf{B} \in \mathbb{R}^{L \times n} = s_{B}(x) \\
\mathbf{B} \in \mathbb{R}^{n \times d} & \rightarrow \mathbf{B} \in \mathbb{R}^{L \times n} = s_{B}(x) \\{C} \in \mathbb{R}^{n \times d} & \rightarrow \mathbf{C} \in \mathbb{R}^{L \times n} = s_{C}(x) \\
\Delta \in \mathbb{R}^{d} & \rightarrow \Delta \in \mathbb{R}^{L \times d} = \tau_{\Delta}(Parameter + s_{\Delta}(x))
\label{eq:Mamba}
\end{aligned}
\end{equation}
where $s_{B}$, $s_{C}$ and $s_{\Delta}$ are the Linear layer and $\tau_{\Delta}=softplus$.
Meanwhile, Mamba introduces the hardware-aware algorithm that computes the model recurrently with a parallel scan instead of convolution. Notably, it avoids IO access between different levels of the GPU memory hierarchy without modeling the expanded state. Finally, Mamba has 5x higher throughput than the  Transformer and has linear scaling over sequence length, which is friendly with long context capabilities.

Many methods have been proposed to optimize Mamba~\cite{chen2024stuffedmambastatecollapse, yuan2025remambaequipmambaeffective,song2025sparsified}. 
For example,
ReMamba~\cite{yuan2025remambaequipmambaeffective} employs selective compression and adaptation methods to compress and retain essential information, minimizing data degradation and reducing state space updates to alleviate information loss.
Similar to ReMamba, TAIPAN ~\citep{vannguyen2024taipanefficientexpressivestate} combines Mamba-2 with selective attention layers for enhancing performance and ensuring computational efficiency,
where the selective attention layer is used to filter out irrelevant information.  DeciMamba~\citep{benkish2024decimambaexploringlengthextrapolation} introduces a context-extension technique tailored for Mamba,
which leverages a concealed filtering mechanism, and allows the trained model to effectively extrapolate without further training.

\paragraph{Linear Attention}

Transformer has a significant advantage in long context tasks, and the current model has been extended to 128k or higher. But the time complexity and space complexity of the Transformer are both $O(n^2)$. As a result, when $n$ is significantly large, the computational burden of the Transformer model becomes challenging. Recently, considerable efforts have been made to lessen the computational demands of self-attention in Transformer~\citep{shen2021efficient, qin2022cosformerrethinkingsoftmaxattention, han2023flattentransformervisiontransformer, arora2024simplelinearattentionlanguage, xiong2021nystromformernystrombasedalgorithmapproximating}. {In this section, we will discuss mainstream works on linear attention as follows:}

First, 
\citet{katharopoulos2020transformersrnnsfastautoregressive} propose the linear transformer model by using a kernel-based formulation~\cite{radford2018improving} of self-attention and the associative property of matrix products to calculate the
self-attention weights,
which greatly reduces memory usage and scales linearly to the context length.
\citet{choromanski2022rethinkingattentionperformers} introduce the Performers,
which utilize the  fast attention via positive orthogonal random features and implement the kernelizable
attention mechanisms beyond softmax. 

RetNet~\cite{sun2023retentivenetworksuccessortransformer} includes three computation paradigms, i.e., parallel,
recurrent, and chunk-wise recurrent, and the chunk-wise recurrent paradigm achieves linear complexity for sequence modeling.
which achieves training parallelism, good performance, and low inference cost simultaneously.
\citet{munkhdalai2024leavecontextbehindefficient} propose the Infini-attention and include a compressive memory into the vanilla attention mechanism,
which uses  masked local attention and long-term linear attention modules in the Transformer block.
There are similar methods, such as
Lightning Attention-2~\citep{qin2024lightningattention2freelunch} employs a tiling strategy to distinctly manage intra-block and inter-block elements during linear attention computation,
where the MiniMax-01~\citep{minimax2025minimax01scalingfoundationmodels} model uses this attention module. It adopts standard attention mechanisms for intra-block processing while implementing linear attention kernel techniques for inter-block operations.
Transformer-VQ~\citep{lingle2024transformervqlineartimetransformersvector}  is a transformer decoder that performs dense self-attention computations in linear time relative to sequence length. This efficiency is achieved by integrating vector-quantized keys, localized positional biases, and a compressive cache designed for efficient attention processing.

\paragraph{RWKV Family} Unlike previous attempts with Recurrent Transformers, RWKV~\citep{peng2023rwkvreinventingrnnstransformer} introduces an enhanced linear attention mechanism, merging the computational efficiency of RNNs with the parallelism and expressive power of Transformers. This enables highly parallelized training and efficient inference, with its linear time complexity making it particularly suitable for long-sequence tasks.  The acronym "RWKV" represents four key elements:
\begin{itemize}
    \item \textbf{R (Receptance):} A gating vector that integrates historical information.
    \item \textbf{W (Weight):} A trainable decay factor applied across positions.
    \item \textbf{K (Key):} Functions similarly to the key vector in standard attention mechanisms.
    \item \textbf{V (Value):} Operates like the value vector in traditional attention systems.
\end{itemize}

RWKV-4 is the first publicly released version, following experimental iterations (RWKV-1, RWKV-2, and RWKV-3). The RWKV-4 model consists of multiple residual blocks, each featuring time-mixing and channel-mixing components.
Building on RWKV-4, RWKV-5 (Eagle) and RWKV-6 (Finch)~\citep{peng2024eaglefinchrwkvmatrixvalued} introduce further innovations. Eagle enhances expressive power by replacing vector-valued states with multi-head matrix-valued states and refining the learning decay strategy. It also reconfigures receptive states and adds gating mechanisms for improved performance.
 RWKV-6 (Finch) further boosts expressiveness and adaptability by integrating data-driven functions, such as parameterized linear interpolation, into the time-mixing and token-shifting modules. It also introduces low-rank adaptation functions, making weight matrices trainable and enabling context-sensitive optimization of decay vectors. These advancements maintain RNN-like inference efficiency while significantly enhancing the model's capabilities.

\paragraph{Test-Time Training} 
Another emerging approach to modeling long context is Test-Time Training (TTT)~\citep{sun2024learning}, which dynamically stores context information in the model's adaptable weights (referred to as fast weights~\citep{schlag2021linear}). The core idea is to frame the update of these adaptable weights as online gradient descent, as exemplified by systems like Titans~\citep{behrouz2024titans}, Atlas~\citep{behrouz2025atlas}, and TTT Done Right~\citep{zhang2025test}. When the update logic for adaptable weights is linear, TTT becomes equivalent to linear attention. Therefore, TTT can be viewed as a more expressive framework than linear attention, capable of capturing more complex dependencies through its nonlinear update mechanisms.

\subsubsection{Hybrid Architecture}
\label{ssec:hybrid_arch}
\citet{waleffe2024empirical} and \citet{parkcan} demonstrate that although the Mamba and Mamba-2 models perform well in language modeling, 
they fall short compared to Transformer models in long context tasks, such as in-context learning and long context retrieval.
\citet{chen2024stuffed} and \citet{waleffe2024empirical} find that adding a few standard Transformer layers back into the architecture enables the model to overcome these issues. 
The essence of hybrid architecture involves combining linear complexity attention with standard attention modules. This integration has developed into three main approaches: 
\begin{itemize}
    \item First, a layer-wise hybrid architecture where full attention and linear attention are mixed at the layer level. 
It's worth noting that models incorporating Sliding Window Attention (SWA) also fall into this category, even though early researchers might not have classified them as hybrid architectures. 
\item Second, a
prefilling-decoding hybrid architecture where different architectures are used for prefilling and decoding stages. This approach exclusively uses linear complexity attention layers during the prefilling phase, while the decoding stage implements a hybrid architecture that incorporates full attention mechanisms. The pioneering work in this direction is YOCO~\cite{sun2025you}.
\item Third, a head-wise hybrid architecture that leverages the multi-head principle of attention mechanisms. This method assigns specific attention heads to perform full attention while designating other heads to execute linear attention operations. Hymba~\cite{dong2024hymba} stands as the seminal work in this approach.

\end{itemize}

The following sections will explore these three approaches in detail.

\paragraph{Layer-Wise Hybrid Architecture}

The evolution of hybrid architecture in large language models represents a significant advancement in balancing performance and efficiency, particularly for handling long contexts. The layer-wise approach, combining linear complexity attention with standard attention modules, has emerged as the predominant strategy. Jamba~\cite{lieber2024jamba} pioneered this hybrid architecture at scale, effectively combining Transformer and Mamba layers with a MoE module, discovering that a 7:1 ratio (Mamba to Transformer layers) provided optimal balance. Jamba 1.5~\cite{team2024jamba} scaled this approach beyond 100B parameters, maintaining the same 7:1 ratio while requiring only 9GB for its KV cache with a 256K context length—compared to 80-88GB for similarly sized transformer models.
Google's RecurrentGemma~\cite{recurrentgemma} employed the Griffin architecture combining linear recurrences with local attention, using a mixture of mechanisms within each layer rather than interleaving different layer types. Meanwhile, Microsoft's Samba~\cite{ren2024sambasimplehybridstate} combined Mamba with SWA rather than full attention, maintaining linear complexity while providing the benefits of attention for local contexts. Comprehensive experimentation demonstrated that this hybrid approach outperformed both pure Mamba and pure attention models of similar size across standard benchmarks.
Zyphra's Zamba~\cite{glorioso2024zamba} innovated by incorporating a shared attention mechanism—utilizing a global shared attention block that appeared every few Mamba blocks but shared parameters across all instances, reducing memory requirements while maintaining strong performance. Zamba2~\cite{glorioso2024zamba2} expanded on these innovations in multiple size levels, switching from Mamba to Mamba-2 and using two alternating shared attention blocks with non-shared low-rank adapters.
MiniMax-01~\cite{minimax2025minimax01scalingfoundationmodels} became the first open-source hybrid model to achieve true state-of-the-art performance across comprehensive benchmarks, using a configuration where one transformer block with softmax attention followed every seven Transformer blocks with lightning attention~\citep{qin2024variouslengthsconstantspeed}. It matched the performance of leading commercial models while supporting context lengths up to 4 million tokens.

The sliding window approach represents another variant of hybrid attention. Google's Gemma~\cite{team2024gemma} employed a more balanced 1:1 ratio of full attention to SWA layers, emphasizing balanced performance across tasks rather than maximizing context length. Cohere's Command R-7B~\cite{c4ai-command-r7b-12-2024} implemented a hybrid approach with SWA, enabling efficient processing of longer contexts while maintaining strong benchmark performance. Research from Character.ai~\cite{character_ai} found that approximately 6:1 (linear complexity to full attention) yielded optimal results—aligning with findings from other hybrid models. This convergence on ratios of approximately 6:1 or 7:1 across multiple independent research efforts indicates a sweet spot in the design space, suggesting that hybrid architectures represent not just a compromise but potentially the optimal approach for next-generation language models efficiently handling diverse contexts.

The approaches discussed above primarily rely on standard full attention to capture global information. Recently, there have also been attempts~\citep{fang2025artificial} to compress global information into fixed-size representations, which preserve competitive performance while significantly lowering the compute and memory costs of the global part.

\paragraph{Prefilling-Decoding Hybrid Architecture}

The prefilling-decoding hybrid approach optimizes language model operation by applying different attention mechanisms based on the operational phase—using linear complexity attention during prefilling and implementing hybrid architectures during decoding. YOCO~\cite{sun2025you} pioneered this paradigm with a decoder-decoder structure that dramatically reduces memory requirements. The self-decoder processes input context using linear-complexity attention and produces a single global key-value cache, which is then reused by all layers of the cross-decoder. This design reduces KV cache memory requirements by approximately a factor equal to the number of model layers, 
which enables early exit during prefilling and maintains performance comparable to traditional Transformers.
GoldFinch~\cite{goldstein2024goldfinch} represents another innovation in this category, stacking a GOLD transformer on top of an enhanced Finch (RWKV-6) architecture. Its key contribution is the ``TokenCat'' mechanism, which generates a highly compressed key cache
in linear time and space. This approach yields remarkable efficiency gains, with cache size savings 756-2550 times smaller than traditional transformer caches.
Both approaches demonstrate that prefilling-decoding hybridization offers a powerful alternative to pure architectures, enabling dramatic efficiency improvements for long context inference without sacrificing model quality.

\paragraph{Head-Wise Hybrid Architecture}
The head-wise hybridization approach combines linear and attention-based mechanisms in parallel within the same layer, allowing different attention heads to process identical inputs using different mechanisms simultaneously.
Hymba~\cite{dong2024hymba} pioneered this approach with a hybrid-head architecture that integrates transformer attention mechanisms with state space models within each layer. Attention heads provide high-resolution recall capabilities while SSM heads enable efficient context summarization. This design can be interpreted as mimicking human cognition: attention heads function like snapshot memories storing detailed recollections, while SSM heads act as memories that retain core information while forgetting details.
 Samba~\cite{ren2024sambasimplehybridstate} combined the Mamba block and sliding window attention block together with an MLP layer. It combines local and global attention mechanisms. Local attention maintains sensitivity to near-neighbor information, while global attention allows the model to capture dependencies over long distances.

\section{Workflow Design}
\label{sec:workflow}

\begin{figure}[t]
    \centering
    \includegraphics[width=\linewidth]{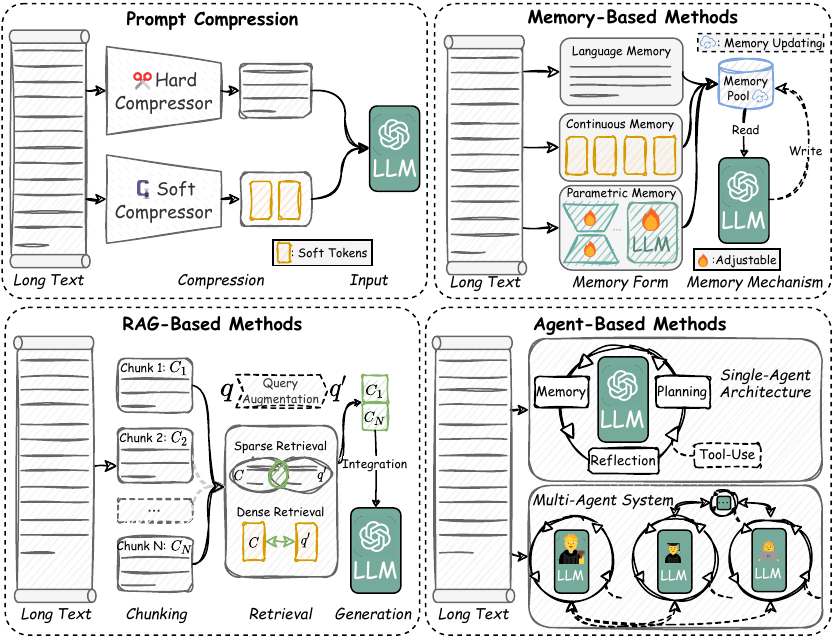}
    \caption{Design of LCLMs based on Workflow strategy.}
    \label{fig:workflow}
\end{figure}

Recent advancements in long context modeling often focus on modifying model parameters or altering the architecture to process long contexts, typically involving fine-tuning large language models or implementing complex internal mechanisms. In contrast, the methods presented in this section aim to enhance the long context processing capabilities of LLMs without altering their parameters, instead leveraging external components to augment the model's ability to handle long contexts. As shown in Figure \ref{fig:workflow}, the strategies discussed in this chapter are as follows:~(1)\textbf{Prompt Compression}~(Sec.~\ref{sec:prompt_compression}) reduces the size of the input context while retaining essential information.~(2)\textbf{Memory-Based Methods}~(Sec.~\ref{sec:memory_based_methods}) utilize an external memory module to store long contexts for efficient retrieval.~(3)\textbf{RAG-Based Methods}~(Sec.~\ref{sec:rag_based_methods}) extract or recall specific information from long contexts to reduce the context.~(4)\textbf{Agent-Based Methods}~(Sec.~\ref{sec:agent_based_methods}) leverage the LLM agent's memory, planning, and reflection capabilities to handle long contexts effectively.

\subsection{Prompt Compression}
\label{sec:prompt_compression}
As real-world tasks such as Chain-of-Thought (CoT), In-Context Learning (ICL) and Retrieval-Augmented Generation (RAG) become more complex, prompts for LLMs often grow longer to include detailed requirements, contextual information and  examples~\cite{li2024promptcompressionlargelanguage}. Nevertheless, lengthy prompts reduce inference speed, increase memory costs or API bills and degrade user experience. Prompt compression seeks to enhance LLM efficiency by reducing input complexity of LLMs. These techniques typically require an additional module to compress prompts while make minimal or no changes to the original LLMs' parameters, allowing for plug-and-play integration. Based on the form of the compressed prompt, prompt compression methods can be divided into two types: 1) \textbf{Hard Prompt Compression} maintains the use of natural language words or sub-words~(Sec.~\ref{sec:hard-prompt-compression}); 2) \textbf{Soft Prompt Compression} converts natural language into embedding representations~(Sec.~\ref{sec:soft-prompt-compression}).

\subsubsection{Hard Prompt Compression}
\label{sec:hard-prompt-compression}
Hard prompts are natural language prompts composed of tokens from the LLM's vocabulary, representing specific words or subwords. Hard prompt compression aims to streamline natural language prompts by reducing their length or complexity, while still maintaining their effectiveness in eliciting the desired response from the model.
This process typically includes two approaches:~(1) \textbf{selecting} relevant tokens of the prompt or (2) \textbf{rewriting} it for brevity and clarity.

\paragraph{Selecting}
SelectiveContext~\cite{li-etal-2023-compressing} enhances LLM inference efficiency by calculating token self-information with a small language model, grouping tokens into lexical units, and removing redundant content.
AdaComp~\cite{zhang2024adacompextractivecontextcompression} adaptively selects relevant documents by considering both query complexity and retrieval quality. CPC~\cite{liskavets2024promptcompressioncontextawaresentence} introduces a sentence-level compression technique leveraging a context-aware sentence encoder to rank sentences by their the embedding similarity to the query and removing less relevant sentences. TCRA-LLM~\cite{liu-etal-2023-tcra} proposes a token compression scheme with two complementary methods: summarization compression which uses a T5-based model to reduce token size, and semantic compression which eliminates words with lower semantic impact.

Besides, reinforcement learning (RL) has been applied to improve the efficiency and effectiveness of prompt compression methods. DynaICL~\citep{zhou2023dynaicl} enhances in-context learning efficiency by using reinforcement learning to fine-tune a meta-controller that dynamically adjusts the number of few-shot examples based on input query difficulty, balancing efficiency and performance. PCRL~\cite{Jung_2024} applies a discrete prompt compression method with reinforcement learning. TACO-RL~\cite{shandilya2024tacorltaskawareprompt} leverages task-specific reward signals to fine-tune a encoder-based compression model using on-policy RL, enabling effective compression while maintaining performance. 

In addition, the LLMLingua series aims to build a specialized language for LLMs through prompt compression, which accelerate model inference, reduce costs and improve downstream performance.
LLMLingua~\cite{jiang-etal-2023-llmlingua} leverages a smaller language model to calculate perplexity (PPL) and remove redundant tokens, featuring a budget controller, iterative prompt algorithm, and alignment techniques to achieve up to 20x prompt compression while preserving semantic integrity. LongLLMLingua~\cite{jiang-etal-2024-longllmlingua} proposes a question-aware coarse-to-fine compression method, a document reordering mechanism, dynamic compression ratios and a subsequence recovery strategy to improve LLMs’ ability to perceive and prioritize key information. LLMLingua-2~\cite{pan-etal-2024-llmlingua} introduces a task-agnostic prompt compression method, trained via data distillation from GPT-4 with a BERT-level encoder, achieving 3x-6x speed improvements over LLMLingua while preserving crucial information.

\paragraph{Rewriting}
An alternative approach involves rewriting prompts instead of selecting them. Nano-Capsulator~\cite{chuang-etal-2024-learning} encapsulate lengthy prompts into shorter ones while adhering to specific generation length constraints, maintaining performance through an explicit semantic-preserving objective with reward scoring. CompAct~\cite{yoon-etal-2024-compact} captures pivotal information from extensive documents by dynamically retaining essential contexts and incorporating information. FAVICOMP~\cite{jung2024familiarityawareevidencecompressionretrievalaugmented} is a decoding-time evidence compression approach that produces a refined evidence set more familiar to the target model, enhancing RAG performance while seamlessly integrating parametric knowledge.

\subsubsection{Soft Prompt Compression}
\label{sec:soft-prompt-compression}
Soft prompts~\cite{lester-etal-2021-power} present an innovative paradigm by eliminating the reliance on discrete tokens. Instead, they directly learn a series of continuous embeddings via backpropagation, which can be fed into the transformer without requiring a mapping to any actual language tokens. Such soft prompts can serve as an efficient alternative to conventional plain-text context, significantly reducing computational overhead during inference. Depending on whether the origin LLMs' parameters are updated, soft prompt compression can be broadly categorized into two types: 1) \textbf{LLM-fixed} methods and 2) \textbf{Gist token-based} methods.

\paragraph{LLM-fixed}
LLM-fixed methods focus on compressing prompts without updating the original LLMs' parameters, offering an efficient solution for managing input context length. Instead, they involve training additional modules that transform discrete tokens into continuous embeddings, facilitating efficient prompt compression. Contrastive conditioning~\cite{wingate-etal-2022-prompt} focuses on learning compact soft prompts to simulate the original natural language prompt by minimizing the Kullback-Leibler~(KL) divergence. However, contrastive conditioning incurs substantial computational overhead, as it requires retraining from scratch for each new incoming prompt, making it impractical for broad application.
ICAE~\cite{ge2024incontext} generates compact and informative memory slots to represent the original context, enabling LLM to encode more information within the same context length. This improves the model’s capability to handle long contexts and reduces computation and memory overheads during inference. 
500xCompressor~\cite{li2024500xcompressorgeneralizedpromptcompression} builds upon ICAE but utilizes KV values for compressed tokens instead of embeddings,achieving remarkable compression ratios. xRAG~\cite{cheng2024xrag} uses a frozen embedding model as the encoder and a trainable adapter between the encoder and decoder LLM. Through modality fusion, it integrates document embeddings into the LLM's representation space, eliminating the need for textual representations and achieving high compression rates.
UniICL~\cite{gao2024unifyingdemonstrationselectioncompression} compresses demonstrations into compressed features, which are then converted into compressed virtual tokens via a learnable projection layer. These virtual tokens replace original demonstrations to shorten input length and help select potential demonstrations. The queries and virtual tokens are then fed into the frozen LLM for response generation.

In summary, LLM-fixed methods offer diverse strategies for prompt compression by optimizing input representations while preserving the frozen parameters of the original LLM.
As an upper-bound reference, \citet{kuratov2025cramming} demonstrate near-lossless \(\sim 1500\times\) compression by optimizing a small set of soft `memory' vectors for a frozen LLM, clarifying the achievable limits of soft-prompt compression and pointing to directions for future encoders compression ratios.
In general, these methods significantly enhance computational and memory efficiency, enabling more effective utilization of LLMs for a broad range of tasks.

\paragraph{Gist Token-based}
Gist token-based methods aim to compress prompts by condensing the context into a small set of special tokens, referred to as ``\textit{gist tokens}"~\cite{NEURIPS2023_3d77c6dc}. These tokens replace the original context, enabling efficient caching and reuse while reducing computational costs.
However, such methods typically require modifying the parameters of the LLM. 
Gist~\cite{NEURIPS2023_3d77c6dc} introduces a method to train models for prompt compression without additional cost over standard instruction finetuning, which inserts gist tokens after the prompt and modifies Transformer attention masks to prevent tokens after the gist tokens from attending to those before them, allowing the model to learn both prompt compression and instruction following simultaneously. 
AutoCompressors~\cite{chevalier-etal-2023-adapting} process long documents by recursively generating gist tokens which are passed as soft prompts to all subsequent segments and result in more compact soft prompts. Activation Beacon~\cite{zhang2025long} serves as a plug-in for Transformer-based LLMs that enables effective, efficient, and flexible compression of long contexts, featuring a progressive compression workflow that distills the context into a small set of activations.

In summary, gist token-based methods provide a powerful framework for reducing input length by converting context into compact, reusable tokens. These approaches offer significant efficiency gains, though they typically involve updating the LLM’s parameters to enable effective compression and integration.

\subsection{Memory-Based Methods}
\label{sec:memory_based_methods}
Memory-based methods aim to utilize an external module (i.e., a memory module) to store long contexts. This approach effectively reduces the computational burden of directly relying on long contexts and mitigates potential information reduction associated with prompt compression. Generally, memory modules not only store essential historical information but also dynamically update, enabling efficient management of long context environments.

Memory augmentation has long been an actively explored topic. Even before the emergence of ChatGPT, numerous methods had been proposed and discussed. For example, 
MemPrompt~\citep{memprompt} and~\cite{towards-teachable-reasoning-systems} store user feedback on model responses in textual form, allowing for more personalized and accurate responses in subsequent interactions. 
Socratic Models~\citep{zeng2022socratic} enables LLMs (Large Language Models), ALMs (Audio-Language Models), and VLMs (Vision-Language Models) to share a unified memory unit that stores logs of numerous observations captured in egocentric videos. By extracting key information from these logs, LLMs can operate effectively within long context environments. 
Token Turing Machine~\citep{token-turing-machine} utilizes an external memory unit composed of continuous vectors. Through cross-attention mechanisms, the model can read information from the external environment for processing or write processed information back into the memory module. This enables the model to achieve long-horizon robotic control. 

Since the emergence of ChatGPT, LLMs have entered the era of agentic workflow. This paradigm involves equipping LLMs with abilities such as memorization, retrieval, planning, reflection, and tool-use, enabling them to adapt and respond contextually to complex, real-world tasks. 
In this section, we focus on how these LLMs or agents are augmented by memorization to understand long contexts or maintain consistency during long text generation. It is important to note that the discussions in ~\S\ref{sec:agent_based_methods} differ from this focus. 
Instead, the latter ones emphasize leveraging the agentic workflow specifically for enhancing LLMs’ long-text capabilities.

Memory-based methods for long-text language models can be categorized into three paradigms based on the form of memories: (1) continuous memory (latent vector representations), (2) language memory (textual content), and (3) parametric memory (model weights).

\paragraph{Language Memory.} 
Language Memory stores historical information as human-readable text fragments and typically reduces its length by retrieval and extraction. 
For example, 
Generative Agents~\citep{generative-agents} employs a ``memory stream'' with a triple scoring mechanism for item retrieval: (1) \textit{Recency}, which prioritizes recent interactions using an exponential decay function; (2) \textit{Importance}, where an LLM assigns a 1–10 score to each memory item based on its semantic significance; and (3) \textit{Relevance}, computed as the embedding similarity between the current query and memory item. This mechanism effectively manages complex contexts with extremely long interaction records in a sandbox environment.
Reflexion~\citep{reflexion} stores textual feedback from self-reflection in a memory bank. When similar errors occur in subsequent generations, the model retrieves historical reflection records for real-time correction. 
MemoryBank~\citep{DBLP:conf/aaai/ZhongGGYW24} incorporates the Ebbinghaus Forgetting Curve to gradually weaken infrequently accessed memories while reinforcing frequently used ones, simulating the formation of long-term human memory. 
AdaPlanner~\citep{adaplanner} and Voyager~\citep{voyager} address long-horizon task planning through a skill library mechanism, where successful textual action plans from previous interactions are archived as reusable templates. These stored plans can be dynamically synthesized into complex strategies for novel tasks. This capability can be seen as simplifying past memories, which are in the form of long texts, into skill items, and retrieving and synthesizing them into manageable and effective contexts for subsequent text generation tasks based on on-the-fly needs. 
RecurrentGPT~\citep{recurrentgpt} enhances the coherence for long-text story generation by combining long-term memory retrieval of relevant paragraphs with short-term memory maintained through iterative plot summarization, mimicking human cognitive processes for sustained narrative consistency.

\paragraph{Continuous Memory.} 
Continuous Memory encodes long context information into latent vector representations, enabling efficient retrieval of historical context without explicitly storing raw text. 
For instance, 
LongMem~\citep{longmem} splits lengthy texts into fixed-length segments and caches their key-value pairs from intermediate Transformer layers in an external memory bank. During inference, it retrieves top-k relevant historical key-value pairs through query-key attention operation and integrates them with current hidden states via a joint attention mechanism. A newly designed cross-network residual connection between the LLM backbone and a trainable Transformer-based SideNet conducts such memory retrieval and fusion operations. 
MemoryLLM~\citep{memoryllm} embeds trainable memory tokens within each Transformer layer as a fixed-size memory pool, implementing a self-update mechanism that selectively overwrites less frequently accessed memory information with new information by replacing only a portion of memory units per update cycle. 

\paragraph{Parametric Memory.} 
Parametric Memory internalizes long context information within the model's weights. 
For example, 
DSI (Differentiable Search Index)~\citep{dsi} reformulates document retrieval as a generation task, training the model to directly output document IDs for queries, which allowing for the memorization of document-query mappings in models' parameters. This approach eliminates the need for external storage and retrieval. 
Further, DSI++~\citep{dsiplusplus} addresses the catastrophic forgetting issues of DSI in continual learning settings by introducing a sharpness-aware loss function: $min_{\theta}max_{||\epsilon||_2\le\rho)}\mathcal{L}(\theta+\epsilon)$, where $\theta$ is the model parameters, and $\rho$ is a threshold. This loss encourages the model to converge to flatter minima, which are empirically shown to improve memory retention. Additionally, DSI++ employs a generative memory technique for memory replay, where a generator synthesizes pseudo-queries for previously indexed documents. These queries are mixed with new document data during continual training to mitigate forgetting and maintain updating over time. 
Generative Adapter~\citep{chen2024generative} dynamically generates lightweight adapter modules (\(\Delta_t\)) for previous context chunks (\([C_1, C_2, ..., C_t]\)) based on their hidden states (\([H_1, H_2, ..., H_t]\)) from the base LM, using an adapter generator (\(G\)) during test-time contextualization. Then, during inference, the generated adapter is integrated into the base LM for text generation. YORO~\citep{kobayashi-etal-2025-read} internalizes the database schema in input prompts into parametric knowledge of LMs via fine-tuning, significantly reduces the input token length by 66\%-98\% for Text-to-SQL task.

\paragraph{Pros \& Cons.} 
Memory-based methods provide critical theoretical advantages for long-text language models by addressing the inherent limitation of Transformer-based LMs imposed by their finite context windows, which either truncate long sequences or incur quadratic computational costs from attention mechanisms. 
By externalizing the storage of long context, these methods bypass the architectural bottleneck of vanilla Transformer while retaining critical historical information. 
The computational completeness of memory augmentation is further underscored by \citet{schuurmans2023memory}, who prove that vanilla Transformer-based LMs are computationally limited, which can be overcome through read-write memory augmentation. Their experiments demonstrate that a 540B-parameter LLM (\textit{i.e.}, Flan-U-PaLM-540B) with associative read-write memory can simulate a universal Turing machines. 
Similar conclusions are drawn from \cite{dehghani2018universal}. 
Nevertheless, memory-based methods face practical challenges including retrieval latency, inconsistencies between memorized and live contexts, and the complexity of memory updating, among others.

\subsection{RAG-Based Methods}
\label{sec:rag_based_methods}
In line with memory-based approaches for LLMs’ long-text capabilities, Retrieval Augmented Generation (RAG) aims to extract or recall specific information from the long texts to reduce the context. The retrieval sources can be training corpus, knowledge base, Internet, the provided long context, or the aforementioned external memories\footnote{Typically, the retrieved memories are so-called long-term memories~\citep{recurrentgpt}.}.

Specifically, leveraging RAG techniques for enhancing long-text capabilities of LLMs involves a three-stage workflow: 
(1) chunking to partition long context into manageable units, 
(2) retrieval to extract the most relevant information, 
and (3) generation to synthesize retrieved knowledge with intrinsic model understanding into coherent and contextually rich outputs.

\paragraph{Chunking.} 
Vanilla chunking units involve fixed token length~\citep{lewis2020retrievalaugmented}, sentences or paragraphs, delimiters such as ``\textbackslash n'' or ``\textbackslash t'', structural markers like Markdown headers or LaTeX sections, and semantically similar sentence neighbors\footnote{\url{https://github.com/FullStackRetrieval-com/RetrievalTutorials/blob/main/tutorials/LevelsOfTextSplitting/5_Levels_Of_Text_Splitting.ipynb}}. 
However, these approaches have several key issues as follows: 

\begin{enumerate}
    \item \textbf{Determining Optimal Chunk Size}: Overly large chunks may exceed input limits of the embedding models, while excessively small chunks can fragment context, making retrieval more challenging and potentially degrading the quality of retrieved information.
    \item \textbf{Maintaining Contextual Integrity}: Ensuring that splitting does not disrupt semantic meaning or omit critical information such as anaphoric references (e.g., ``it'', ``they'').
    \item \textbf{Balancing Efficiency and Effectiveness}: Chunking must be computationally efficient while preserving sufficient detail for downstream tasks.
    \item \textbf{Adapting to Data Formats}: Different data formats (e.g., text, code, or tables) may require specific chunking strategies.
\end{enumerate}

To mitigate these issues, 
for example, 
Late Chunking~\citep{günther2024late} first embeds the long document with an embedding model supporting long inputs and then conducts chunking on the embeddings. This approach ensures that each chunk embedding captures the whole contextual semantics. 
Sliding Window Chunking\footnote{\url{https://safjan.com/from-fixed-size-to-nlp-chunking-a-deep-dive-into-text-chunking-techniques/}} divides the text into overlapping chunks of a fixed size, ensuring that contextual dependencies across chunk boundaries are preserved. 
Contextual Retrieval~\citep{contextual-retrieval} leverages a long context LLM to augment the target chunk before embedding by taking the entire document as its context. The LLM generates an enriched version of the target chunk by integrating relevant contextual information from the document.

\paragraph{Retrieval.} 
The retrieval stage can be categorized into two approaches according to the retriever types: 
(1) Sparse Retriever which refers to the retriever that primarily uses sparse representations of text data such as TF-IDF\citep{drqa}, BM25\citep{toolformer} or Jaccard index\citep{zhang2023repocoder}; and (2) Dense Retriever which uses dense vectors with continuous values such as BERT\citep{devlin-etal-2019-bert} features. Among these two approaches, dense retriever is mostly used due to its ability to capture semantic similarities rather than relying on lexical matching. By encoding queries and documents (e.g., long context) into a shared latent space, dense retriever achieves higher retrieval accuracy and extends its functionalities such as instruction-following\citep{gritlm}, conversational search\citep{mo2024survey}, complex reasoning\citep{joshi2024reaper}, fine-grained noise filtering~\citep{zhang2025finefilter}, as well as scaling laws\citep{fang2024scaling} when combined with particular techniques. 
For example, 
BGE-M3\citep{chen2024bge} is a multilingual retrieval model that integrates sparse and dense retrieval, supporting an extended context window of up to 8192 tokens. 
ModernBERT\citep{warner2024smarter0} is a BERT-based model trained on 2 trillion tokens with an 8192-token context length, showcasing exceptional performance in natural language understanding (NLU) and long context retrieval tasks. 
REAPER~\citep{joshi2024reaper} enhances retrieval by incorporating complex reasoning, breaking down the retrieval query into a series of planned and chained steps.

Moreover, several techniques augment the queries to enhance the retrieval quality. 
For example, 
Query2Doc~\citep{wang2023query2doc0} generates pseudo-documents using few-shot prompting and expands the original query with these documents. 
HyDE~\citep{hyde} produces a hypothetical document through an instruction-following language model, encodes it with a contrastively learned encoder, and retrieves similar real documents from the corpus using the resulting embedding vector. 
Rewrite-Retrieve-Read~\citep{ma2023query} employs a trainable small language model as a rewriter, optimized via reinforcement learning with feedback from the LLM, to refine the query before retrieving contexts from a web search engine and processing them with the LLM.

\paragraph{Generation.} After the long context being reduced by chunking and retrieval, the following step is to integrate the reduced information into the LLMs for generation. Typically, without modeling training, the reduced information is fed into the LLMs by simple context concatenation~\citep{replug, realm, atlas}, or after further prompt compression~\citep{recomp}. For example, Fusion-in-Decoder~\citep{fusion-in-decoder} initially encodes all retrieved passages along with the corresponding questions into a sequence of soft features, which are then concatenated and input into the decoder-only LLM. The retrieved information can also be leveraged to adjust the decoding process. For example, kNN-LM~\citep{knn-lm} blends the model's prediction distribution with that of the nearest neighbors from the retrieval knowledge to adjust the generation on the output side. For certain specialized model architectures, such as Retro~\citep{retro}, trained cross-attention modules are employed to integrate the information.

\subsection{Agent-Based Methods}
\label{sec:agent_based_methods}

A LLM agent is an autonomous LLM entity that possesses a cognitive architecture consisting of several key components: perception capabilities to receive and process inputs, memory systems (both short-term and long-term) for information storage, action generation abilities for text output and tool manipulation, reflection mechanisms for self-evaluation, planning capabilities for goal setting and task decomposition, and reasoning abilities for logical thinking and decision-making~\citep{agent-survey-1, agent-survey-2}. In the context of long context LLMs, the LLM agent leverages its memory, planning, and reflection capabilities to process long texts effectively. For example, Generative Agent~\citep{generative-agents} demonstrates these capabilities through a comprehensive architecture where agents can extract and retrieve relevant information from their memory stream, generate periodic reflections to synthesize high-level insights from low-level observations, and create detailed plans that can be dynamically adjusted based on new information or interactions. This cognitive architecture enables agents to maintain coherent understanding and generate appropriate responses even when dealing with extensive context.

Agent-based approaches for enhancing LLMs' long-text capabilities can be broadly categorized into two main types: single-agent and multi-agent approaches. These two paradigms differ in their architectural design and operational mechanisms: (1) single-agent architectures, and (2) multi-agent systems. Single-agent approaches focus on developing comprehensive cognitive architectures within a single LLM agent entity. These methods emphasize building robust memory management systems, sophisticated planning mechanisms, and effective reflection processes. In contrast, multi-agent systems distribute the cognitive load across multiple specialized agents, each handling specific aspects of the long-text processing task. This paradigm emphasizes division of labor and collaborative problem-solving. Agents can engage in structured dialogues, debates, or collaborative analysis to understand or generate long texts more effectively.

\paragraph{Single-Agent Architectures.} 
For \textbf{long-text understanding}, 
ReadAgent~\citep{readagent} features a three-stage memory management approach powered by LLMs. The system first groups related content into coherent memory units, then distills these units into condensed summaries for efficient storage. When specific information is needed, it can intelligently navigate back to the source text to retrieve precise details for task completion. 
PEARL~\citep{pearl} employs a three-stage prompt framework to enhance reasoning about long documents: action mining, plan generation, and plan execution. When given a question about a long document, PEARL breaks it down into specific actions (such as summarizing, finding events, and identifying relationships), then executes these actions on the document to derive an answer. 
Self-Notes~\citep{self-notes} generates multiple QA notes interleaved with the input context and the question to handle the long context reasoning problems. 
MemWalker~\citep{memwalker} functions as an interactive reading system that breaks down long documents into a hierarchical structure of summaries. When given a query, it systematically explores this structured information, moving through different summary levels until it finds and collects the relevant details needed to provide an accurate response. 
GraphReader~\citep{graphreader} introduces an innovative approach that transforms lengthy documents into navigable graph structures. The system deploys an AI agent that intelligently traverses this graph representation. When presented with a query, the agent begins by conducting a detailed analysis to create a search strategy. Using specialized navigation tools, it systematically moves through the graph, examining both individual nodes and their connections in a hierarchical manner. The agent maintains a dynamic record of its discoveries and continuously evaluates its progress, adjusting its approach as needed until it collects enough data to construct a comprehensive response. 
RoleAgent~\citep{liu2024roleagent} employs a hierarchical memory system to distill long-text observations into shorter events, key points, and insights, enabling a coarse-to-fine information seeking during user-LLM interactions.

As for \textbf{long-text generation}, 
Re3~\citep{re3} follows an iterative five-stage process for long story generation. The process begins with the premise, where a given premise is provided to initiate the story. Based on this premise, the plan stage generates the setting, characters, and an outline using an LLM. Following this, the draft stage involves writing story continuations by prompting the model with both the plan and previously generated content. The process then enters an iterative cycle between the rewrite and edit stages: in the rewrite stage, the generated story continuations are reranked for their coherence with the plot and relevance to the premise, while in the edit stage, selected continuations are refined to ensure long-range factual consistency. This cycle of drafting, rewriting, and editing continues until the final story is generated, where the text flows naturally and aligns with the original premise. 
RecurrentGPT~\citep{recurrentgpt} builds two streams of memory flow for long-text generation, where the LLM agent is equipped with both short-term and long-term memory streams. During the generation of long text, the agent maintains and updates a short summary of the previously generated paragraphs as a part of context, which is also termed as short-term memory. Moreover, to integrate more detailed history paragraph information, the LLM agent is also enhanced with a retrieval mechanism to recall the most semantically relevant history paragraphs with the current generated paragraph as the query, which is also termed as long-term memory. Both memories are updated when new paragraph is generated.

\paragraph{Multi-Agent Systems.} 
Chain of Agent (CoA)~\citep{chain-of-agent} utilizes a multi-agent framework to process long context tasks. It divides the context into smaller segments, each handled by a worker agent. These worker agents sequentially communicate with a central manager agent, which synthesizes the contributions into a coherent output. This approach ensures efficient context processing by assigning each agent a short context and interleaving reading and reasoning. 
Similarly, LongAgent~\citep{longagent} employs a leader-agent model to process long texts. The leader understands the user’s intent and directs member agents to extract information from the document. An inter-agent communication mechanism resolves conflicts between agents’ responses, ensuring the leader gathers accurate data. This multi-agent approach enables efficient handling of long contexts while mitigating hallucination issues.

\section{Infrastructure}
\label{sec:infra}

In this section, we investigate the AI infrastructures that support LCLM training and inference, highlighting the key differences from those techniques typically used for general LLM.
The majority of the methods discussed are designed primarily for enhanced efficiency.
The others represent the perspective of the combination of engineering and algorithmic approaches, initially proposed to improve the model performance brought by the long context.

\begin{table*}[!htp]
\centering
\resizebox{\linewidth}{!}{
\begin{tabular}{c|cccc}
\toprule
\multirow{2}{*}{\textbf{Efficient  Strategies}} & \textbf{Computational} & \textbf{I/O} & \textbf{GPU HBM} & \textbf{Communication} \\
       & \textbf{Overhead }     & \textbf{Overhead} & \textbf{Memory}    & \textbf{Overhead}      \\
 \midrule
 \multicolumn{5}{c}{Training}\\
 \midrule
 Fundamental I/O optimization  & \textbf{--} & \Checkmark & \textbf{--} & \textbf{--} \\
 Data packing  & \textbf{--} & \Checkmark & \textbf{--} & \textbf{--} \\
 File systems & \textbf{--} & \Checkmark & \textbf{--} & \Checkmark \\
 Mixed-precision training & \Checkmark & \textbf{--} & \Checkmark & \Checkmark \\
 Low-precision training & \Checkmark & \textbf{--} & \Checkmark & \textbf{--} \\
 Optimized memory access & \Checkmark & \textbf{--} & \Checkmark & \textbf{--} \\
 Computation partition & \Checkmark & \textbf{--} & \Checkmark & \XSolidBrush \\
 Communication-computation overlapping & \Checkmark & \textbf{--} & \textbf{--} & \Checkmark \\
 \midrule
 \multicolumn{5}{c}{Inference}\\
 \midrule
 Quantization & \XSolidBrush& \Checkmark & \Checkmark & \textbf{--} \\
 Virtual Memory management  & \textbf{--} & \Checkmark & \Checkmark & \Checkmark \\
 Scheduling Strategies & \Checkmark & \textbf{--} & \textbf{--}& \textbf{--} \\
  Prefilling-Decoding Disaggregation & \Checkmark & \Checkmark & \textbf{--} & \Checkmark \\
 GPU-CPU Parallel Inference & \Checkmark & \XSolidBrush & \Checkmark & \textbf{--} \\
 Speculative Decoding & \XSolidBrush & \Checkmark & \XSolidBrush & \textbf{--} \\

 \bottomrule
\end{tabular}
}
\caption{Comparison of Optimization in AI Infrastructure. \Checkmark~means optimization in this aspect, \XSolidBrush~denotes a negative impact on this aspect, while \textbf{--} means no impact on this aspect or not involved.}
\label{tab:infra_stats}
\end{table*}

\subsection{Training of LCLMs}

With advancements in LLM algorithms, optimizations in training efficiency have also undergone significant innovation.
Given the scale of parameters and computational complexity, distributed training is the standard solution for LLMs.
Since the advent of ChatGPT, mainstream NVIDIA GPU compute resources have evolved through Volta, Turing, Ampere, Hopper, and Blackwell architectures~\footnote{https://www.nvidia.com/en-us/technologies/}, resulting in a hundred-fold increase in single-GPU compute capability (e.g., from 16.4 TFLOPS single precision for V100S to 2,250 TFLOPS for B200).
Maximizing this compute power in a distributed setting is the key challenge in LLM training.
Optimizations in computing, communication, memory management, and parallelization strategies have become standard practice and yield substantial performance gains.
These approaches are integrated into prominent training frameworks such as Megatron~\cite{shoeybi2019megatron}, DeepSpeed~\cite{rasley2020deepspeed}, and FSDP~\cite{zhao2023pytorch}, facilitating model development and training.
Numerous surveys \cite{liu2024understanding,duan2024efficient,brakel2024model} have documented these foundational techniques.

Nevertheless, long context lengths introduce further challenges.
Limited GPU memory, in particular, renders most common training optimizations ineffective.
Reducing batch size may enable execution, but inefficient computation and memory access bottlenecks severely degrade overall training efficiency.
High-performance long context training primarily explores parallel computation strategies and sophisticated communication-computation overlap to maximize hardware utilization and continuity.
Specifically, these methods address I/O, GPU resource constraints, and communication bottlenecks, as illustrated in Table~\ref{tab:infra_stats}.

\subsubsection{I/O Optimization}

\paragraph{Ultimate Foundational Optimization of I/O}
Training LCLMs inherently requires larger token batch sizes and significantly varying data lengths. 
Given constraints in memory, network bandwidth, and PCIe bandwidth, reading and transferring these large data volumes from memory to the GPU substantially slows down batch construction. 
Because I/O performance typically lags behind modern GPU compute performance, I/O becomes a major bottleneck in the training process. 
Strategic approaches, such as increasing the number of I/O threads and utilizing pinned memory, are commonly employed to mitigate this issue. 
However, the optimal hyperparameters for these strategic I/O optimizations require case-by-case tuning based on model size, context window length, and hardware configuration; otherwise, CPU cores or memory may become new bottlenecks.

\paragraph{Sophisticated Data Packing} Data packing concatenates samples into longer sequences.
To maximize effective data utilization in batches, packing requires the appropriate combinatorial arrangement and essential truncation~\cite{staniszewski2023structured,lu2024datasculpt}.
Packing would alter the data reading order and the actual training distribution, potentially affecting the training performance. 
Attention masks can differentiate packed samples~\cite{kundu2024enhancing}, ensuring isolation when necessary.
However, constructing non-causal attention masks introduces fragmented operations, and non-causal attention negatively impacts training efficiency.
Some methods~\cite{bai2024longalign,krell2021efficient} employ data sampling to ensure that sample lengths follow a predefined maximum. 
There exist other solutions which dynamically adjust context lengths.
For example, Hydraulis~\cite{li2024demystifying} avoids the usage of fixed context windows, but uses dynamic programming to address data sampling imbalance and data packing imbalance.
Data Decomposition~\cite{pouransari2024dataset} curates data to different windows using multiple bucket sizes.
Building upon these established data organization strategies, many recent pre-trained models have adopted a progressive length extension approach. This method has demonstrated significant performance improvements on commonly used long text benchmarks~\cite{zhao2024longskywork,prolong,gao2024prolong,yang2025qwen2}.

\paragraph{Distributed File Systems and Pre-fetching} In the LLM training process, data is read from disk/network to host cache, then transferred to GPU via PCIe.
In distributed training, data retrieval and distribution by CPUs are limited by PCIe bandwidth.
Pre-fetching with near-end data workers~\cite{zhao2023goldminer} or caching~\cite{dong2020eflops} can effectively overlap I/O and computation, even hiding and eliminating the I/O latency.
Upon the observation that SSD throughput and RDMA bandwidth were underutilized, 3FS~\cite{deepseek3fs} introduces a distributed, random-access approach, which enhances both performance and usability by leveraging the available resources more effectively.

\subsubsection{Optimizations on GPU Constraints and Memory Access}

\paragraph{Mixed-precision Training}
GPU memory is a significant bottleneck in long context LLM training, with parameters, gradients, optimizer states, and especially activations (proportional to sequence length) consuming substantial resources.
Reducing numerical precision~\cite{guan2024aptq} through low-precision or mixed-precision computation is a natural solution.
Mixed-precision training~\cite{micikevicius2017mixed} is a common memory optimization technique.
Typically, floating-point operations use FP32 (single-precision), allocating 8 exponent bits, and 23 mantissa bits.
In the memory-constrained scenarios, sacrificing some precision reduces storage requirements, e.g., FP16 requires half the storage.
However, FP16's maximum integer value of 65,536 can lead to the NaN exception or Inf exception owing to the numerical overflow. 
BF16, another half-precision format (requiring Ampere architecture or later), increases exponent bits, mitigating overflow at the cost of some precision.
FP16/BF16 are now standard for LLM training, with FP32 reserved~\cite{wang2024precision} for precision-sensitive operations (RoPE, LayerNorm, Softmax) and communication reductions (where BF16 should be avoided).

\paragraph{Quantization and Low-Precision Training} Recent quantized training methods incorporate quantized fine-tuning, using 8-bit floats (FP8) or integers (INT8), and preserve accuracy in the inference phase.
FP8 are currently usable in TransformerEngine and other derivative acceleration libraries \cite{deepgemm2025} on Hopper GPUs, primarily for less precision-sensitive matrix multiplications in Transformer layers.
It applies E4M3 (4-bit exponent, 3-bit mantissa) for forward and E5M2 for backward passes, potentially offering a 2x performance boost over BF16. 
INT8 integers are more widely applicable, forming the basis of most current quantization methods. 
Weight quantization is the most intuitive approach, mapping values to 256 discrete levels within a statistically or pre-defined range for each tensor. 
Quantization significantly reduces memory footprint, and with appropriate hardware and software support, dequantization overhead can be fused and minimized, significantly boosting performance.
However, reduced precision can irreversibly impact model performance.
Some studies observe that the error can be properly compensated at critical locations in the training process~\cite{peng2023fp8,yu2024collage,huang2024slim}.
Optimizer states are more sensitive to precision, requiring careful training with dynamic range scaling to align their distribution with the FP8 representation, reducing both memory footprint and quantization error~\cite{xi2024coat,dettmers2024qlora}.
Apart from the weights and optimizer state quantization, activation quantization is also attracting significant research.
Intuitively, quantizing activations leads to substantial information loss.
Fortunately, latest research discovers that this loss primarily attributes to the activation outliers, where suppressing these outliers during training can enable reliable activation quantization~\cite{li2023fptq,lin2024awq,xiao2023smoothquant}.

\paragraph{Optimized Memory Access and Blockwise Computation}
The quadratic time and memory complexity of the Transformer's multi-head attention mechanism presents significant computational and memory challenges, particularly with long input sequences. 
FlashAttention~\cite{dao2022flashattention} addresses this by leveraging the high bandwidth, but limited capacity, of GPU shared memory (e.g., 19TB/s bandwidth with a 20MB limit) for block-wise processing.
It employs a log-sum-exp trick for optimized softmax computation and caches key values and cumulative sums to minimize memory access.  FlashAttention demonstrates substantial performance gains in both standard and long context training scenarios.
Alongside these optimizations for standard attention, variants like sparse FlashAttention~\cite{pagliardini2023fast} have gained traction.
FlashAttention-v2~\cite{dao2024flashattention} refines the computation order of its predecessor, reducing non-matrix multiplications and optimizing warp scheduling within thread blocks to minimize shared memory access and maximize GPU parallelism.
Furthermore, leveraging the Transformer Engine introduced with the Hopper GPU architecture, FlashAttention-v3~\cite{shah2025flashattention} incorporates FP8 acceleration and three key techniques: asynchronous data loading and computation via a producer-consumer model, overlapping softmax and GEMM computations, and block-wise quantization.
This block-wise optimization paradigm facilitates further algorithmic and engineering co-optimization, as exemplified by NSA's~\cite{yuan2025native} hybrid approach combining token compression, selection, and a sliding window, 
and MoBA's~\cite{lu2025moba} block-wise attention mechanism, achieving a favorable balance between full and sparse attention using a MoE-like strategy.
Similar solutions have already been verified on MLA \cite{flashmla2025}.

\paragraph{Partition on Computation}
Distributed computing with sharding effectively mitigates the memory pressure imposed by long context windows.
Standard self-attention has a computational complexity of O(n$^2$), where n is the sequence length.
Ring attention~\cite{liu2023ring} reduces this to O(n) by restricting each token's attention to a fixed number of surrounding tokens, significantly decreasing computational and memory costs, and enabling the processing of longer sequences.
Hybrid approaches combining local and global attention mechanisms further improve performance by maintaining long-range dependency modeling while reducing computational overhead~\cite{liu2024world,brandon2023striped}.
More direct methods employ various parallelization strategies: 
Sequence Parallelism~\cite{korthikanti2023reducing} distributes model layers across devices, reducing individual device load but requiring substantial inter-device communication, and currently only shards Dropout and LayerNorm activations, leaving room for optimization.
Context Parallelism~\footnote{https://docs.nvidia.com/megatron-core/developer-guide/latest/api-guide/context\_parallel.html} divides the context window into segments processed in parallel, subsequently aggregating their representations. 
This effectively reduces memory requirements and increases training speed. 
Ulysses Parallelism~\cite{jacobs2023deepspeed}, integrated into the DeepSpeed framework, combines the advantages of both by sharding both model layers and the context window with an interleaved strategy, minimizing communication overhead while maximizing parallel efficiency, demonstrating significant advantages for extremely long context windows.

\subsubsection{Optimizations on Communication-Computation Overlapping}

Distributed LLM training requires inter-node communication for gradient aggregation and intermediate calculation results.
Performance discrepancies between computation and communication lead to hardware underutilization, exacerbated in long context training.
Since CPUs are heavily utilized for I/O in such scenarios, offloading computation to CPUs may not yield expected gains, and thus communication optimization is crucial.
Overlapping techniques enable parallel execution of forward computation with parameter gathering or backward gradient aggregation, which improves the overall efficiency.
Advanced bidirectional pipeline parallelism techniques effectively mitigate pipeline bubbles~\cite{deepseekai2024deepseekv3technicalreport}.
Gradient Accumulation (GA) is a crucial method for improving the overlap between computation and communication.
In LCLM training, high per-sample memory usage restricts local batch sizes on individual nodes.
GA addresses this by implementing mini-batches: performing forward passes on smaller batches and accumulating gradients across multiple mini-batches before a single, consolidated parameter update.
This allows for full parallelization of mini-batch forward passes with the preceding backward pass communication, effectively simulating large-batch training under memory constraints.
However, integrating GA with DeepSpeed-ZeRO requires careful consideration.
While ZeRO-1 necessitates only a single communication step during backward propagation due to its non-partitioned gradients, ZeRO-2, which partitions gradients across nodes, requires communication after each mini-batch forward pass to gather all gradient partitions, potentially hindering complete overlap.
Training engines often incorporate GA as a component, leveraging different CUDA streams to enable customized optimization~\cite{sourouri2014effective}.
In LLM training, strategically utilizing distinct CUDA streams within kernels allows for optimal overlapping strategies tailored to diverse model architectures and hardware configurations~\cite{chang2024flux,kim2024tccl}, effectively addressing discrepancies between communication and computation speeds.

\subsection{Inference of LCLMs}

Inference can be divided into two distinct phases: i) The prefill phase, where the input prompt is processed to generate the KV cache for large language models.
ii) The decoding phase, where the model utilizes the KV cache to generate subsequent tokens. Typically, the prefill phase is compute-bound, meaning that reducing computational overhead is key to accelerating this step. In contrast, the decoding phase is bandwidth-bound, requiring optimization of memory transfer during computation to improve efficiency.  In long context inference, the infrastructure faces four primary challenges:
First, during the prefill stage, the computation of attention incurs a quadratic time complexity with respect to the sequence length. Besides, during generation, the use of an extremely large KV cache imposes significant memory and I/O pressure. When performing multi-device or cross-device computations, there is also a communication overhead. The challenges addressed by various approaches are delineated in the table~\ref{tab:infra_stats}.

\begin{itemize}
\item \textit{Computational Overhead}.  During the prefill stage, the computation of KV cache incurs a quadratic time complexity with respect to the sequence length. Research efforts aim to accelerate the prefilling process~\citep{jiang2024minference,li2024scbench}, primarily mitigate the computational overhead. In contrast, the decoding phase is predominantly bandwidth-limited.  Optimizations for decoding, such as Quantization and Speculative Decoding, often strategically trade increased computational workload for reduced memory transfer requirements.  
\item \textit{I/O Overhead}. In Autoregressive Decoding, I/O operations become the primary bottleneck for speed. Each I/O operation requires transmitting both the model parameters and the KV cache into the computation unit to generate the next token. For long inputs, compressing the KV cache memory can significantly improve the speed. For long outputs, Spectulative Decoding and Model Parameters Quantization accelerate the process by reducing the memory of transmissions of model parameters. 
\item \textit{GPU HBM Memory}. 
Long-text tasks also impose significant HBM pressure on GPUs. As a result, some approaches offload the KV cache to the CPU and use retrieval techniques to alleviate the memory usage on HBM. Quantization applied to the model parameters or the KV cache, can also help reduce the pressure on HBM memory.
\item \textit{Communication Overhead}. In large-scale deployments, it is common to cache the precomputed KV cache of prompts, and when receiving the same prompt requests, the KV cache is transmitted between different machines via communication. Memory Management and Prefilling-Decoding Disaggregation methods can optimize this aspect.
\end{itemize}

In the following subsections, we will introduce several common techniques for enhancing inference performance through AI infrastructure improvements. These techniques include: quantization, memory management, PD Disaggregation, GPU-CPU Parallel Inference, and Speculative Decoding.

\subsubsection{Quantization}

In long context LLMs, processing extended input sequences leads to a significant increase in Key-Value (KV) cache size.  Furthermore, generating lengthy output sequences necessitates repeated transfers of both the KV cache and model parameters to computational units, exacerbating  bandwidth demands.  To mitigate these challenges and accelerate the decoding phase, quantization emerges as a crucial strategy by reducing the volume of data transferred.  Quantization for long context LLMs can be applied to either the KV cache alone or both the model parameters and KV cache.

One strategy focuses on quantizing solely the KV cache ~\citep{pmlr-v235-liu24bz,kang2024gearefficientkvcache,hooper2025kvquant,zhang2024kv,dong2024qaqqualityadaptivequantization}.  However, due to the prevalent lack of native hardware support for mixed-precision operations (e.g., FP16 x INT4) on mainstream architectures, efficient fusion computations often necessitate the development of specialized kernels.  Within KV cache quantization, common optimization techniques include: quantizing the Key and Value with different methods~\citep{pmlr-v235-liu24bz,hooper2025kvquant}, filtering outliers~\citep{hooper2025kvquant,zhao2024atom}, recording or adjusting the size of the channel~\citep{zhao2024atom,duanmuskvq,xiao2023smoothquant}. In contrast, quantizing both model weights and activations to a uniform low precision \citep{yao2022zeroquant,zhao2024atom,xiao2023smoothquant,sheng2023flexgen,liu2024unlocking,yue2024wkvquant} allows for the direct utilization of existing hardware-supported low-precision operations.

\subsubsection{Memory Management}

\paragraph{Virtual Memory Management}  In long context inference, the KV cache can grow extremely large, leading to memory fragmentation and inefficient usage. Virtual memory management techniques address these challenges by optimizing how KV cache memory is allocated and accessed, reducing waste and improving performance for long sequences. PagedAttention~\citep{kwon2023efficient} utilizes virtual memory to place the KV cache of tokens at the same position across different layers and heads within the same memory page. vTensor~\citep{xu2024vtensor} introduces a virtual memory abstraction that decouples computation from defragmentation. KV-Compress~\citep{rehg2024kv} extends PageAttention to support token-based KV compression. 
\paragraph{Scheduling Strategies}  When dealing with long context, especially in batched inference or scenarios with shared prefixes (like in conversational AI), efficient scheduling of KV cache and computations becomes critical.  Scheduling strategies tackle this by organizing data structures to facilitate deduplication and sharing, maximizing memory utilization and computational efficiency for long inputs. ChunkAttention~\citep{ye2024chunkattention} and MemServe~\citep{hu2024memserve} organize data structures to enable efficient cache deduplication and sharing of common prefixes, thereby improving both memory utilization and computational efficiency. SGLang~\citep{zheng2023efficiently} use RadixAttention for sharing common prefixes between batches.

\subsubsection{Prefilling-Decoding Disaggregated Architecture} 

Demanding both increased computation and memory for KV caches, long context LLMs inherently suffer from higher latency, resource consumption, and potential bottlenecks. To address these challenges, Prefilling-Decoding disaggregation decouples the computationally intensive prefill stage from the bandwidth-sensitive decode stage, assigning each to dedicated server pools optimized for their distinct resource demands~\citep{patel2024splitwise,qin2024mooncake}. By strategically allocating hardware, PD disaggregation boosts inference efficiency, demonstrably improving both Time to First Token (TTFT) for the prefill phase and Time Per Output Token (TPOT) for decoding phase~\citep{zhong2024distserve}.
Within this paradigm, a series of research efforts have emerged to optimize different aspects of the disaggregated architecture.
Distserve~\citep{zhong2024distserve} customizes resource allocation, parallelism strategies, deployment algorithms, and runtime scheduling optimizations for each stage. Splitwise~\citep{patel2024splitwise} investigates how to allocate machines within a cluster to handle the prefill phase and decoding phase effectively. Mooncake~\citep{qin2024mooncake} particularly excelling in long context scenarios and under heavy user loads, developed a prediction-based early rejection policy. CacheGen~\citep{liu2024cachegen} reduces the transmission time of precomputed KV caches across machines by adopting an optimized storage format. 
LoongServe~\citep{wu2024loongserve} proposes an elastic sequence parallelism method that dynamically adapts to different requests and phases, enhancing computation, communication, and GPU memory efficiency.
These studies collectively highlight the potential of PD disaggregation to accelerate LLM inference, paving the way for more scalable and cost-effective deployment of these powerful models.

\subsubsection{GPU-CPU Parallel Inference} GPU memory, designed for high-bandwidth access by computational units, is inherently limited and frequently inadequate to accommodate the ever-growing size of the KV cache, especially in long context scenarios. A cost-effective strategy to mitigate this constraint is offloading the KV cache to CPU memory, with the potential for further offloading to hard disk or network storage~\citep{xuanlei2024hetegen,jiang2024efficient}. While offloading reduces GPU memory pressure, it introduces a new bottleneck: the slow PCIe bus becomes a limiting factor when transferring the KV cache from CPU to the GPU for computation~\citep{yu2024twinpilots,he2024fastdecode}. To mitigate the issue of slow PCIe bandwidth, GPU-CPU Parallel Inference methods leverage concurrent CPU computation during PCIe transfers to either reduce the amount of data that needs to be transferred or optimize subsequent GPU computations, thereby improving overall efficiency. FlexGen~\citep{sheng2023flexgen} and PipeSwitch~\citep{bai2020pipeswitch} are techniques that attempt to overlap GPU-based computation of the current layer with the concurrent loading of the KV cache for the subsequent layer. FastDecode~\citep{he2024fastdecode} proposes computing attention scores directly on the CPU, leveraging its faster memory access to the KV cache relative to the GPU. Another methods employ CPU-GPU heterogeneous execution strategies to mitigate data transfer overhead by strategically performing computations on the CPU~\citep{xuanlei2024hetegen,yu2024twinpilots,park2024improving}.

\subsubsection{Speculative Decoding}
In long output scenarios, the decoding phase becomes a dominant factor in overall inference time. Greedy Decoding requires transmitting the model parameters for every new token computation. Speculative Decoding, on the other hand, accelerates this process by generating multiple potential tokens in a single pass and processing them together, thereby reducing the frequency of transferring model parameters from HBM to the computation units. Specifically, Speculative Decoding~\citep{leviathan2023fast} consists of a smaller draft model and a larger target model, the core idea is: (1) The draft model generates $\gamma$ candidate tokens. (2) The target model verifies the $\gamma$ tokens simultaneously. (3) The target model then either generates the first rejected token or adds a new token if all previous tokens are accepted.
Self-Speculative Decoding~\citep{hooper2023speed,zhang-etal-2024-draft,elhoushi-etal-2024-layerskip} leverage layer-skipping techniques to use the target model itself as the draft model, thereby reducing the overhead associated with maintaining a separate draft model. This approach sometimes allows the draft model and the final model to share the KV cache~\citep{elhoushi-etal-2024-layerskip}, further optimizing resource usage. MagicDec~\citep{chen2024magicdec} and TRIFORCE~\citep{suntriforce} use a draft model with a fixed KV budget using 
sparse attention. While the aforementioned methods effectively reduce or eliminate the overhead of a separate draft model, an alternative body of research demonstrates that significant performance gains can be achieved by carefully designing the architecture of a dedicated draft model.  To enhance draft model accuracy in speculative decoding, various architectural modifications have been proposed. One line of work involves adding specialized components to the draft model; for example, Medusa~\citep{cai2024medusa} employs multiple FFN heads, while Eagle~\citep{li2024eagle} incorporates features from the larger model's embeddings. Another promising direction explores alternative model architectures entirely, such as using a Mamba-based model for token prediction~\citep{choi2025mamba}, fully leveraging the efficiency of State Space Models(SSMs). More details about Speculative Decoding can be seen in this survey~\citep{xia2024unlocking}. 


\section{Evaluation}
\label{sec:evaluation}

\newcommand{\ColoredFiveStar}[1]{\textcolor{#1}{\raisebox{-2pt}{\FiveStar}}}
\newcommand{\ColoredBlock}[1]{\textcolor{#1}{$\blacksquare$}}

\newcommand{\boxedR}[2][]{%
    \begin{tikzpicture}[baseline=(char.base)]
        \node[draw,rectangle,rounded corners=2pt,scale=0.8,inner sep=2pt,#1] (char) {#2};
    \end{tikzpicture}}

\newcommand{\ColoredBoxR}[1]{%
    \IfStrEqCase{#1}{%
        {RQA}{\boxedR[fill=blue!10]{1} }%
        {RSum}{\boxedR[fill=green!10]{2} }%
        {RDRR}{\boxedR[fill=orange!10]{3} }%
        {RRAG}{\boxedR[fill=red!10]{4} }%
        {RICL}{\boxedR[fill=purple!10]{5} }%
        {RCode}{\boxedR[fill=cyan!10]{6} }%
        {S}{\boxedR[fill=yellow!10]{S} }%
    }%
}

\definecolor{Retrieval}{HTML}{72ac4d}
\definecolor{Aggregation}{HTML}{f2af38}
\definecolor{Reasoning}{HTML}{de5a32}

\definecolor{MCQ}{HTML}{F27970}
\definecolor{GEN}{HTML}{BB9727}
\definecolor{LM}{HTML}{54B345}
\definecolor{CLS}{HTML}{05B9E2}

This section introduces the evaluation of long context models. We generally divide the capacity of long context modeling into two aspects: processing long inputs and generating long outputs, namely \textit{Long Context Comprehension} and \textit{Long-Form Generation}.

For long context comprehension, we will first introduce its evaluation paradigm~(Sec.~\ref{sec:ueval_paradigm}), then summarize recent benchmarks~(Sec.~\ref{sec:ueval_summary_of_benchmarks}), and discuss how to design better benchmarks for long context comprehension~(Sec.~\ref{sec:ueval_benchmarks_discuss}). For long-form generation, we will first review its definition and introduce representative benchmarks~(Sec.~\ref{sec:long-form-evaluation-benchmark}). Then, we summarize the data sources~(Sec.~\ref{sec:long-form-data-source}) and present the commonly
used evaluation methods~(Sec.~\ref{sec:long-form-evaluation-methods}). After that, we discuss the challenges and trends in
long-form generation~(Sec.~\ref{sec:long-form-discussion}).

\subsection{Evaluating Long Context Comprehension}

\subsubsection{Evaluation Paradigm}
\label{sec:ueval_paradigm}

\begin{figure}[t]
    \centering
    \includegraphics[width=1.0\textwidth]{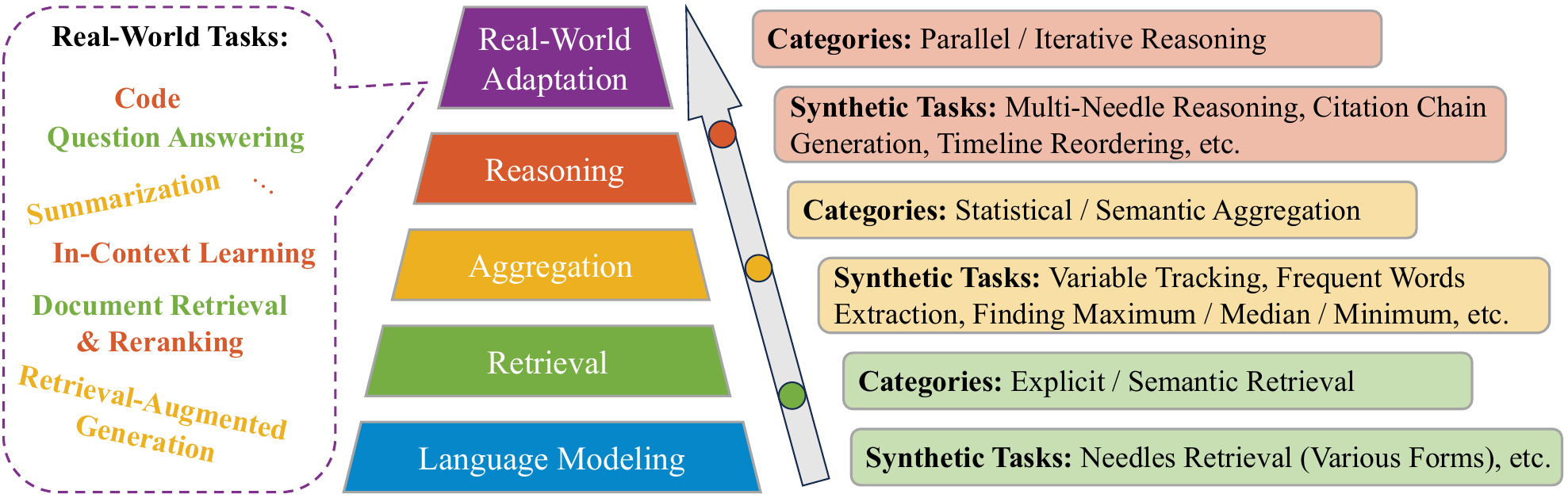}
    \caption{Evaluation paradigm for long context comprehension.}
    \label{fig:eval_paradigm}
\end{figure}

This section presents the evaluation paradigm for long context comprehension. As shown in Figure~\ref{fig:eval_paradigm}, we categorize the capabilities of LCLMs in processing long inputs into five hierarchical levels: \textit{language modeling}, \textit{retrieval}, \textit{aggregation}, \textit{reasoning}, and \textit{real-world adaptation}. 
At the foundation lies language modeling, which is the most basic capability for understanding the input text. 
Building upon this foundation, retrieval, aggregation, and reasoning constitute the core capabilities for long-text modeling. For each of these three capabilities, we provide a detailed taxonomy along with corresponding synthetic tasks specifically designed to evaluate each aspect.
At the highest level, real-world adaptation examines how well LCLMs can leverage their long context capabilities in practical scenarios. 
Specifically, we provide a detailed introduction to the most representative long context tasks, including question answering, summarization, document retrieval \& reranking, retrieval-augmented generation, many-shot in-context learning, coding, etc.
Performance on these tasks are among the most important indicators for assessing LCLMs' long-text understanding abilities.

\paragraph{Language Modeling} Language modeling is the most fundamental capability for long context comprehension. Typically, a low perplexity is an indicator that the model have good understanding of the given input document. In the realm of long context modeling, it is not only required that LCLMs have an overall low perplexity, but also that this perplexity decreases as the context window grows, indicating that the models can make more accurate predictions of target tokens by fully utilizing the rich context information. In practice, there are two common ways to examine the efficacy of LCLMs through the lens of perplexity. The first way is to inspect \textit{cumulative average negative log-likelihood (NLL)} as token position in long documents increases, as exemplified by Gemini-1.5~\citep{team2024gemini}. A decreasing trend in this cumulative NLL indicates that model can make better predictions of future tokens as the context window grows. An alternative way is to inspect PPLs calculated by the sliding window approach~\citep{press2021train,chen2023extending,zhu2023pose} under various window sizes. Specifically, the sliding window approach have a pre-defined window size $w$. The model first predicts the probability of the beginning $w$ tokens, then gradually slides the window forward to process the remaining tokens and calculate overall PPL. In this case, a decreasing PPL as the window size increases indicates that the model can effectively utilize long context information.

\paragraph{Retrieval} Retrieval requires LCLMs to identify and extract target information from one or multiple locations within long contexts. This capability underlies most long context tasks, as they typically require implicit retrieval of relevant content before subsequent processing. 
\begin{itemize}
    \item \textbf{Categories}\quad The retrieval capability comprises two distinct levels based on their complexity: \textit{Explicit Retrieval} and \textit{Semantic Retrieval}. Explicit retrieval refers to string matching based on given queries, where models must locate and extract matching content from the source text. Semantic retrieval, which is typically more challenging, requires models to extract semantically relevant content based on the semantic meaning of queries.
    \item \textbf{Synthetic Tasks}\quad \textit{Needle Retrieval (or Needle-in-a-Haystack, NIAH)} serves as a prototypical synthetic task to evaluate retrieval capabilities. This task inserts one or more needles into a long sequence and queries the model to retrieve the corresponding needles~\citep{mohtashami2023landmark,needleinhaystack,hsieh2024ruler,li2024needlebench,xiong2024artificial,zhu-etal-2024-longembed,li2023long,liu2024repoqa,bai2023longbench,roberts2024needle}. The specific instantiation of \textit{needles} and \textit{documents} takes on various forms. The needles can be n-digit numbers, specific sentences, UUIDs, dictionaries, functions, passages, etc. while documents can be repeated noisy sentences, meaningful content extracted from essays, code repositories, or other sources. In line with capability categorization, the needle retrieval task includes explicit and semantic variants. In \textit{Explicit (or Literal) Needle Retrieval}, models simply need to perform exact matching of needles, such as locating the completion of a partial sentence within a long text~\citep{mohtashami2023landmark,needleinhaystack,hsieh2024ruler,li2024needlebench,li2023long}. \textit{Semantic Needle Retrieval}, however, requires models to identify content based on semantic correspondence~\citep{modarressi2025nolima}, such as retrieving paragraphs based on their abstracts~\citep{bai2023longbench,zhu-etal-2024-longembed} or functions based on their behavioral descriptions~\citep{liu2024repoqa}. Notably, while these needle retrieval tasks primarily assess retrieval capabilities, they can be highly challenging and thus serve as crucial evaluation tasks for long context modeling abilities.

\end{itemize}

\paragraph{Aggregation} Aggregation refers to the model's ability to integrate information from multiple locations or even globally across long contexts. Building upon retrieval capabilities, this competency encompasses two key aspects: first, it requires the model to process the text piece-by-piece across multiple locations or even the entire context; second, it demands the model to recognize meaningful connections between these pieces and synthesize them into coherent higher-level representations.

\begin{itemize}
    \item \textbf{Categories}\quad Based on the nature of aggregation targets, Aggregation can be further divided into two types: \textit{Statistical Aggregation} and \textit{Semantic Aggregation}. Statistical Aggregation requires models to perform quantitative analysis over long contexts, while Semantic Aggregation focuses on combining and synthesizing semantic information from different parts of the context.
    \item \textbf{Synthetic Tasks}\quad A variety of synthetic statistical tasks have been proposed to evaluate Statistical Aggregation capabilities~\citep{song2024counting,hsieh2024ruler,zhang2024bench}. These tasks include tracing variable states, extracting frequent patterns, computing descriptive statistics (e.g., maximum, median, mode), and performing other numerical operations across extended sequences. For Semantic Aggregation, SummHay~\citep{laban2024SummHay} requires LCLMs to process the synthesized Haystack of documents and generate a summary. Additionally, some synthetic tasks specifically focus on solely evaluating the model's piece-by-piece processing capability across long context, such as stacked news labeling~\citep{qiu2024clongeval} and stacked typo detection~\citep{qiu2024clongeval}.
\end{itemize}

\paragraph{Reasoning} Long context reasoning refers to LCLMs' ability to perform logical inference over information distributed across long contexts. While both reasoning and aggregation involve identifying and processing multiple pieces of information, reasoning further emphasizes the logical deduction and inference process rather than mere information collection and summarization. 
\begin{itemize}
    \item \textbf{Categories}\quad In the realm of long context comprehension, Reasoning can be further categorized into \textit{Parallel Reasoning} and \textit{Iterative Reasoning}. Parallel Reasoning involves gathering all relevant information first before conducting the reasoning process, while Iterative Reasoning requires a step-by-step approach, where each reasoning step informs the next information gathering target, forming a continuous cycle until reaching the final conclusion.
    \item \textbf{Synthetic Tasks}\quad In this aspect, \textit{multi-needle reasoning} serves as a prototypical synthetic task that emphasizes both parallel reasoning and iterative reasoning capabilities. It requires models to reason across multiple \textit{needles} distributed throughout a long document~\citep{li2024needlebench,kuratov2024babilong}. These needles are typically logically related facts extracted from existing reasoning benchmarks. For example, the BABILong benchmark~\citep{kuratov2024babilong} utilizes samples from the bAbI dataset~\citep{weston2015towards} as needles, while NeedleBench~\citep{li2024needlebench} opts for the R4C dataset~\citep{inoue2020r4c}. 
\end{itemize}
\paragraph{Real-World Adaptation} Real-world adaptation represents the highest level in the capability hierarchy for long context comprehension, where models must effectively integrate and apply their fundamental capabilities (language modeling) and core capabilities (retrieval, aggregation, and reasoning) to address practical challenges. Unlike synthetic tasks that evaluate specific capabilities in controlled settings, real-world tasks present complex scenarios that often require multiple capabilities working in concert. These tasks not only test models' individual capabilities but also their ability to appropriately combine and deploy these capabilities based on task requirements. Introductions to the most representative real-world tasks involving long context comprehension are provided below. For each task, we use stars to denote which of the three core capabilities - retrieval~\ColoredBlock{Retrieval}, aggregation~\ColoredBlock{Aggregation}, and reasoning~\ColoredBlock{Reasoning} - the task tends to rely more heavily on.

\begin{itemize}
    \item \textbf{Question Answering}~[\ColoredBlock{Retrieval} / \ColoredBlock{Aggregation} / \ColoredBlock{Reasoning}]\quad The Question Answering task requires models to provide accurate answers based on real-world long contexts and queries. These questions and contexts span diverse domains,  including literature (NarrativeQA~\citep{kovcisky2018narrativeqa}, QuALITY~\citep{pang2022quality}, LStQA~\citep{qiu2024clongeval}, NoCha~\citep{nocha-2024-karp-thai-et-al}), academic papers (Qasper~\citep{dasigi2021dataset}), encyclopedias (WikiQA~\citep{yang2015wikiqa}, HotpotQA~\citep{yang2018hotpotqa}, 2WikiMultihopQA~\citep{ho2020constructing}, DuReader~\citep{he2018dureader}, AltQA~\citep{pal2023giraffe}), conversation history (LCvMem~\citep{qiu2024clongeval}), financial reports~(DocFinQA~\citep{reddy2024docfinqa}), structured data~(Table QA~\citep{zhang2024tablellm}), hybrid scenarios (MultiFieldQA~\citep{bai2023longbench}), etc. Notably, 
    due to the diversity of questions and domains, different QA tasks may emphasize different aspects of long context capabilities - some requiring primarily information retrieval, others focusing on information aggregation across multiple parts, and yet others demanding complex reasoning over the aggregated information.

    \item \textbf{Summarization}~[\ColoredBlock{Aggregation}]\quad Summarizing key information from long sequences of text has long been a crucial research area in NLP. Throughout the development of this field, researchers have created diverse summarization datasets spanning nearly all sorts of relevant domains, including novels~(\citep{zhang2024bench}, LStSum~\citep{qiu2024clongeval}, SQuALITY~\citep{wang2022squality}), government reports~(GovReport~\citep{huang2021efficient}), meeting scripts~(QMSum~\citep{zhong2021qmsum}, VCSUM~\citep{wu2023vcsum}), news articles~(MultiNews~\citep{fabbri2019multi}), patents~(BigPatent~\citep{sharma2019bigpatent}), screenplays~(SummScreen~\citep{chen2022summscreen}), legal documents~(MultiLexSum~\citep{shen2022multi}), and more. Notably, long context summarization also stands as one of the primary application scenarios for long context models.
    \item \textbf{Document Retrieval and Reranking}~[\ColoredBlock{Retrieval} / \ColoredBlock{Reasoning}]\quad Document Retrieval and Reranking are crucial components of modern Information Retrieval (IR) systems, responsible for retrieving relevant documents from a pool of candidates and reordering them based on their relevance to the query, respectively. The advent of LCLMs has enabled a new paradigm where models directly process all candidate documents and produce their rankings in a generative manner~\citep{DBLP:conf/emnlp/0001YMWRCYR23,ma2024fine,wang2024large}. Conversely, retrieval and reranking performance can serve as indicators of these models' long context capabilities, with the former emphasizing the ability to retrieve relevant information and the later focusing on reasoning across different portions of long context~\citep{lee2024can,yen2024helmet}. For example, the HELMET benchmark includes a Reranking test where candidate documents are sampled from the MSMARCO~\citep{nguyen2016ms} retrieval dataset, and LCLMs are challenged with generating top-10 document IDs ranked by relevance. Beyond relevance-based reranking, the task encompasses broader forms of sequence organization, such as timeline ordering~\citep{wang2024leave,li2023loogle,zhang2023marathon} and paragraph order reconstruction~\citep{dong2023bamboo}.
    \item \textbf{Retrieval-Augmented Generation}~[\ColoredBlock{Aggregation}]\quad Retrieval-augmented generation enhances the accuracy and reliability of generative models by providing them with a corpus of facts retrieved from external sources. In long context scenarios, where the input corpus can extend up to 1M tokens~\citep{lee2024can}, relevant information becomes more dispersed and sparse, challenging LCLMs' ability to aggregate information across extended contexts. Existing evaluation approaches~\citep{lee2024can,yen2024helmet} primarily employ open-domain QA to assess LCLMs' performance in retrieval-augmented generation. These evaluations typically process open-domain QA datasets such as Natural Questions~\citep{kwiatkowski2019natural}, TriviaQA~\citep{joshi2017triviaqa}, and HotpotQA~\citep{yang2018hotpotqa} by concatenating their passages into a large corpus. The model input consists of this corpus along with queries targeting specific passages, requiring LCLMs to locate relevant information within the extended corpus to generate appropriate answers.
    \item \textbf{In-Context Learning}~[\ColoredBlock{Reasoning}]\quad With extended context window size, In-Context Learning (ICL) expands the scale of demonstration examples from dozens, as seen in traditional ICL frameworks, to hundreds or even thousands. This scenario challenges LCLMs to generate precise predictions based on this substantially larger set of examples\citep{li2024long,xu2024stress,lee2024can,bertsch2024context}. Recent empirical studies have revealed that existing LCLMs exhibit several limitations in this setting: they demonstrate significant performance degradation beyond certain context lengths, show susceptibility to example ordering effects, display recency bias by favoring predictions aligned with labels presented near the sequence terminus, etc.~\citep{li2024long,xu2024stress} These findings indicate that there remains a substantial journey ahead for LCLMs to fully unleash the potential of ICL.
    \item \textbf{Code Relevant Real-World Tasks}~[\ColoredBlock{Reasoning}]\quad Processing repository-level code represents a particularly compelling application scenario that necessitates LCLMs. Among various specific tasks, code completion has garnered significant attention due to its comprehensive reflection of a model's capability to process entire repositories~\citep{bogomolov2024long,liu2024repobench,zhang2023repocoder,jimenez2024swebench}. Additionally, several tasks address other potential challenges in repository-level code processing, including CI builds repair, commit message generation, bug localization, module summarization, execution simulation, code translation, etc.~\citep{bogomolov2024long,zhang2024bench,wang2024repotransbench}. Despite significant advances, the intrinsic complexity of repository-level code still poses substantial challenges for current LCLMs in addressing real-world code processing requirements, as demonstrated by their performance limitations on SWE-Bench~\citep{jimenez2024swebench}.
\end{itemize}

\subsubsection{Evaluation Benchmarks}
\label{sec:ueval_summary_of_benchmarks}

\begin{table*}[!htp]
    \centering
    \footnotesize
    \renewcommand{\arraystretch}{1.25}
    \setlength\tabcolsep{2pt}
    
    \resizebox{\textwidth}{!}{%
    \begin{tabular}{lclllc}
    \toprule
    \multirow{2.5}{*}{\textbf{Benchmark}} & \textbf{Support} & \multicolumn{1}{c}{\multirow{2.5}{*}{\textbf{Characteristics}}} & \textbf{Target} & \textbf{Eval} & \multirow{2.5}{*}{\textbf{Metrics}} \\ 
    & \textbf{Length} & & \textbf{Aspects} & \textbf{Setup} & \\\midrule
    \multicolumn{6}{c}{\textit{General Domain Synthetic Benchmarks}} \\ \midrule
    Ada-LEval~\citep{wang2024adaleval} & 128k & Adaptable length & \ColoredBlock{Aggregation} \ColoredBlock{Reasoning} & M G & Auto \\
    BABILong~\citep{kuratov2024babilong} & 10m & BAbI in long context & \ColoredBlock{Reasoning} & G & Auto \\
    DENIAHL~\citep{dai2024deniahl} & 4k & Different NIAHs & \ColoredBlock{Retrieval} & G & Auto \\
    HoloBench~\citep{maekawa2025holistic} & 64k & Database aggregation\&reasoning & \ColoredBlock{Aggregation} \ColoredBlock{Reasoning} & G & Auto \\
    LIFBENCH~\citep{wu2024lifbench} & 128k & Instruction-following\&stability, new metric & \ColoredBlock{Retrieval} \ColoredBlock{Aggregation} \ColoredBlock{Reasoning} & G & Auto \\
    
    LongIns~\citep{gavin2024longins} & 16k & Instruction-centric & \ColoredBlock{Retrieval} \ColoredBlock{Aggregation} \ColoredBlock{Reasoning} & G & Auto \\
    LongPiBench~\citep{tian2024distance} & 256k & Focusing on position bias & \ColoredBlock{Retrieval} \ColoredBlock{Aggregation} \ColoredBlock{Reasoning} & G & Auto \\
    LongRangeArena~\citep{tay2020longrangearenabenchmark} & 16k & Early attempts, multiple modalities & \ColoredBlock{Retrieval} \ColoredBlock{Aggregation} \ColoredBlock{Reasoning} & C & Auto \\
    LongReason~\citep{ling2025longreason} & 128k & Long Context reasoning & \ColoredBlock{Reasoning} & M & Auto \\
    mLongRR~\citep{agrawal2024evaluating} & 64k & Multi-lingual retrieval\&reasoning & \ColoredBlock{Retrieval} \ColoredBlock{Reasoning} & G & Auto \\
    M4LE~\citep{kwan2023m4le} & 8k & Bilingual, semi-realistic & \ColoredBlock{Retrieval} \ColoredBlock{Aggregation} \ColoredBlock{Reasoning} & M G & Auto \\
    Michelangelo~\citep{vodrahalli2024michelangelo} & 128k & Latent structure queries & \ColoredBlock{Aggregation} \ColoredBlock{Reasoning} & G & Auto \\
    MLNeedle~\citep{hengle2024multilingualneedlehaystackinvestigating} & 32k & Multilingual NIAH & \ColoredBlock{Retrieval} & G & Auto \\
    NeedleThreading~\citep{roberts2024needle} & 900k & Different NIAHs & \ColoredBlock{Retrieval} \ColoredBlock{Aggregation} & G & Auto \\
    NoLiMA~\citep{modarressi2025nolima} & 32k & Semantic NIAHs w/ minimal lexical overlap & \ColoredBlock{Retrieval} & G & Auto \\ 
    RULER~\citep{hsieh2024ruler} & 128k & Different NIAHs + other synthetic tasks & \ColoredBlock{Retrieval} \ColoredBlock{Aggregation} & G & Auto \\ 
    S3Eval~\citep{lei2024s3eval} & 80k & SQL-centric & \ColoredBlock{Retrieval} \ColoredBlock{Aggregation} \ColoredBlock{Reasoning} & G & Auto \\
    SummHay~\citep{laban2024SummHay} & 100k & Synthetic tasks for summarization & \ColoredBlock{Aggregation} & G & LLM \\
    \midrule
    \multicolumn{6}{c}{\textit{Specific Domain Synthetic Benchmarks}} \\ \midrule
    LongHealth~\citep{adams2024longhealth} & 8k & Medical, fictional patient cases & \ColoredBlock{Retrieval} \ColoredBlock{Aggregation} & M & Auto \\
    MathHay~\citep{wang2024mathhay} & 128k & Mathematical retrieval\&reasoning & \ColoredBlock{Retrieval} \ColoredBlock{Reasoning} & G & Hybrid \\
    RepoQA~\citep{liu2024repoqa} & 16k & Code-style NIAH & \ColoredBlock{Retrieval} & G & Auto \\
     \bottomrule
    \end{tabular}%
    }
    \caption{Overview of synthetic benchmarks for long context comprehension. \textit{Support Length} is the maximum token count of samples in the benchmark (k=$2^{10}$, m=$2^{20}$). \textit{Characteristics} describe the unique features of the benchmark. \textit{Target Aspects} indicate the core long context capabilities the benchmark aims to evaluate, including Retrieval \ColoredBlock{Retrieval}, Aggregation \ColoredBlock{Aggregation}, and Reasoning \ColoredBlock{Reasoning}. \textit{Eval Setup} denotes the evaluation format the benchmark adopts: M for multi-choice question answering, G for free-form generation, C for classification. \textit{Metrics} lists the evaluation metrics used in the benchmark, including automatic metrics (Auto), LLM-as-a-Judge approaches (LLM), and hybrid styles combining both (Hybrid).}
    \label{tab:synbench_overview}
\end{table*}

Following the evaluation paradigm established in Section~\ref{sec:ueval_paradigm}, we present a comprehensive overview of recent long context comprehension benchmarks. These benchmarks can be categorized into two groups: synthetic benchmarks (Table~\ref{tab:synbench_overview}) that are entirely constructed using heuristic rules, and real-world benchmarks (Table~\ref{tab:realbench_overview}) that primarily consist of real-world tasks with human annotations. For each benchmark, we present its supported length (maximum token count), characteristics, evaluation setup, and evaluation metrics. The evaluation setup encompasses four main formats: multiple-choice question answering with provided options, free-form generation, classification, and language modeling. The evaluation metrics include automatic metrics, LLM-as-a-Judge approaches, and hybrid methods combining both.
Additionally, for synthetic benchmarks, we annotate their primary focus on specific aspects of long context comprehension capabilities, as these artificial tasks are typically designed to test particular comprehension skills. For real-world benchmarks, we document their covered scenarios, including Question Answering, Summarization, Document Retrieval \& Reranking, Retrieval-Augmented Generation, In-Context Learning, Code-Relevant Tasks, and various synthetic tasks.
\begin{table*}[!htp]
    \centering
    \footnotesize
    \renewcommand{\arraystretch}{1.25}
    \setlength\tabcolsep{2pt}
    
    \resizebox{\textwidth}{!}{%
    \begin{tabular}{lclllc}
    \toprule
    \multirow{2.5}{*}{\textbf{Benchmark}} & \textbf{Support} & \multicolumn{1}{c}{\multirow{2.5}{*}{\textbf{Characteristics}}} & \multicolumn{1}{c}{\multirow{2.5}{*}{\textbf{Scenarios}}} & \textbf{Eval} & \multirow{2.5}{*}{\textbf{Metrics}} \\ 
    & \textbf{Length} & & & \textbf{Setup} & \\ \midrule
    \multicolumn{6}{c}{\textit{General Domain Real-World Benchmarks}} \\ \midrule
    BAMBOO~\citep{dong2023bamboo} & 16k & Multi-task & \ColoredBoxR{RQA} \ColoredBoxR{RCode} \ColoredBoxR{S} & M G L & Auto \\
    CLongEval~\citep{qiu2024clongeval} & 100k & Chinese, human-curated (partly) & \ColoredBoxR{RQA} \ColoredBoxR{RSum} \ColoredBoxR{S} & G & Auto \\
    DetectiveQA~\citep{xu2024detectiveqa} & 250k & Bilingual, human-curated, detective novels & \ColoredBoxR{RQA} & M & Auto \\
    ETHIC~\citep{lee2024ethic} & 100k & Requiring high information coverage & \ColoredBoxR{RQA} \ColoredBoxR{RSum} \ColoredBoxR{S} & G & Hybrid\\
    
    InfinityBench~\citep{zhang2024bench} & 100k & Bilingual, long average length & \ColoredBoxR{RQA} \ColoredBoxR{RSum} \ColoredBoxR{RCode} \ColoredBoxR{S} & G & Auto \\
    HELMET~\citep{yen2024helmet} & $\sim$128k & Application-centric, robust eval & \ColoredBoxR{RQA} \ColoredBoxR{RSum} \ColoredBoxR{RDRR} \ColoredBoxR{RRAG} \ColoredBoxR{RICL} \ColoredBoxR{S} & M G & Auto \\
    L-CiteEval~\citep{tang2024lciteeval} & $\sim$48k & Answer with citations (faithfulness) & \ColoredBoxR{RQA} \ColoredBoxR{RSum} \ColoredBoxR{S} & G & Auto \\
    L-Eval~\citep{an2023eval} & 200k & Diverse data, improved metrics & \ColoredBoxR{RQA} \ColoredBoxR{RSum} & G & Hybrid \\
    LIBRA~\citep{churin2024long} & 128k & Long benchmark for Russian analysis & \ColoredBoxR{RQA} \ColoredBoxR{S} & G & Auto \\
    LOFT~\citep{lee2024can} & 1m & Real tasks of extreme lengths & \ColoredBoxR{RDRR} \ColoredBoxR{RRAG} \ColoredBoxR{RICL} \ColoredBoxR{S} & M G & Auto \\
    Long2RAG~\citep{qi2024long2rag} & 32k & Long retrieval\&output + new metric & \ColoredBoxR{RRAG} & G & Auto \\
    LongBench~\citep{bai2023longbench} & 16k & Bilingual, application-centric & \ColoredBoxR{RQA} \ColoredBoxR{RSum} \ColoredBoxR{RICL} \ColoredBoxR{RCode} \ColoredBoxR{S} & G & Auto \\
    LongBench-v2~\citep{bai2024longbench2} & $\sim$2m & Challenging tasks, expert curation & \ColoredBoxR{RQA} \ColoredBoxR{RICL} \ColoredBoxR{RCode} & M & Auto \\
    LongBench-Cite~\citep{zhang2024longcite} & 70k & QA with citations (trustworthiness) & \ColoredBoxR{RQA} \ColoredBoxR{RSum} \ColoredBoxR{RCode} & G & Auto \\
    LongICLBench~\citep{li2024long} & 50k & Extreme-label classification & \ColoredBoxR{RICL} & M G & Auto \\
    LongMemEval~\citep{wu2024longmemeval} & 1.5m & Long-term chat memory & \ColoredBoxR{RQA} & G & Hybrid \\
    Loong~\citep{wang2024leave} & 250k & Bilingual, semi-realistic, human-curated & \ColoredBoxR{RQA} \ColoredBoxR{RDRR} \ColoredBoxR{S} & G & LLM \\
    LooGLE~\citep{li2023loogle} & $\sim$24k & Human curation, long dependency & \ColoredBoxR{RQA} \ColoredBoxR{RSum} \ColoredBoxR{S} & G & Hybrid \\
    LV-Eval~\citep{yuan2024lveval} & 256k & Adaptable length, less knowledge leakage & \ColoredBoxR{RQA} & G & Auto \\
    ManyICLBench~\citep{zou2024retrieval} & 128k & Extensive ICL test, new metric & \ColoredBoxR{RICL} & M G & Auto \\
    Marathon~\citep{zhang2023marathon} & $\sim$80k & Human curation, multi-choice QA & \ColoredBoxR{RQA} \ColoredBoxR{RDRR} \ColoredBoxR{S} & M & Auto \\
    
    NoCha~\citep{nocha-2024-karp-thai-et-al} & 336k & Global reasoning over entire novels & \ColoredBoxR{RQA} & G & Auto \\
    TCELongBench~\citep{zhang2024analyzing} & $\sim$12k & Temporal complex events & \ColoredBoxR{RQA} \ColoredBoxR{RDRR} & M G & Auto \\
    ZeroSCROLLS~\citep{shaham2023zeroscrolls} & 8k & Early attempts & \ColoredBoxR{RQA} \ColoredBoxR{RSum} \ColoredBoxR{S} & G & Auto \\
    \midrule
    \multicolumn{6}{c}{\textit{Specific Domain Real-World Benchmarks}} \\ \midrule
    DocFinQA~\citep{reddy2024docfinqa} & $\sim$200k & Finance, FinQA in long context & \ColoredBoxR{RQA} & G & Auto\\
    FinTextQA~\citep{chen2024fintextqa} & 30k & Finance & \ColoredBoxR{RQA} \ColoredBoxR{RRAG} & G & Hybrid \\
    LongCodeArena~\citep{bogomolov2024long} & 2m+ & Code-centric & \ColoredBoxR{RCode} & G & Auto \\
    MedOdyssey~\citep{DBLP:journals/corr/abs-2406-15019} & 200k & Medical, diverse tasks, new metrics & \ColoredBoxR{RQA} \ColoredBoxR{S} & G & Auto \\
    NEPAQuAD1.0~\citep{phan2024examining} & 600k & Environmental statements\&acts & \ColoredBoxR{RQA} & G & Hybrid \\
    \bottomrule
    \end{tabular}%
    }
    \caption{Overview of real-world benchmarks for long context comprehension. \textit{Support Length} is the maximum token count of samples in the benchmark. \textit{Characteristics} describe the unique features of the benchmark. \textit{Scenarios} indicate the real-world task types covered by each benchmark: Question Answering \protect \ColoredBoxR{RQA}, Summarization \protect \ColoredBoxR{RSum}, Document Retrieval \& Reranking \protect \ColoredBoxR{RDRR}, Retrieval-Augmented Generation \protect \ColoredBoxR{RRAG}, In-Context Learning \protect \ColoredBoxR{RICL}, Code relevant tasks \protect \ColoredBoxR{RCode}, and synthetic tasks \protect \ColoredBoxR{S}. \textit{Eval Setup} denotes the evaluation format the benchmark adopts: M for multi-choice question answering, G for free-form generation, L for language modeling. \textit{Metrics} lists the evaluation metrics used in the benchmark, including automatic metrics (Auto), LLM-as-a-Judge approaches (LLM), and hybrid styles combining both (Hybrid). }
    \label{tab:realbench_overview}
\end{table*}

Based on Table~\ref{tab:synbench_overview}, we make several observations about existing synthetic benchmarks. First, a significant portion of synthetic benchmarks consists of various NIAH task variants, featuring more complex forms or different domains, essentially evaluating the robustness of models' long context retrieval capabilities. Second, most synthetic benchmarks rely solely on automatic metrics such as accuracy and F1 scores, with only a few requiring LLM or hybrid evaluation approaches when assessing longer generated outputs like summaries or mathematical reasoning processes. Third, the controlled construction process of synthetic benchmarks enables systematic comparisons of specific factors such as language, position, and patterns under controlled variables.

Analysis of Table~\ref{tab:realbench_overview} reveals several patterns in real-world benchmarks. First, Question Answering emerges as the most prevalent scenario, which aligns with the inherent diversity of QA tasks, followed by summarization tasks. Second, some real-world benchmarks adopt multiple-choice QA formats to guide model generation, possibly due to the diverse nature of real-world task outputs where unconstrained generation might complicate evaluation. Third, interestingly, not all real-world benchmarks rely entirely on human annotations; many incorporate synthetic tasks, likely to ensure comprehensive evaluation coverage.

Furthermore, examining both synthetic and real-world benchmarks collectively reveals that medical, financial, and code-related domains particularly demand strong long context processing capabilities.

\subsubsection{Discussion}
\label{sec:ueval_benchmarks_discuss}

\paragraph{What Makes Good Benchmarks for Long Context Comprehension} From a high-level perspective, an effective long context comprehension benchmark should satisfy three key requirements: coverage of sample lengths matching models' context windows, evaluation of fundamental long context modeling capabilities, and assessment of downstream task performance. Given that current LCLMs typically feature context windows exceeding 128k tokens, benchmarks primarily focusing on documents under 32k tokens may inadequately assess the capabilities of mainstream LCLMs. For fundamental capabilities, including retrieval, aggregation and reasoning as discussed in Section~\ref{sec:ueval_paradigm}, the main challenges lie in the dispersion of key information and the difficulty of extraction~\citep{goldman2024really}, with more dispersed and semantically oriented (rather than literal) information posing greater challenges~\citep{modarressi2025nolima}. Although synthetic tasks are particularly well suited for evaluating these aspects, research indicates that excellence in synthetic tasks alone does not guarantee downstream competence. Therefore, comprehensive benchmarks must incorporate specific downstream long context tasks, particularly question answering, summarization, RAG, and in-context learning, which serve as a relatively complete proxy for real-world applications. A final notable trend in evaluation methodology is the increasing formulation of tasks such as multiple choice question answering, allowing direct performance assessment through accuracy measurements.

\subsection{Evaluating Long-Form Generation}

\begin{figure}
    \centering
    \includegraphics[width=\linewidth]{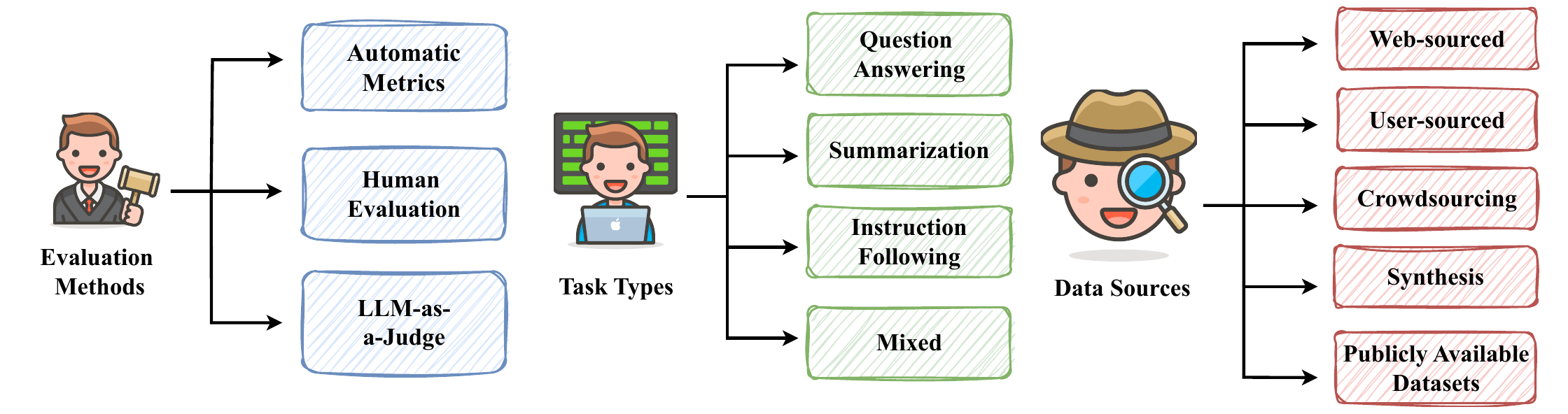}
    \caption{Overview of evaluating long-form generation.}
    \label{fig:long-form-gene}
\end{figure}

Beyond long context comprehension, where the inputs are lengthy, another common paradigm for long context modeling is long-form generation, where the outputs are lengthy. Long-form generation is a classic task in the field of NLP~\citep{wiseman2017challenges,bosselut2018discourse,cho2018towards}. In the realm of LLMs, the increasing practical demand for document-level text generation~\citep{yuan2022wordcraft,venkatraman2024collabstory,luo2024repoagent} and repository-level code completion~\citep{liu2024stall+,wang2024teaching} has brought growing attention to long-form generation, leading to the emergence of several evaluation benchmarks and improvement approaches~\citep{que2024hellobench,bai2024longwriter}. In this subsection, we first review the definition of the long-form generation and introduce representative benchmarks. Then, we summarize the data sources and present the commonly used evaluation methods. After that, we discuss the challenges and trends in long-form generation. The overview of long-form generation is illustrated in Figure ~\ref{fig:long-form-gene}.

\subsubsection{Evaluation Benchmark}
\label{sec:long-form-evaluation-benchmark}

long-form generation is generally defined as generating long, coherent, and contextually relevant text in response to a given input or instruction. To better understand the scope of the long-form generation, we provide the following clarification:

\noindent \textbf{First, the instruction must either explicitly or implicitly indicate the necessity for a long response.} 
Explicit requirements refer to instructions that clearly specify the expectation for a long response, either through a specified word count or a direct statement.
Implicit requirements, in contrast, do not specify a desired length but instead arise from tasks that inherently demand detailed responses. For instance, open-ended questions often require comprehensive examples or in-depth discussions to effectively convey a perspective, naturally leading to longer responses compared to close-ended questions.

\noindent \textbf{Second, the definition of what qualifies as a ``long'' response varies based on the nature of the task.}
In the context of a writing task, a response is typically considered long if it exceeds 1,000 words, whereas in a question-answering task, a response may be regarded as long if it surpasses 500 words. Accordingly, long-form generation can be defined as tasks that require responses substantially exceeding the average length associated with the specific task type.

\begin{table*}[!htp]
    \centering
    \footnotesize
    \renewcommand{\arraystretch}{1.25}
    \setlength\tabcolsep{10pt}
    
    \begin{tabular}{llllll}
    \toprule
    \textbf{Benchmark} & \textbf{Size} & \textbf{Task} & \textbf{Target Aspect} & \textbf{Source} & \textbf{Eval Method} \\ 
    \midrule
    ELI5~\citep{fan2019eli5} & 272K & QA & General & Web & Auto, Human \\
    MS-NLG~\citep{nguyen2016ms} & 183K & QA & General & User & Auto \\
    ExpertQA~\citep{malaviya2023expertqa} & 2K & QA & General & CrowdSrc & Auto \\
    ProxyQA~\citep{tan2024proxyqa} & 100 & QA & General & CrowdSrc & LLM \\
    LongGenBench-HKUST~\citep{liu2024longgenbench} & 16K & QA & General & PADs & Auto \\
    ASQA~\citep{stelmakh2022asqa} & 6K & QA & Ambiguous & PADs & Auto, Human \\
    QASA~\citep{lee2023qasa} & 2K & QA & Scientific & CrowdSrc & Auto \\
    CLAPNQ~\citep{rosenthal2024clapnq} & 5K & QA & RAG & PADs & Auto, Human \\
    Long2RAG~\citep{qi2024long2rag} & 280 & QA & RAG & PADs & Auto \\
    LFMedQA~\citep{hosseini2024benchmark} & 1K & QA & Medical & User & LLM \\
    MedLFQA~\citep{jeong2024olaph} & 5K & QA & Medical & PADs & Auto \\
    FActScore~\citep{min2023factscore} & 183 & QA & Factual & Web & LLM \\
    LongFact~\citep{wei2024long} & 1K & QA & Factual & Synthesis & LLM \\
    LTF-TEST~\citep{jeung2024large} & 12K & QA & Fairness & Synthesis & LLM \\
    AQUAMUSE~\citep{kulkarni2020aquamuse} & 6K & Summ & General & Web, PADs & Auto, Human \\
    Multi-News~\citep{fabbri2019multi} &  56K & Summ & General & Web & Auto, Human \\
    LCFO~\citep{costa2024lcfo} & 252 & Summ & General & PADs & Auto, LLM, Human \\
    LongForm-C~\citep{koksal2023longform} & 28K & IF & General & PADs & Auto \\
    Suri~\citep{pham2024suri} & 20K & IF & General & PADs & Auto, Human \\
    LongBench-Write~\citep{bai2024longwriter} & 120 & IF & Writing  & User & Auto, LLM \\
    LonGen~\citep{quan2024language} & 240 & IF & Writing & User & Auto, LLM \\
    LOT-OutGen~\citep{guan2022lot} & 2K & IF & Writing & Web & Auto, Human \\
    LongLaMP~\citep{kumar2024longlamp} & 63K & IF & Personalized & PADs & Auto \\
    DoLoMiTes~\citep{malaviya2024dolomites} & 2K & IF & Structured & Web & LLM \\
    LongGenBench-SUTD~\citep{wu2024longgenbench} & 400 & IF & Structured & Synthesis & Auto \\
    LongProc~\citep{ye2025longproc} & 2K & IF & Structured & PADs & Auto \\
    HelloBench~\citep{que2024hellobench} & 647 & Mixed & General & Web, PADs & LLM, Human\\
    FACTS Grounding~\citep{jacovi2025facts} & 2K & Mixed & Factual & CrowdSrc & LLM \\
    \bottomrule
    \end{tabular}
    \caption{Representative benchmarks for long-form generation. \textit{Web} refers to web-sourced data, meaning the data comes from the web. \textit{User} means the data comes from real users. \textit{CrowdSrc} is the abbreviation for crowdsourcing, indicating that the data comes from annotation teams. \textit{PADs} stands for Publicly Available Datasets, referring to the data sourced from publicly available datasets. \textit{Synthesis} means the data is synthetic. \textit{Auto} refers to automatic evaluation metrics. \textit{Human} represents human evaluation. \textit{LLM} refers to the evaluation based on LLMs, also known as LLM-as-a-Judge.}
    \label{tab:long-form-bench}
\end{table*}

Table ~\ref{tab:long-form-bench} lists representative benchmarks for long-form generation.
We summarize the task types of long-form generation into four categories: \textit{Question Answering (QA)}, \textit{Summarization (Summ)}, \textit{Instruction Following (IF)}, and \textit{Mixed}.

\paragraph{Question Answering (QA)} In long-form generation, QA mainly refers to long-form QA. \textit{ELI5}~\citep{fan2019eli5} is the first large-scale benchmark for long-form QA, where the task is to generate multi-sentence explanations in response to open-ended questions sourced from Reddit. Similarly, \textit{MS-NLG} is a subset of \textit{MS MARCO}~\citep{nguyen2016ms} that focuses on natural language generation. In addition to general long-form QA, \textit{ASQA}~\citep{stelmakh2022asqa} and \textit{QASA}~\citep{lee2023qasa}, which focus on ambiguous questions and scientific questions, respectively. In the realm of LLMs, \textit{ExpertQA}~\citep{malaviya2023expertqa} and \textit{ProxyQA}~\citep{tan2024proxyqa} are high-quality long-form QA benchmarks with questions crafted by experts across various fields, along with expert-verified golden answers. Retrieval Augmented Generation (\textit{RAG}) is an important application of LLMs, \textit{CLAPNQ}~\citep{rosenthal2024clapnq} and \textit{Long2RAG}~\citep{qi2024long2rag} are benchmarks designed to evaluate LLM-based RAG systems. Besides, generating long-form and well-supported responses is particularly essential in high-stakes domains such as medicine~\citep{hosseini2024benchmark,jeong2024olaph}. 
Although LLMs have demonstrated strong performance in many tasks, they often contain factual errors when generating long responses~\citep{wei2024long}. \citet{min2023factscore} propose a novel framework for factual evaluation, comprising three steps: decomposing the response into atomic facts, annotating each fact as supported, unsupported, or irrelevant, and computing the proportion of supported facts. 
In addition to concerns related to factuality, long-form responses are also susceptible to biases. To evaluate fairness in long-form generation, \citet{jeung2024large} propose an evaluation framework that covers 14 topics and 10 demographic axes.

\paragraph{Summarization (Summ)} Summarization is a classic task in long context modeling. As input documents become increasingly numerous and lengthy, the length of summaries has also grown significantly. Compared to early summarization benchmarks where summaries were around 50 words long~\citep{nallapati2016abstractive}, recent summarization benchmarks feature much longer summaries. As a result, summarization not only evaluates a model's ability to comprehend and condense long input documents but also tests its capability for generating long-form summaries. \textit{Multi-News}~\citep{fabbri2019multi} is the first large-scale multi-document news summarization benchmark, collected from \texttt{newser.com}. \textit{AQUAMUSE}~\citep{kulkarni2020aquamuse} proposes a scalable method to automatically mine summarizations from \textit{Natural Questions}~\citep{kwiatkowski2019natural} and \textit{Common Crawl}. \textit{LCFO}~\citep{costa2024lcfo} is a human-annotated benchmark for evaluating long context summarization and summary expansion, consisting of 252 long documents from diverse domains.

\paragraph{Instruction Following (IF)} 
Apart from using questions as inputs for LLMs, instructions and other forms of prompts can also serve as inputs. Such tasks prioritize the evaluation of a model's ability to follow instructions, encompassing both straightforward instruction adherence and more complex, constrained instruction-following.
To evaluate the ability of LLMs in structured problem-solving, \textit{DoLoMiTes}~\citep{malaviya2024dolomites} consists of 519 domain-specific long-form methodical tasks with 1,857 examples collected from experts across 25 fields. 
Similarly, \textit{LongGenBench-SUTD}\footnote{In fact, there are two benchmarks named LongGenBench~\citep{wu2024longgenbench, liu2024longgenbench}. To distinguish them, we add the suffix of the respective primary institution.}~\citep{wu2024longgenbench} and \textit{LongProc}~\citep{ye2025longproc} are designed to evaluate the capability of LLMs to generate high-quality long-form text while adhering to complex instructions.
Besides, \textit{LongLaMP}~\citep{kumar2024longlamp} is a benchmark for personalized generation, covering four tasks: personalized email completion, abstract generation, review writing, and topic writing. 
Recently, increasing attention has been given to aligning LLMs with long-form output instructions, such as creative writing and story generation. To enhance instruction tuning for long-form generation, \textit{LongForm-C}~\citep{koksal2023longform} gathers instruction-following examples via reverse instructions on C4 and English Wikipedia. \citet{bai2024longwriter} propose a long instruction-following dataset \textit{LongWriter-6K} aimed at extending the output length of existing models to over 10,000 words. 
Similarly, \textit{Suri}~\citep{pham2024suri} is a benchmark consisting of 20,000 long-form human-written texts, each paired with backtranslated instructions with multiple constraints. \citet{quan2024language} propose Self-Lengthen, an iterative training framework that utilizes LLMs' intrinsic knowledge without relying on auxiliary data. To evaluate its effectiveness, they introduce \textit{LongGen}, a benchmark for evaluating LLMs' long-form generation capabilities in Chinese and English across various tasks, with length-constrained user instructions and corresponding responses. 

\paragraph{Mixed} Some benchmarks include multiple task types to enable more comprehensive evaluation. \textit{Facts Grounding}~\citep{jacovi2025facts} evaluates LLMs' ability to generate factually accurate long-form text based on a given context and user requests, encompassing tasks QA, summarization, and document rewriting. \textit{HelloBench}~\citep{que2024hellobench} is a comprehensive benchmark for evaluating LLM's long-form generation capabilities, consisting of 647 samples across 38 subcategories and 5 tasks (QA, summarization, chat, completion, generation), sourced from real-world scenarios and publicly available datasets.

\subsubsection{Data Source}
\label{sec:long-form-data-source}

The data source of a benchmark determines its data quality. A reasonable, high-quality, and easily accessible data source can significantly enhance the overall quality of the benchmark. Data sources for long-form generation benchmarks can be categorized into five types:
\begin{itemize}
    \item \textbf{Web-Sourced Data}: Web contains abundant and vast amounts of text content. The advantage of web-sourced data is its richness and diversity. However, not all web content meets high-quality standards and typically necessitates processes such as data cleaning and deduplication. Among long-form generation benchmarks, ELI5~\citep{fan2019eli5} collects QA pairs from Reddit, FActScore~\citep{min2023factscore} gathers people biographies from Wikipedia, and LOT-OutGen~\citep{guan2022lot} collects human-written stories by crawling web pages.
    \item \textbf{User-Sourced Data}: The main characteristic of user-sourced data is its practicality, as it reflects real-world scenarios faced by users. For example, MS-NLG~\citep{nguyen2016ms} collects data from users' search logs on Bing. LongBench-Write\citep{bai2024longwriter}, LonGen~\citep{quan2024language}, and LFMedQA~\citep{hosseini2024benchmark} collect user data from their respective platforms: GLM, Qwen, and Lavita Medical AI Assist.
    \item \textbf{Synthetic Data}: Synthetic data is also a commonly used method for constructing benchmarks. This method refers to predefining a template and then filling it with different content to achieve both structure and diversity. The content can either be predefined or AI-generated. Benchmarks like LongFact~\citep{wei2024long}, LTF-TEST~\citep{jeung2024large}, and LongGenBench-SUTD~\citep{wu2024longgenbench} are constructed using this method.
    \item \textbf{Crowdsourcing (CrowdSrc)}: Data collection processes are usually non-automated, requiring human involvement to ensure data quality. ExpertQA~\citep{malaviya2023expertqa}, ProxyQA~\citep{tan2024proxyqa}, and QASA~\citep{lee2023qasa} collect questions from experts. Specifically, ExpertQA recruits experts through Prolific to write questions within their areas of expertise. ProxyQA manually creates meta-questions with the help of five experienced researchers. QASA involves AI/ML researchers to label data. Meanwhile, FACTS Grounding~\citep{jacovi2025facts} instructs third-party human raters to design prompts that require both the processing of long-form input and the generation of long-form output.
    \item \textbf{Publicly Available Datasets (PADs)}: ``Standing on the shoulders of giants'' is also an effective strategy for data collection. On one hand, the data quality of previous datasets is generally reliable, on the other hand, previous datasets often exhibit consistency, making it easier to process and adapt them. Many long-form generation benchmarks source their data from publicly available datasets. For example, LongGenBench-HKUST~\citep{liu2024longgenbench} collects its data from MMLU~\citep{hendrycks2020measuring}, GSM8K~\citep{cobbe2021training}, and CommonSenseQA~\citep{talmor2018commonsenseqa}. ASQA~\citep{stelmakh2022asqa} sources its data from AMBIGQA~\citep{min2020ambigqa}. Long2RAG~\citep{qi2024long2rag} sources its data from ELI5.
\end{itemize}

\subsubsection{Evaluation Paradigm}
\label{sec:long-form-evaluation-methods}

Evaluation methods are critical for benchmarks, as an appropriate evaluation method can provide a unified standard for evaluating the performance of different models, highlight their strengths and weaknesses, and guide model iteration. Evaluating the performance of long-form generation presents significant challenges. Here, we summarize the evaluation methods used in existing long-form generation benchmarks and categorize them into three types, offering insights for the development of future evaluation methods.
\paragraph{Automatic Metrics}

Automatic metrics are evaluation methods that evaluate model-generated responses without the need for additional human involvement or the use of LLMs. Most traditional metrics are classified as automatic metrics. 
Existing automatic metrics in the field of long-form generation can be categorized into four types:

\begin{itemize}
    \item \textbf{Semantic Aspect}: This type of metrics evaluates the semantic aspect of the response. Metrics like ROUGE~\citep{lin2004rouge} and BLEU~\citep{papineni2002bleu} are the most commonly used and were initially developed for summarization tasks and machine translation tasks. ROUGE and BLEU measure the semantic similarity between the response and the reference answer, providing an indication of the generation quality of LLMs. They are adopted as evaluation metrics in 13 out of 28 benchmarks. Similarly, METEOR~\citep{banerjee2005meteor} is also used to measure the semantic similarity and is applied in LongForm-C~\citep{koksal2023longform} and LongLamp~\citep{kumar2024longlamp}. Besides, BERTScore~\citep{zhang2019bertscore} is a metric that evaluates the semantic similarity of two texts using a finetuned BERT model~\citep{jeong2024olaph}.
    \item \textbf{Repetitive Aspect}: These metrics evaluate the repetitiveness or fluency of the responses. For instance, Perplexity (PPL)~\citep{liang2023open} calculates the overall probability of the response. Lower perplexity indicates higher fluency and suggests the sentence is more likely to be generated by an LLM. Metrics like Repetition~\citep{shao2019long} and Distinct~\citep{li2015diversity} are used to measure repetitiveness. Repetition-n calculates the percentage of n-grams that appear at least twice in the text, while Distinct-n quantifies text diversity by computing the number of unique n-grams in the output. These metrics are used in the LOT-OutGen~\citep{guan2022lot} and LCFO~\citep{costa2024lcfo}.
    \item \textbf{Accuracy-Based}: Accuracy is more commonly used in short-form generation scenarios like multiple-choice QA. 
    In long-form generation, LongGenBench-HKUST~\citep{liu2024longgenbench} combines multiple questions into a single input, requiring the model to generate responses for all of them. Each response is paired with a corresponding standard answer (either a specific option or a number), making it well-suited for accuracy-based evaluation.
    \citet{wu2024longgenbench} propose the CR and STIC metrics, both designed to measure the accuracy of responses in specific aspects.
    In addition, QASA~\citep{lee2023qasa} uses the standard accuracy as the evaluation metric.
    \item \textbf{Task-Specific}: Due to the varying focus of different benchmarks, task-specific metrics are often used. For example, ExpertQA~\citep{malaviya2023expertqa} uses QAFactEval~\citep{fabbri2021qafacteval} to evaluate the factual consistency of the responses, while ASQA~\citep{stelmakh2022asqa} uses Disambiguation Metrics to evaluate whether the response resolves ambiguities. Retrieval-oriented metrics, such as nDCG~\citep{rosenthal2024clapnq} and KPR~\citep{qi2024long2rag}, are used in CLAPNQ and Long2RAG to evaluate the effectiveness of the LLM-based RAG systems.
\end{itemize}

\paragraph{Human Evaluation}

Although automatic metrics are convenient, recent studies have shown that the correlation between automatic metrics and human evaluation is quite low~\citep{krishna2021hurdles,xu2023critical}. 
To achieve more accurate evaluations, the best approach is to establish standardized evaluation criteria and have human evaluators evaluate the responses of LLMs~\citep{ruan2024defining}. ELI5~\citep{fan2019eli5}, ASQA~\citep{stelmakh2022asqa}, Long2RAG~\citep{qi2024long2rag}, AQUAMUSE~\citep{kulkarni2020aquamuse}, Multi-News~\citep{fabbri2019multi}, LongForm-C~\citep{koksal2023longform}, LongBench-Write~\citep{bai2024longwriter}, LOT-OutGen~\citep{guan2022lot}, and HelloBench~\citep{que2024hellobench} have all conducted experiments involving human evaluation or provided human evaluation criteria.

\paragraph{LLM-as-a-Judge}

The main drawback of human evaluation lies in its time-consuming, labor-intensive, and inefficient nature. Recently, there has been a growing trend of using strong-performing LLMs to replace humans for fine-grained evaluations~\citep{zheng2023judging}. 
This strategy is also known as \textit{LLM-as-a-Judge}. 
In ProxyQA~\citep{tan2024proxyqa}, LLMs are used to evaluate whether the responses to proxy questions are correct. In LFMedQA~\citep{jeong2024olaph}, GPT-4o and Claude-3.5 are used for pairwise comparisons.
In HelloBench~\citep{que2024hellobench}, predefined checklists are set for each sub-task, and the LLM determines whether each checklist is satisfied. The overall score is calculated using human-correlated weights.

\subsubsection{Discussion}
\label{sec:long-form-discussion}

\paragraph{What Kind of Data Sources Are Better?}

The data source lays the foundation for the quality of a benchmark. In Section ~\ref{sec:long-form-data-source}, we categorize the data sources for long-form generation into five types: web-sourced data, user-sourced data, synthetic data, PADs, and CrowdSrc. Web-sourced data is advantageous in terms of scale and accessibility, but it often suffers from low quality. Similarly, while user-sourced data is directly obtained from users, its quality can also vary significantly. Synthetic data has the advantage of making evaluation easier, as it allows for the use of automatic metrics with high evaluation accuracy. However, such data often lacks alignment with real-world scenarios. Previous available datasets pose risks such as data leakage. While CrowdSrc provides high-quality data, it may still deviate from practical use cases. It is clear that each data source has its own limitations. So, what type of data can be considered good for long-form generation? We argue that user-sourced data combined with detailed post-processing is a promising direction for future research. Specifically, the optimization goal of LLMs is to serve users, which makes understanding the real issues users encounter critical. As such, obtaining first-hand interaction data directly from users is ideal. However, user-sourced data often contains low-quality samples, making data filtering a top priority. We believe that human involvement in the data post-processing phase is essential to achieve better outcomes. As long as data quality is ensured, having slightly fewer evaluation samples is unlikely to introduce significant evaluation bias. 

\section{Analysis} 
\label{sec:analysis}

\begin{figure}[!htp]
    \centering
    \includegraphics[width=1.0\linewidth]{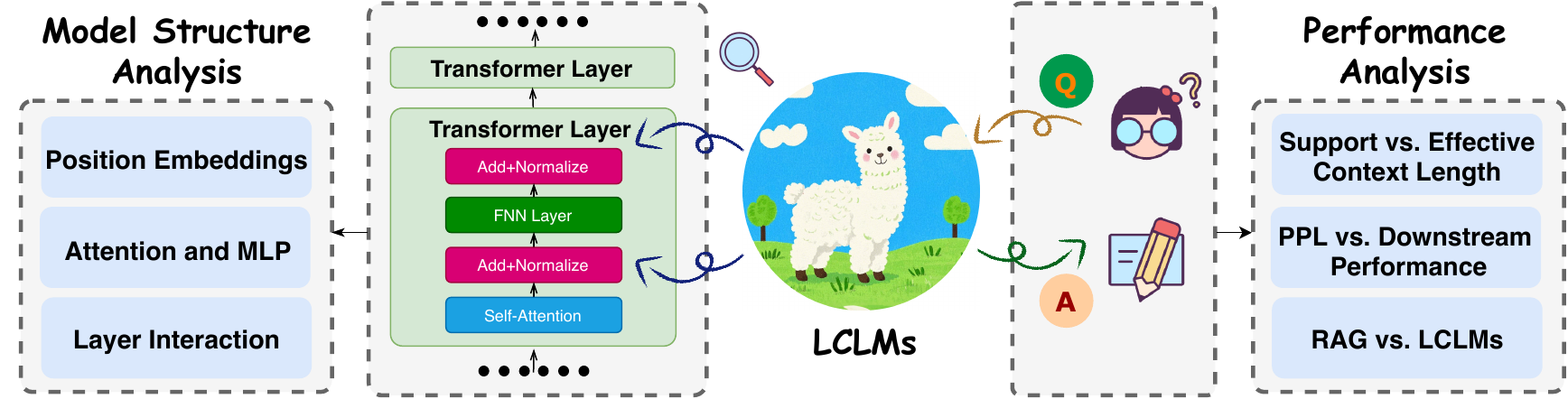}
    \caption{Illustration of analysis of LCLMs.}
    \label{fig:analysis}
\end{figure}
As shown in Figure~\ref{fig:analysis},
we provide the 
performance analysis and 
model structure analysis to understand neural network models externally and internally. 
This section provides a detailed review of these works.

\subsection{Performance Analysis} 
\label{ssec:blackbox_analysis}

First, LCLMs are analyzed as black boxes to examine their behavioral characteristics. In this subsection, we will first dive into the false promise of \textit{support} context length, related to the well-known \textit{lost-in-the-middle} phenomenon. Then, we explore the relationship between long context perplexity and real-world performance. We also discuss the ongoing debate between Retrieval-Augmented Generation (RAG) and long context modeling.

\subsubsection{The False Promise of \textit{Support} Context Length}

Recent years have witnessed remarkable growth in model context length, expanding from 4K tokens \citep{touvron2023llama} to 128K tokens \citep{grattafiori2024llama}, and even 10M tokens \citep{team2024gemini}. Despite this impressive scaling of context extension, there exists a significant gap between the context length these models claim to support and the actual context length they can process effectively \citep{liu2024lost,hsieh2024ruler,kuratov2024babilong}. A particularly noteworthy phenomenon in this regard is the \textit{lost-in-the-middle} effect. \citet{liu2024lost} demonstrate that many LCLMs exhibit a distinctive U-shaped performance curve: performance remains robust when information is located at the beginning or end of the input, but deteriorates substantially when critical information is positioned in the middle. \citet{an2024make} and \citet{he2023never} also observed that models progressively lose track of target information as the relative distance increases. To systematically assess the prevalence of this gap between supported and effective context lengths, RULER~\citep{hsieh2024ruler} conducted comprehensive evaluations of more than ten models, including both open source and proprietary ones, as illustrated in Table~\ref{tab:model_lengths}. The findings reveal a consistent pattern: For most models across both categories, the effective context length rarely exceeds half of the claimed length. This discrepancy underscores that, alongside pursuing ever-larger context window sizes, improving model performance within already supported context lengths is equally important. Results on BABILong~\citep{kuratov2024babilong} benchmark also support these evidences.

\begin{table*}[!htp]
\centering
\footnotesize
\begin{tabular}{@{}lcccc@{}}
\toprule
\textbf{Models}               & \textbf{OpenSource}       & \textbf{Claimed Length} & \textbf{Effective Length} & \textbf{Effective Ratio} \\ \midrule
Llama2 (7B)                   & \Checkmark                    & 4K                     & -                          & -                     \\
Gemini-1.5-Pro                & \XSolidBrush                    & 1M                     & $>128K$                    & $>12.8\%$                      \\
GPT-4                         & \XSolidBrush                    & 128K                   & 64K                        & 50\%                     \\
Llama3.1 (70B)                & \Checkmark                    & 128K                   & 64K                        & 50\%                     \\
Qwen2 (72B)                   & \Checkmark                    & 128K                   & 32K                        & 25\%                      \\
Command-R-plus (104B)         & \Checkmark                    & 128K                   & 32K                        & 25\%                     \\
GLM4 (9B)                     & \Checkmark                    & 1M                     & 64K                        & 6.4\%                     \\
Llama3.1 (8B)                 & \Checkmark                    & 128K                   & 32K                        & 25\%                     \\
GradientAI/Llama3 (70B)       & \Checkmark                    & 1M                     & 16K                        & 1.6\%                     \\
Mixtral-8x22B (39B/141B)      & \Checkmark                    & 64K                    & 32K                        & 50\%                     \\
Yi (34B)                      & \Checkmark                    & 200K                   & 32K                        & 16\%                     \\
Phi3-medium (14B)             & \Checkmark                    & 128K                   & 32K                        & 25\%                     \\
Mistral-v0.2 (7B)             & \Checkmark                    & 32K                    & 16K                        & 50\%                     \\
LWM (7B)                      & \Checkmark                    & 1M                     & $<4K$                      & $<4\%$                      \\
DBRX (36B/132B)               & \Checkmark                    & 32K                    & 8K                         & 25\%                     \\
Together (7B)                 & \Checkmark                    & 32K                    & 4K                         & 12.5\%                      \\
LongChat (7B)                 & \Checkmark                    & 32K                    & $<4K$                      & $<12.5\%$                      \\
LongAlpaca (13B)              & \Checkmark                    & 32K                    & $<4K$                      & $<12.5\%$                     \\ \bottomrule
\end{tabular}
\caption{Comparison of Claimed Length and Effective Length of Various Models~\cite{hsieh2024ruler}.}
\label{tab:model_lengths}
\end{table*}

\subsubsection{Relevance of Long Context Perplexity and Real-World Performance}
Beyond measuring language modeling capabilities, empirical evidence has demonstrated that language models' perplexity on short texts exhibits strong correlation with their performance on short-text downstream tasks~\citep{huangcompression}. Specifically, lower perplexity scores on a held-out set of short texts consistently predict superior performance on downstream short-text tasks. However, this correlation becomes substantially weaker in long context scenarios. \citet{hu2024can}, \citet{an2023eval}, and \citet{sun2021long} all observed that different models' perplexity scores on long contexts fail to correlate with their long context comprehension capabilities.

Recent work has reestablished the role of perplexity in evaluating models' long context modeling capabilities. Through comprehensive experiments, \citet{lu2024controlled} demonstrate that when fine-tuning a single base model (LLaMA2-7B) with various context-extension methods—including PI~\citep{chen2023extending}, YaRN~\citep{peng2023yarn}, NTK~\citep{ntk}, LongLora~\citep{chen2023longlora}, Landmark Attention~\citep{mohtashami2023landmark}, and CLEX~\citep{chen2024clex}—the resulting models' perplexity scores on the GovReport~\citep{huang2021efficient} dataset exhibit significant correlation with their performance on long context downstream tasks (including Needle-in-a-Haystack, LongBench, and RULER). 
Concurrently, \citet{fang2024wrong} introduced LongPPL, a refined metric that eliminates interference from context-irrelevant tokens by computing perplexity exclusively on context-sensitive token distributions. 
Under this improved metric, models' long context perplexity scores show robust correlation with their long context downstream performance. 
Together, these investigations reestablish perplexity as an effective indicator of long context modeling capabilities, providing valuable reference for future reliable evaluation of LCLMs.

\subsubsection{RAG Meets Long Context LLMs}

Since the emergence of LCLMs, they have frequently been compared with Retrieval-Augmented Generation (RAG) in terms of performance. Given a query, the RAG pipeline first retrieves key information from corpora, then uses an LLM to generate answers based on this information, while LCLMs process entire corpora as input, implicitly identifying relevant information before producing answers. \citet{li2024retrieval,lee2024can} revealed that when ample computational resources are available, LCLMs deliver superior average performance compared to RAG, despite not being specifically trained for these tasks.

However, compared to RAG, directly employing LCLMs to generate responses based on entire corpora presents significant efficiency limitations. Therefore, instead of comparing RAG against LCLMs, current research has shifted toward combining these approaches to get the best of both worlds. \citet{li2024retrieval} proposed the Self-Route method, which dynamically directs queries to either RAG or LCLMs based on the model's self-assessment, thereby optimizing the balance between performance and computational cost. \citet{jiang2024longrag} leveraged LCLMs to obtain larger, semantically more coherent retrieval units for RAG. \citet{DBLP:journals/corr/abs-2410-05983} discovered that "hard-negative" examples significantly impact LCLM-based RAG, and therefore designed both training-free and training-based approaches to mitigate this challenge.

\subsection{Model Structure Analysis}

Section \ref{ssec:blackbox_analysis} examines LCLMs as \textit{black boxes}, discussing the false promise of supported context length, the relevance between long context perplexity and real-world performance, and the ongoing debate between LCLMs and RAG. Taking a step further, this section approaches LCLMs from a \textit{glass-box} perspective, providing deeper insights at the granularity of individual model components (e.g., positional embeddings, attention and MLP modules, transformer layers), thereby illuminating the internal workings that drive long context capabilities.
\subsubsection{Positional Embedding} 

In long context modeling, the larger the model's context length is, the more possibilities there are in downstream applications. However, models are typically trained on a small and fixed context length during pretraining. \textit{Length extrapolation} studies the problem of enabling models to digest longer texts than the training distribution. In this section, we review how position embeddings (specifically RoPE~\citep{su2024roformer}) could be extended for length extrapolation.
\paragraph{RoPE Essentials} 
RoPE~\citep{su2024roformer} explicitly encodes the relative position between tokens into attention.
Recall that for a model with even hidden size \(d\), the RoPE embedding~\citep{jianlin_rope_beta_base} for absolute position \(n\) would be:
\begin{align}
    \left[ \cos\left( \frac{n}{\beta^{\frac{0}{d}}} \right), \sin\left( \frac{n}{\beta^{\frac{0}{d}}} \right), \cos\left( \frac{n}{\beta^{\frac{2}{d}}} \right), \sin\left( \frac{n}{\beta^{\frac{2}{d}}}\right), \ldots , \cos\left( \frac{n}{\beta^{\frac{d - 2}{d}}} \right), \sin\left( \frac{n}{\beta^{\frac{d - 2}{d}}} \right)    \right] \label{eq:rope}
\end{align} 

The \(\beta\) in Equation \ref{eq:rope} is the rotation base. The trigonometric functions could be associated with hypothetical ``wave lengths'' \(\lambda_{2i} =\left\{ 2\pi \beta^{\frac{2i}{d}} \right\}_{i = 0}^{\frac{d}{2} - 1} \), which are indicative of the model's context capacity. For a given \(\beta\), the wave lengths range from \(2\pi\) to \(2\pi \beta\)~\citep{peng2023yarn}. A typical choice for \(\beta\) is 10000~\citep{su2024roformer}, yielding max wave length \(\sim 63K\). The hypothetical ``frequency'' associated with a wave length is \(f_{2i} = \frac{2\pi}{\lambda_{2i}}\). Therefore, the larger the base, the larger the wavelength and the lower the frequency is.

\paragraph{Position Interpolation (PI)}
Position interpolation~\citep{chen2023extending} scales positions linearly to enlarge context length. 
Suppose that the model is pretrained with context length \(L\) and we wish to extend it to \(L'> L\).
Let \(s =\frac{L'}{L}\) denote the scaling factor. 
PI maps the original position \(n\) to a hypothetical position \(n' = \frac{n}{s}\). The largest position \(L'= sL\) would be mapped to \(L\). 
Therefore, all mapped positions still fall within the training range. 
From the perspective of wavelength, new wavelengths would be \(\lambda_{2i}^{PI} =\left\{  2\pi s \beta^{\frac{2i}{d}}\right\}_{i = 0}^{\frac{d}{2}- 1} \). 
PI enables length extrapolation with minimal fine-tuning steps (\(\sim 1000\)).
\begin{figure}[!htp]
    \centering
    \includegraphics[width=0.7\textwidth]{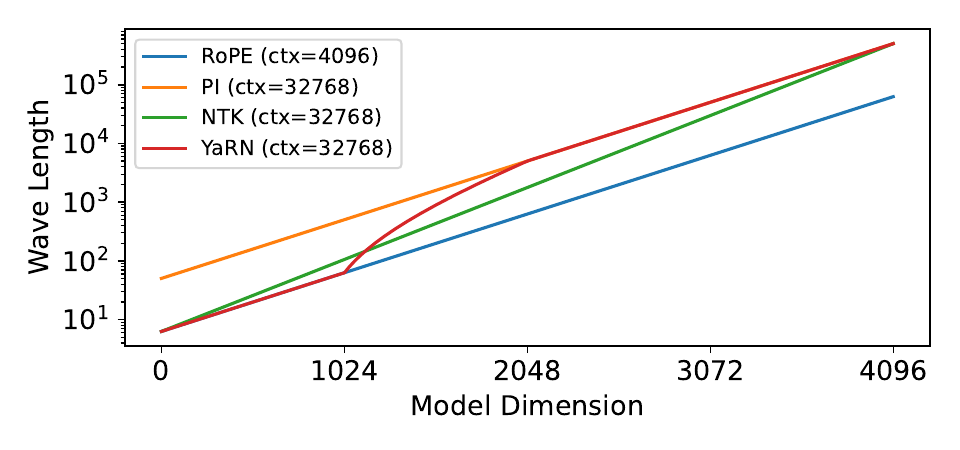}
    \caption{Associated wave lengths of RoPE, PI, NTK, and YaRN. Figure taken from \citet{peng2023yarn}. The y-axis uses log scale. For the sake of illustration, \(\alpha\) and \(\beta\) parameters for YaRN are set to \(\frac{1024}{2}\) and \(\frac{2048}{2}\). In practice, they are set to \(\frac{2}{2}\) and \(\frac{64}{2}\).} 
    \label{fig:rope_wave_lengths}
  \end{figure}
\paragraph{Frequency-Specific Scaling}
Cited works in this paragraph build on an important insight: injecting \textit{high frequency} embeddings improves performance and convergence~\citep{nips_fourier_high_freq}. On the contrary, PI scales up wavelengths, which transforms high-frequency embeddings into low-frequency ones. An initial solution named ``NTK-aware scaling''~\citep{ntk_aware_reddit} starts from unscaled high-frequency embeddings, then gradually transits to scaled low-frequency embeddings. Concretely, \(\lambda_{2i}^{NTK} =\left\{ 2\pi \left( s ^{\frac{d}{d - 2}}\beta \right)^{\frac{2i}{d}}  \right\}_{i = 0}^{\frac{d}{2}- 1} \). YaRN preserves unscaled high-frequency embeddings more conservatively than NTK, and transits to scaled low-frequency embeddings faster. Figure \ref{fig:rope_wave_lengths} illustrates the wavelengths of RoPE, PI, NTK, and YaRN.

\paragraph{Critical Rotation Base}
\citet{liu2023scaling} investigates the impact of both scaling up and scaling down the rotation base. Their setting is similar to PI, and they calculate critical scaling factors. Suppose that the model is pretrained with context length \(L\). Denote \(L\) as \(\lambda_{\text{train}}\) to connect it with trigonometric functions. The critical rotation base has the largest wave length that is smaller than \(\lambda_{\text{train}}\). Let \(I_{c}\) denote the critical dimension index, then \(I_{c} = 2\left\lfloor \frac{d}{2}\log_{\beta}\frac{\lambda_{\text{train}}}{2\pi} \right\rfloor\). Associated with \(I_{c}\) is the critical rotation base \(\beta_{c}\) and the critical wave length \(\lambda_{c}\).

When scaling down the rotation base from \(\beta\) to \(\beta_{\text{down}}\), phase change happens when the largest wave length \(2\pi \beta_{\text{down}}\) is equal to \(\lambda_{c}\). The gains of scaling down the rotation base is prominent until \(\beta_{c}\). When scaling up the rotation base from \(\beta\) to \(\beta_{\text{up}}\), \(\lambda_{c}\) would correspond to \(\left\{ \lambda_{c} \right\}_{\text{up}} \), which is indicative of the phase change point: model's perplexity would explode on texts longer than \(\left\{ \lambda_{c} \right\}_{\text{up}} \).

\citet{men2024baseropeboundscontext} further scrutinizes scaling rotation bases. They point out that scaling down rotation bases can only achieve superficial long context capabilities. The authors take one more step to derive the lower bound of the rotation base for an expected model context length. The derivation is based on the following theoretical insight: the model's ability to attend to a similar token decays as the distance increases, but the decay rate is slower for larger bases. Empirically, there is a polynomial relationship between effective context lengths and rotation bases.

\subsubsection{Attention and MLP Analysis}

\paragraph{Attention Head}
Attention heads typically serve specialized purposes. For example, previous research finds \textit{induction heads}, \textit{name mover heads} and \textit{arithmetic heads} in LLMs \citep{anthropic_induction_heads,wang_interpretability_wild_2023,arith_head_icml}.
Strides have been made in identifying crucial components for long context modeling. 
\citet{wu2025retrieval_head} identifies ``retrieval heads'', which are responsible for extracting information from long contexts. 
Ablating such heads leads to significant degradation across model families and model sizes.
Razor Attention~\citep{tang2024razorattention} refines retrieval heads into two subsets: echo heads and induction heads. Echo heads attend to identical tokens from the current location, while induction heads attend to the token after the identical token. ~\citet{fu2025not} further points out Retrieval-Reasoning (R2) Heads, which combines retrieval capabilities with reasoning abilities.

\paragraph{The Softmax Function and Attention Patterns}
The \texttt{softmax} function poses severe limitations on long context modeling.
As sequences grow longer, attention scores become increasingly uniform, thus the model cannot focus effectively~\citep{han-etal-2024-lm}. To mitigate this during length extrapolation, scaling coefficients are commonly adopted~\citep{peng2023yarn}.
Another prominent phenomenon is the attention sink effect~\citep{attn_sink}, where attention heads put heavy scores on the first token. The removal of these tokens from the KV cache can drastically impair model performance. 
Two solutions have been proposed to address this issue:
1. Introducing a dedicated trainable ``Sink Token'' to hold excessive attention scores.
2. Implementing alternative attention mechanisms to \texttt{softmax}, such as SoftMax-One~\citep{attn_off_by_one}.
\paragraph{MLP Layers}
\citet{voita-etal-2024-neurons_func} finds that the activation of certain MLP neurons is highly correlated with the current token's position. It also reveals that neurons could act as n-gram detectors, capturing local patterns. While the authors inspect only absolute positional embeddings, one could expect that there also exists neurons that activate strongly on certain relative positions.
\subsubsection{Layer Interaction}
Modern LLMs are composed of a stack of transformer layers each incorporating attention and MLP modules. Interestingly, researchers have discovered that alternating between full attention and linear-complexity mechanisms (such as window attention~\citep{yang_hybrid_attn_rope_nope} or lightning attention~\citep{minimax2025minimax01scalingfoundationmodels}) across layers yield superior extrapolation performance. Regarding position embedding,  \citet{yang_hybrid_attn_rope_nope} further discover that alternating between NoPE~\citep{kazemnejad2024impact} and RoPE can get the best of both worlds, as NoPE offers advantages in information retrieval, while RoPE better models local information due to its built-in recency bias.

\begin{figure}[!t]
    \centering
    \includegraphics[width=1.0\textwidth]{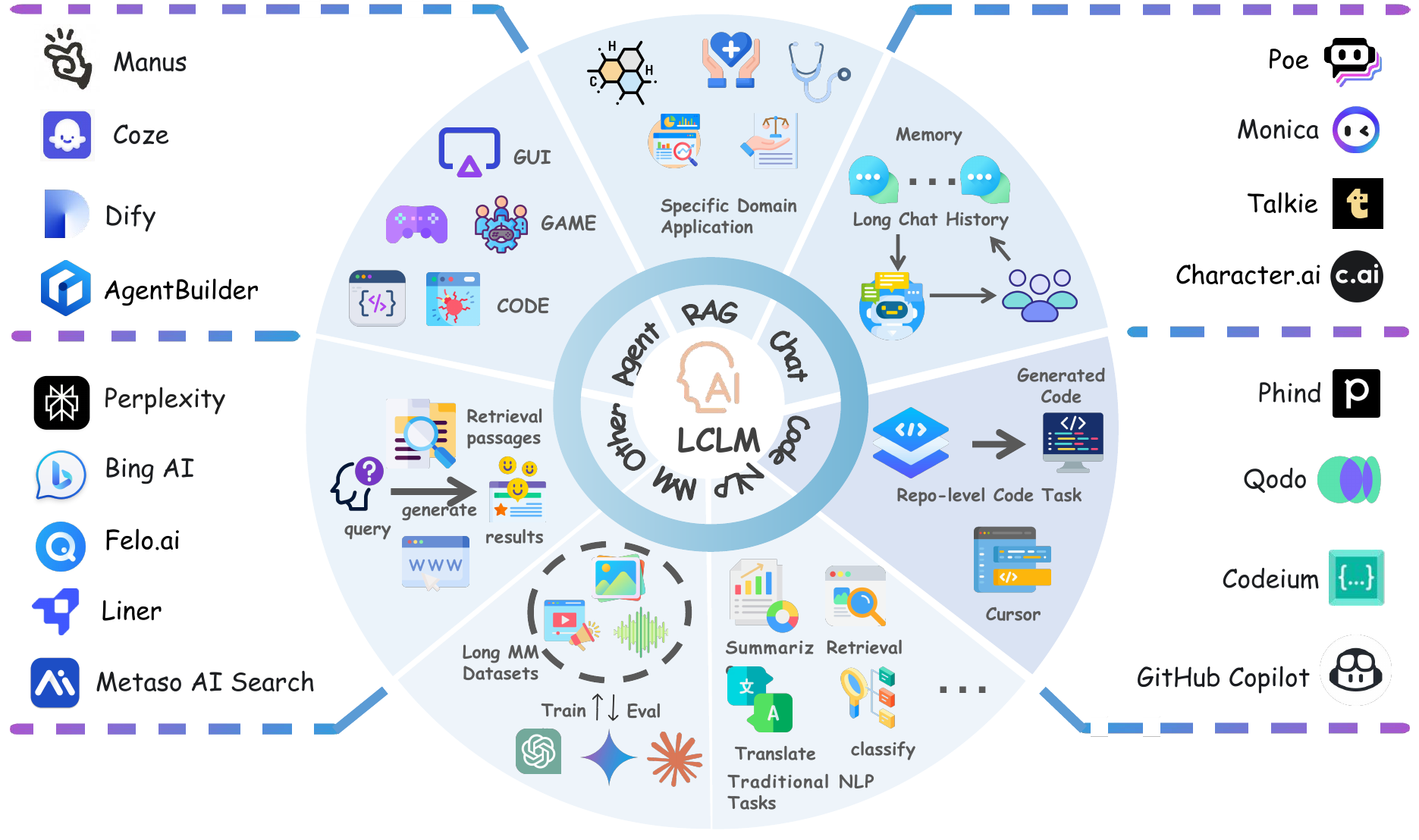}
    \caption{Overview of applications of long context language models and techniques across various tasks.}
    \label{fig:fig_applications_main}
    \vspace{3mm}
\end{figure}

\section{Application}
\label{sec:application}
The strong multitasking capabilities enable LLMs to be adapted to a wide range of applications. In real-world scenarios, tasks frequently require the processing of extensive contextual information. Consequently, long context LLMs significantly enhance the practical utility of LLMs in various domains, including GUI copilots and RAG-based systems, as illustrated in Figure~\ref{fig:fig_applications_main}. In this section, we briefly discuss the applications of long context LLMs and the associated techniques.

\subsection{Application in Agent}

LLM-based agents complete tasks via iterative interactions with environments, predicting next steps based on previous interaction histories~\citep{DBLP:conf/iclr/YaoZYDSN023,DBLP:journals/corr/abs-2309-07864,DBLP:journals/corr/abs-2309-07870}. Since both the environment observations and the interaction trajectories can be very lengthy, long context capability is crucial for developing effective agents.

Long context comprehension capabilities enable LLM agents to process complex agent tasks with extended observations. 
One example is GUI agent task~\citep{hu2024agents}, where the agents are required to comprehend the rich layout and text information~\citep{zhou2023webarena,DBLP:journals/corr/abs-2404-05955,xie2024osworld}. Another example is software engineering agent~\citep{jimenez2024swebench}. The agents are asked to interact with a repository to solve real-world coding tasks~\citep{DBLP:conf/nips/YangJWLYNP24,DBLP:conf/issta/0002RFR24,xia2024agentless,wang2024openhands_}.
Some agent tasks inherently require long-horizon reasoning and planning capabilities. 
For example, gaming agents must track the game state and plan actions over an extended horizon to progress or win~\citep{DBLP:journals/corr/abs-2403-03186}.
Tasks like planning multi-day travel itineraries~\citep{chen2024travelagent} or solving complex machine learning problems~\citep{DBLP:journals/corr/abs-2410-07095,DBLP:journals/corr/abs-2310-03302} also necessitate long-horizon planning. Automated agent optimization~\citep{DBLP:conf/icml/ZhugeWKFKS24,DBLP:journals/corr/abs-2406-18532} also requires good long context understanding and generation capabilities so that the LLM-based agent optimizer can effectively reason with long-horizon agentic workflows and intermediate results and generate optimized prompts, tools, and agentic workflows.

The long-context capabilities of large language models have facilitated the development of various agent applications. Open-source LLM agent development platforms, such as Dify~\citep{dify} and Coze~\citep{coze}, enable users to easily orchestrate LLM apps ranging from simple agents to complex AI workflows. General agents, like the OpenAI Computer-Using Agent~\citep{oai2025computeruse} and Manus~\citep{manus}, function as comprehensive AI assistants that handle real-world tasks by performing actions such as web browsing and coding within a virtual PC environment.

\subsection{Application in RAG}
Integrating long context technology into Retrieval-Augmented Generation (RAG) systems significantly enhances performance by enabling the processing of larger text chunks, retrieving more relevant information, and supporting the development of complex systems to address complex queries~\cite{yu2024defense,DBLP:journals/corr/abs-2410-05983,zhu2025doclens}. For instance, LongRAG~\citep{DBLP:journals/corr/abs-2406-15319}, which combines a `long retriever' and a `long reader' demonstrates notable improvements in document question answering. Similarly, studies~\citep{DBLP:journals/corr/abs-2410-05983,DBLP:journals/corr/abs-2405-03085} show that models capable of retrieving longer text fragments significantly enhance answer relevance and completeness. Furthermore, ~\citeauthor{DBLP:conf/acl/LiLL024} introduces `Reasoning with Attributions' which leverages long context models to improve multi-hop reasoning benchmarks by providing a larger context window for complex prompts. These advancements highlight how long context models reduce traditional RAG systems' reliance on external tools, offering a robust end-to-end modeling approach that enhances system performance and robustness.

Due to the aforementioned benefits, the rapid development of long context technology has spurred the growth of RAG systems, leading to popular applications like Perplexity~\citep{perplexity_pages} and Genspark~\citep{genspark}, which focus on retrieval. Other systems, such as ChatGPT~\citep{DBLP:conf/emnlp/0001YMWRCYR23} and Deepseek~\citep{deepseek_web_2023}, have also integrated retrieval capabilities. Notably, Deepsearch~\citep{deepsearch}, developed by xAI, enhances information navigation through advanced context understanding and intent recognition. These innovations underscore the critical role of long context technology in modern retrieval systems.

\subsection{Application in Chatbot}
Recent advancements in long context processing have significantly enhanced dialogue systems by enabling extended memory retention and contextual coherence. Long context windows allow chatbots to process extensive conversational histories, thereby improving interaction fluency and supporting long-term memory capabilities. This technological leap is foundational for applications requiring sustained user engagement, as evidenced by frameworks exploring prompt-based memorization~\cite{lee-etal-2023-prompted}, memory-augmented architectures~\cite{DBLP:conf/aaai/ZhongGGYW24}, and context extension techniques~\cite{longmem,wang2024limitssurveytechniquesextend}.

Due to the aforementioned benefits, long context processing technology has given rise to a diverse array of AI dialogue systems, demonstrating its broad value in practical applications. Major platforms like ChatGPT~\citep{openai2024memory} and Pi~\citep{inflection2023impi} utilize persistent memory to maintain user preferences and conversation history, enabling personalized interactions. Specialized systems such as Character AI~\citep{character_ai} and Talkie~\citep{talkie} further exemplify how long context windows support style-consistent dialogues and contextual continuity. These capabilities prove particularly valuable in longitudinal applications including educational tutoring, healthcare monitoring, and therapeutic counseling, where preserving the context over multiple conversations is critical~\cite{jo2024understanding}.

\subsection{Application in Code}

The advancement of long context technology has significantly enhanced repository-level code tasks and software development agent tasks~\cite{jiang2024survey}. Such tasks often require processing complex and extensive contextual information, which previously constrained models' ability to integrate comprehensive context. For instance, RepoCoder~\cite{zhang2023repocoder} employs a similarity-based retrieval mechanism to enrich context and improve code completion quality. Similarly, the RLPG framework~\cite{Shrivastava2022RepositoryLevelPG} integrates repository structures with relevant file contexts to generate repository-level prompts. These intricate system designs were necessitated by the limitations of earlier context windows. However, breakthroughs in long context technology have effectively alleviated these constraints, enabling the development of more sophisticated systems that enhance task execution precision and efficiency~\cite{shrivastava2023repofusion}. 

Recent models incorporating long context technologies, such as StarCoder2~\cite{lozhkov2024starcoder2stackv2}, Qwen2.5-Coder~\cite{hui2024qwen2}, and Granite Code Models~\cite{DBLP:journals/corr/abs-2405-04324}, have demonstrated superior performance in long context code tasks. Unlike earlier approaches that relied on complex architectures to manage extended contexts, these models leverage robust long context comprehension capabilities, providing scalable and practical solutions for software development. This progress has streamlined tasks such as code completion, contextual understanding, and repository-level code generation.

Due to these advancements, long context technologies have spurred the development of advanced AI-powered code applications, such as GitHub Copilot~\cite{github_copilot} and Anysphere Cursor~\cite{cursor_ai_2025}. These tools utilize long context models to enhance code completion, deliver contextual documentation, and enable predictive debugging. By maintaining a coherent understanding of large codebases, they improve developer productivity, reduce manual effort, and support more intuitive workflows. This integration highlights the transformative impact of long context technology on modern software development.

\subsection{Application in Traditional NLP Tasks}
The introduction of long context technology has led to transformative breakthroughs in numerous traditional NLP tasks by addressing the limitations of conventional methods, which were constrained by restricted context windows. This technological advancement has significantly enhanced the performance and practical applicability of various NLP applications. In this section, we briefly discuss three representative tasks that have particularly benefited from long context technology.

\paragraph{Document Summarization}
Traditional document summarization methods often encounter challenges such as inconsistencies and limited contextual understanding~\cite{zhang2019hibertdocumentlevelpretraining,zhang2020pegasus}. The development of long context models enables the processing of entire documents at once, facilitating a more comprehensive understanding of the original content and improved identification of redundant information~\cite{zaheer2020big,wang2024study,godbole2024leveraging}. Models such as Longformer~\cite{DBLP:journals/corr/abs-2004-05150} and LongT5~\cite{DBLP:conf/naacl/GuoAUONSY22} have demonstrated exceptional performance in document summarization tasks, thereby advancing the practical implementation of applications such as summarizing news articles and academic papers~\cite{jin2024comprehensive}.
\paragraph{Information Retrieval}
Traditionally, most vector models were limited by window size, which required segmenting the data into blocks when performing long context semantic modeling, often resulting in the loss of coherent semantic information~\cite{zhu2023large,DBLP:conf/acl/WangYHYMW24}. However, with advancements in long context technology, semantic vector models can now accommodate longer text inputs, such as text-embedding-3-large~\cite{openai2024embedding}, jina-embeddings-v2~\cite{günther2024jinaembeddings28192token}, and BGE-M3~\cite{chen2024bge}. This enables the processing of longer texts such as chapters and documents, thereby significantly improving the practical usability of semantic vector models in real-world applications~\cite{DBLP:conf/emnlp/ZhuW0SWWL24,saad2024benchmarking}.
~\paragraph{Machine Translation}
Document translation is a key research area in machine translation. Early approaches focused on modifying the transformer architecture to encode more contextual information~\cite{miculicich2018document,zhang2018improving,bao2021g,DBLP:conf/emnlp/WangLJZY0T23}. Long context models, however, can directly translate lengthy, complex documents by processing extended context windows, which improves the translation of polysemous words~\cite{herold2023improving,wang2024benchmarking}. This advancement significantly enhances the translation quality of long documents and makes it increasingly feasible for large language models to translate entire novels and books~\cite{lyu2023paradigm}.

\subsection{Application in Multimodal Tasks}
Understanding extensive videos, large collections of images, audio streams, and lengthy text inputs is often essential in real-world scenarios. Therefore, multimodal long context modeling has become a key area of research in long-text applications. This section provides an overview of relevant data and training strategies, models, and frameworks, as well as benchmarks for long context multimodal large language models (MLLMs).

\paragraph{Data and Training Strategies} To extend the context windows of MLLMs, many studies design the data curation pipeline to filter and synthetic long context multimodal training data. Typically, these datasets, unlike textual long context training data, are constructed by increasing the length of the multimodal part in the context, such as multi-image~\citep{DBLP:journals/corr/abs-2411-19951}, long video~\citep{DBLP:journals/corr/abs-2406-14129,DBLP:journals/corr/abs-2501-00574}. 
Leveraging such data, several studies~\citep{zhao-2024-arxiv-OmChat,li-2024-arxiv-llavaonevision,wang-2024-arxiv-LongLLaVA,hannan-2024-arxiv-revisionllm,zhang-2024-arxiv-intenlmxcomposer25,xue-2024-arxiv-LongVILA,tu2025longwriter} enhance MLLMs' long context capabilities by training on progressively longer multimodal contexts.
After that, some work~\citep{li-2025-arxiv-TPO} utilizes the variants of preference optimization to improve the long context modeling of MLLMs. Alternatively, some work utilizes long context LLMs as the foundational model of MLLMs for training~\citep{jang-2024-emnlp-mate} to inherent long context modeling ability of LLMs.

\paragraph{Model Structure and Framework} To enhance the ability of MLLMs to understand multimodal long context input and reduce the computational costs in model training and inference, some work focuses on modifying position embeddings and introducing emergent modules in MLLMs for better inference efficiency and performance. Next, we will elaborate on the details of these techniques. Unlike text-only long context language modeling, to effectively model the positional information of visual tokens in multimodal long context inputs, some studies~\citep{Qwen2VL,ge-2024-arxiv-v2pe,su-kexuefm-2024-ropetie} propose extending the dimensionality of the ROPE adopted by numerous MLLMs. The dimensional extension enables ROPE can improve the sequential modeling in multimodal tokens, the spatial modeling in images, and the temporal modeling between video frames, thereby improving the differentiating the positional information of visual tokens in multimodal long context inputs. In addition to enhancing the ability of multimodal models to capture positional information, some studies introduce specialized modules to further reduce the context window length occupied by visual tokens with MLP~\citep{Qwen2VL}, pixel shuffle~\citep{chen-arxiv-2024-internvl15}, Q-former-like module~\citep{Alayrac-2022-nips-flamingo,li-2023-icml-blip2,zhu-2024-arxiv-focusllava,ye-2024-arxiv-mPLUG-Owl3}, LoRA~\citep{ma-2024-arxiv-vlora}, which integrate effective important visual information into LLMs eliminating redundant information in visual representations. 

However, the aforementioned methods to reduce the length of visual tokens necessitate training MLLMs from scratch for adaptation. To compress the number of visual tokens to reduce inference costs for well-trained MLLMs, some approaches~\citep{liu-2024-arxiv-mustdrop,han-2024-arixv-ficoco,yang-2024-arxiv-visionzip,vasu-2024-arxiv-fastvlm} in a training-free manner preserve important visual tokens by assessing their relevance with input text. Generally, for multimodal long context inputs, videos occupy the largest portion of the context window due to containing numerous frames. Therefore, some work utilizes tree structure~\citep{wang-2024-arxiv-videotree}, reward models~\citep{yang-2024-arxiv-vca}, similar scores based on CLIP~\citep{liang-2024-arxiv-keyvideollm}, trained video frame selector~\citep{yu-2024-arxiv-framevoyager,hannan-2024-arxiv-revisionllm}, and aggregating methods~\citep{faure-2024-arxiv-hermes} to select keyframes and decrease the length of video in context.

\paragraph{Benchmarks} After introducing the basic and advanced development of long context MLLMs in the community, in this part, we will review existing evaluation benchmarks for assessing the performance of multimodal long context modeling capabilities of MLLMs. Multimodal long context benchmarks can primarily be divided into two types: those that use multiple images and those that use long videos as the context. These benchmarks then require the model to answer questions based on the contextual information to evaluate the MLLMs' capability of multimodal long context modeling.  For benchmarks using multi-image context, some works~\citep{ma-2024-arxiv-mmlongbenchdoc,pramanick-2024-arxiv-spiqa, chia-2024-arxiv-mlongdoc} use a large number of multimodal cases as context to evaluate the model’s many-shot in-context learning capabilities. Other works employ \emph{needle in a haystack}~\citep{wang-2024-arxiv-MMNeedle}, temporal, and semantic multi-image tasks~\citep{song-2024-arxiv-milebench} as the primary evaluation methods. Additionally, some studies~\cite{jiang-2024-arxiv-manyshots,Sharma-2024-arxiv-LOCOVQA,Ruoss-2024-arxiv-lmact} not only assess the understanding of long multimodal contexts but also evaluate the multiple document comprehension capabilities. Taking this further, certain attempts aim for a comprehensive evaluation by integrating multiple task formats. MMLongBench~\citep{wang2025mmlongbench}, for example, covers multimodal needle-in-a-haystack (NIAH), long-document VQA (DocVQA), visual retrieval-augmented generation (VRAG), and many-shot in-context learning (ICL). Compared to multi-image long context data, understanding long videos is more challenging due to their larger context length and the inclusion of more redundant and noisy information. Some work~\citep{zhou-2024-arxiv-mlvu,zou-2025-arxiv-hlv1k,Chandrasegaran-2024-arxiv-videohour,gang-2024-arxiv-longvale} utilizes diverse long video and multiple challenging tasks to evaluate multimodal long context modeling of MLLMs. Other works~\citep{Nagrani-2024-arxiv-Neptune,DBLP:journals/corr/abs-2406-14129,wu-2024-arxiv-longvideobench} collect videos with rich events through both automated and manual methods and pose questions on the events.

\subsection{Application in Specific Domains}

Long context technology has demonstrated significant potential across various domains by enabling more efficient processing of complex and lengthy information~\cite{godbole2024leveraging}. In news, it improves the generation of coherent summaries by aggregating data from multiple sources~\cite{gao2019abstractive}. In the legal field, it simplifies the interpretation of large volumes of documents, helping professionals quickly extract key insights~\cite{kapoor2024promises}. In healthcare, long context models enhance the synthesis of patient records and medical literature, thereby supporting informed decision-making~\cite{DBLP:journals/corr/abs-2406-15019}. Financial applications utilize these models to analyze extensive reports and derive insights into market trends~\cite{reddy2024docfinqa,DBLP:journals/corr/abs-2401-15050,nie2024survey}. In the field of biology, long context technology aids in understanding molecular structures and genomic sequences, fostering advancements in drug discovery and personalized medicine~\cite{hilgert2024evaluating, shao2024long}. Overall, long context models have significantly enhanced the effectiveness and accuracy of information processing in these areas, making data synthesis and information extraction more efficient.

\section{Future Directions}
\label{sec:directions}
\subsection{Long Context Modeling for o1-like Long Reasoning}

Most recently, o1-like long reasoning models have attracted significant attention due to their exceptional performance on complex reasoning tasks~\citep{qin2024o1,huang2024o1,zhang2024o1,zhao2024marco,guo2025deepseek,team2025kimi,muennighoff2025s1,chen2025r1v}. This test-time scaling paradigm that first generates extended CoT reasoning before producing answers essentially equips models with the ability to perform trial-and-error, backtracking, correction, and iteration auto-regressively within the context window, unlocking significant performance improvements~\citep{li2025llms,wang2024drt,li2025small}. Despite its promise, current practices leveraging LongCoT remain far from satisfactory, primarily due to two critical drawbacks: the rough quality of generated CoTs severely impacts reasoning efficiency, containing substantial redundancy and irrelevant information~\citep{zhang2025lightthinker,xia2025tokenskip,chen2024not,hu2025efficient,ma2025cot}; and the difficulty in scaling CoT length significantly constrains the potential for further performance gains, as numerous studies have observed performance degradation in particularly long reasoning chains~\citep{zeng2025revisiting,yang2025towards,wu2025more}.

Addressing these challenges requires advances in long context language models, which are essential for two key aspects: reliable evaluation of long reasoning processes (currently a significant challenge for existing Process Reward Models~\citep{he2025can}), and developing robust long-form generation capabilities for producing longer and more reliable reasoning chains. Furthermore, tailoring efficiency-oriented techniques developed for general long context scenarios, such as KV cache compression and prompt compression, to the specific needs of long reasoning models presents a promising direction for future research.

\subsection{Further Extending Context Window and Improving Modeling Capabilities}
Context length stands as one of the most fundamental attributes of language models. Its evolution closely tracks the development of increasingly sophisticated modeling capabilities - from basic n-gram models constrained to mere tokens of context~\citep{brants2007large}, to BERT-style architectures processing hundreds of tokens for paragraph-level understanding~\citep{devlin-etal-2019-bert}, to early iterations of ChatGPT managing 2k-4k tokens for general-purpose dialogue, and now to unprecedented scales of 100k to 1M tokens which has enabled sophisticated capabilities including long chain-of-thought reasoning~\citep{guo2025deepseek}, long in-context learning~\citep{team2024gemini}, long video processing~\citep{Qwen2.5-VL}, and supporting complex agent systems with extensive interaction histories~\citep{oai2025deepresearch}. This progression demonstrates how expanding context length has consistently unlocked new frontiers in language modeling, enabling increasingly sophisticated applications and use cases. Therefore, extending context windows to even greater scales holds the promise of unleashing more advanced capabilities and transformative applications.
On the other hand, existing research has demonstrated a significant gap between models' claimed supporting context length and their effective context length~\citep{hsieh2024ruler,an2024does}. Consequently, enhancing models' long context modeling capabilities within their supported context lengths remains crucial. Regarding the expansion of context windows and improvement of long context modeling capabilities within supported context windows, we have identified several promising directions:
\paragraph{Long Context Reinforcement Learning} Despite its immense potential, reinforcement learning remains underexplored in long context scenarios. While combining LongCoT with RL has demonstrated significant improvements for tasks with definitive answers, many long context applications still face fundamental challenges as they require referring to lengthy inputs. One example is alignment scenarios, where long context RLHF practices—whether employing long context models as reward models (e.g., GLM-Long~\citep{glm2024chatglm}) or utilizing long context preference data (e.g., LLaMA-3.1~\citep{grattafiori2024llama})—have yet to yield substantial benefits. The core challenge of long context RL lies in developing reward models capable of effectively evaluating extensive inputs, such as lengthy reasoning chains, narrations, and dialogues. Oriented toward this goal, collecting long context preference data, either through human annotation or carefully designed synthesis protocols, merits further investigation.

\paragraph{Recipe for Collecting, Filtering and Synthesizing High-Quality Training Data} The advancement in long context modeling capabilities is fundamentally data-driven. The data recipe for training long context models presents several promising directions for future exploration: (1) developing fine-grained filtering strategies beyond heuristics to identify training data with long-range dependencies; (2) synthesizing challenging queries that require integrating key information dispersed throughout the text, along with their correct answers; (3) exploring task types particularly suited for long context RL training that best generalize to other long context tasks; and (4) optimizing the distribution of domains and sequence lengths to achieve Pareto efficiency between computational cost and model performance. Addressing these challenges will be essential for developing more efficient and effective data recipes for future LCLMs. 

\paragraph{Long Context Distillation} Model distillation is a common strategy to enhance the capabilities of smaller models by leveraging larger, more powerful models. The de facto approach involves generating responses from a strong model for a set of queries, then fine-tuning the smaller model on these question-answer pairs. This method has proven effective in various aspects, including improving human preference ratings for model responses and enhancing models' chain-of-thought reasoning abilities. Within the long context domain, an interesting research question emerges: how can we utilize large models with strong long context modeling capabilities to improve models with weaker long context abilities? Beyond generating high-quality responses~\citep{longalign}, an intriguing observation is that powerful LCLMs can effectively identify and filter training data that exhibits strong long-range dependencies, with more capable models demonstrating superior filtering efficacy~\citep{wu2025longattn}. Given that LCLM training is a complex process spanning both pre-training and post-training stages, we believe that models with strong long context modeling capabilities can play a more significant role in the LCLM training process, improving both training efficiency and resulting model performance.

\paragraph{Performance-Oriented Architecture Design} Architectural design remains an ongoing research topic in long context modeling. Beyond improving training and deployment efficiency, architectural modifications such as enhanced positional encodings and hybrid attention mechanisms can lead to better extrapolation capabilities and improved modeling within supported context lengths, as discussed in Section~\ref{sec:model}. We anticipate continued advances in this direction to further enhance LCLMs' modeling capabilities.
\paragraph{Optimizing Long-Form Generation} Current research on LCLMs predominantly focuses on processing long inputs, such as information retrieval from extensive documents or concise summary generation. By contrast, long-form generation remains relatively unexplored, despite its promising applications in areas such as long-form narrative creation, repository-scale code generation, and long CoT. Advancing long-form generation capabilities presents two significant challenges: First, beyond context comprehension, the automated generation of high-quality long-form content demands sophisticated output planning, requiring a higher level of language modeling capabilities. Second, the evaluation of generated long-form content poses substantial difficulties - automated metrics like ROUGE demonstrate limited reliability for lengthy texts, while manual assessment proves prohibitively time-consuming. Consequently, future advances in long-form generation may need to develop in tandem with improvements in automated evaluation methodologies for long-form content.
\paragraph{Domain-Specific Enhancement} As mentioned above, the expansion of context windows has unlocked diverse application domains for language models, ranging from text-intensive fields such as legal and medical domains to multimodal scenarios where vision-language models process extensive visual content, and to intelligent systems maintaining rich interaction histories. While preserving robust general-purpose long context modeling capabilities remains essential, investigating targeted optimizations for these domain-specific challenges presents a promising research direction.

\subsection{Efficient Architecture Design, Training, and Deployment of LCLMs}

\paragraph{Model Architecture} The inherent limitations of Transformer models become pronounced in long context scenarios, primarily due to the substantial KV-cache memory overhead. The memory significantly exceeds the High Bandwidth Memory (HBM) capacity of contemporary GPUs. Consequently, the exploration of memory-efficient KV-cache architectures, such as the investigation of linear attention mechanisms, emerges as  critical directions for the architectural advancement of Long Context Language Models.

\paragraph{Training and Inference Frameworks}  To mitigate training complexities, the implementation of refined partitioning strategies~\citep{qi2023zero} and local recomputation techniques~\citep{korthikanti2023reducing} assumes paramount importance. These methodologies which balance bandwidth overhead with computation time are crucial to effectively address the growing divergence between increasingly complex model architectures and the continuously evolving capabilities of GPU. For inference frameworks, the strategic deployment of operator fusion and the rearrangement of computational sequences to curtail the generation of extensive intermediate matrices remains a salient avenue for investigation~\citep{dao2022flashattention,dao2024flashattention,shah2025flashattention}. Despite prevailing precision norms anchored at 16-bit, state-of-the-art hardware architectures already accommodate high-throughput computation at reduced precisions, down to 8-bit~\citep{nvidia_h200}. The application of 8-bit quantization to selectively chosen modules within models, applicable to both training and inference paradigms, has demonstrated considerable promise~\citep{liu2024deepseek}.  Prospective research endeavors focused on exploring even lower precision quantization levels, and the expanded application of quantization across a broader spectrum of model modules, are anticipated to yield substantial performance enhancements.
Recently, the advent of long context Reinforcement Learning (RL) training~\citep{guo2025deepseek} introduces novel challenges to existing frameworks, necessitating targeted optimizations to bolster training efficiency and stability. 

\paragraph{Customized Hardware Design} The rapid improvement in hardware computational speed has surpassed the corresponding gains in bandwidth, thereby exacerbating bandwidth bottlenecks, particularly during the decoding phase~\citep{gholami2024ai}.  While existing Prefill-Decoding disaggregated architectures facilitate the independent optimization of prefill and decoding stages, there remains a notable absence of specialized hardware designed for decoding. Decoding processes are inherently characterized by substantial demands for GPU with more HBM capacity and higher bandwidth, while exhibiting comparatively lower computational intensity. Therefore, the development of dedicated hardware specifically engineered for decoding, characterized by enhanced HBM capacity and bandwidth, constitutes a crucial direction for future hardware evolution in this domain.

\subsection{More Reliable Evaluation Framework for Long Context Language Modeling}
As discussed in Section~\ref{sec:evaluation}, the capabilities of long context language modeling can be categorized into two main aspects: long context comprehension and long-form generation. Consequently, future research directions naturally align with these two fundamental tracks.

\paragraph{Towards Real-World and Scenario-Specific Long Context Comprehension}
Regarding long context comprehension, given that numerous synthetic and real-world benchmarks have emerged to evaluate various aspects of this capability, future research can be directed towards more specific real-world applications. This includes mining user logs to understand authentic long context usage patterns, assessing LCLM performance across specialized domains (such as legal, medical, and financial sectors), and evaluating its effectiveness in different academic disciplines (from social sciences to natural sciences), etc. Notably, the extensive input lengths pose significant challenges for obtaining high-quality human annotations. Therefore, developing efficient annotation frameworks specifically designed for long context scenarios presents another valuable research direction.

\paragraph{Coarse-to-Fine Evaluation of Long-Form Generation} 

In the field of long-form generation, human evaluation is frequently used, mainly because automatic metrics struggle to effectively evaluate long responses and are less suited to open-ended generation tasks. 
Human evaluation and automatic metrics are two extremes: automatic metrics are highly efficient but lack precision, while human evaluation offers high accuracy but is significantly resource-intensive. How can we combine the advantages of both? 
By observing the trends of evaluation methods for long-form generation, we can notice a transition from a combination of automatic metrics and human evaluation toward the adoption of LLM-as-a-Judge. 
Nonetheless, fully end-to-end evaluation of long responses remains a challenge for current LLMs. To address this, an evaluation workflow can be designed, decomposing the entire evaluation into granular aspects, following a systematic coarse-to-fine adaptation. For simpler evaluation aspects, LLMs combined with prompt engineering can serve as alternatives to human evaluators. 
For components where LLMs struggle to evaluate, further decomposition of the evaluation workflow or adopting proxy-based methodologies can be explored. This systematic breakdown allows us to preserve high evaluation accuracy while enhancing efficiency, ultimately demystifying the challenges tied to long-form generation evaluation. 
Currently, ProxyQA~\citep{tan2024proxyqa}, HelloBench~\citep{que2024hellobench}, and LongFact~\citep{wei2024long} reflect this trend, but we believe there is still potential for optimizing these systems further to achieve even higher accuracy.

\subsection{Towards Mechanistical Interpretability for Long Context Modeling}
Mechanistic Interpretability (MI)~\citep{nanda2023mechanistic} aims to reverse engineer models at a module level. In MI literature~\citep{meng2022locating,arith_head_icml,wang_interpretability_wild_2023}, researchers typically pinpoint a sparse set of attention heads and MLPs that are responsible for some downstream task.
One exemplar of MI for long context modeling is the identification of \texttt{retrieval heads} \citep{wu2025retrieval_head} that are tied to long-range dependencies. MI for long context modeling is an exciting intersection for both worlds. We point out several promising research questions to answer.

\paragraph{Mechanistic Understanding of Position Embeddings}
How do LLMs make use of positional embeddings? \citet{voita-etal-2024-neurons_func} showcases that the activation weights of certain neurons are highly correlated with the current token's position, yet they analyze only absolute positional embeddings. Future works could shed light on whether the same holds for relative positional embeddings, and on the interaction between MLPs and attention modules when processing positional information. 

\paragraph{Identifying Causes of Long Context Issues}
We advocate a problem-driven approach to MI for long context modeling. For example, \citet{liu2024lost} points out that models have a positional bias for processing information, favoring tokens at the start or at the end. We take one step further and ask: What modules of the model lead to such a bias? Based on possible discoveries,  can we devise a method to mitigate it mechanistically? 

\paragraph{Interpretability-Driven Enhancements}
Closely related is the problem of length extrapolation. \citet{zhou-etal-2024-num_sys} demonstrates that LLMs fail to align digits when performing addition on numbers longer than those in the training distribution. Similarly, we ask: What modules of the model are responsible for this failure? Can we build on the findings to improve length extrapolation?
At a high level, MI provides useful toolchains to understand the inner workings of LLMs and to pinpoint problematic components. On the other hand, long context modeling is a challenging field with its unique interests and a huge application space. MI for long context has huge potential for efficiently identifying and addressing issues in long context modeling.

\section{Conclusion}
In this paper, we provide a comprehensive survey on the recent advances on long context modeling for large language models. 
Specifically, we first discuss the long context data strategy including data filtering, data mixture and data construction strategies for both pre-training and post-training stages.
Then, we provide a comprehensive discussion of the recent advancement on model including Transformer-based LLM architectures and position embeddings, which aim at enhancing the long context capabilities of LLMs.
After that, we provide the workflow-based LCLM methods,
such as prompt compression and agent-based methods, to enhance the long context abilities.
Besides, we discuss the long context evaluation benchmarks including understanding and generation abilities.
Moreover, we provide the AI infrastructure method including training and inference strategies optimized for long context modeling.
In addition, we also provide a detailed analysis for better interpretability of LCLMs.
Furthermore, we summarize existing applications (e.g., agent, RAG, Code) on long context modeling,
and discuss the remaining issues or challenges for future directions. 
Finally, we hope our survey could guide the developers to better understand the related techniques for LCLMs and facilitate the growth of foundation models.

\clearpage
\newpage

\section{Contributions and Acknowledgments}
\paragraph{Project Leaders}
\begin{itemize}
    
    \item Jiaheng Liu, Nanjing University, Institute of Automation, Chinese Academy of Sciences, M-A-P, \textit{Organize the whole project}
    \item Dawei Zhu, Peking University, \textit{Organize the whole project}
\end{itemize}

\paragraph{Core Contributors (Alphabet Order)}
\begin{itemize}
   
\item Zhiqi Bai, Alibaba Group, \textit{Architecture}
\item Yancheng He,  Alibaba Group, \textit{Application}
\item Huanxuan Liao, Institute of Automation, Chinese Academy of Sciences, \textit{Data}
\item Haoran Que, M-A-P, \textit{Evaluation}
\item Zekun Wang, Kuaishou Technology, \textit{Workflow Design}
\item Chenchen Zhang, Tencent, \textit{Analysis}
\item Ge Zhang, ByteDance, M-A-P, \textit{Architecture \& Evaluation}
\item Jiebin Zhang, Peking University, \textit{Infrastructure}
\item Yuanxing Zhang, Kuaishou Technology, \textit{Infrastructure}
\end{itemize}
\paragraph{Contributors (Alphabet Order)}
\begin{itemize}
    \item Zhuo Chen, Alibaba Group, \textit{Architecture}
 \item Hangyu Guo, M-A-P, \textit{Application}
 \item Shilong Li, Alibaba Group, \textit{Application}
 \item Ziqiang Liu, Shenzhen Institutes of Advanced Technology, Chinese Academy of Sciences, \textit{Workflow Design}
 \item Yong Shan, ByteDance,  \textit{Architecture}
 \item Yifan Song, Peking University,  \textit{Application}
 \item Jiayi Tian, Alibaba Group, \textit{Data}
 \item Wenhao Wu, Peking University, \textit{Future Directions}
 \item Zhejian Zhou, University of Southern California, \textit{Analysis}
 \item Ruijie Zhu, UC Santa Cruz, \textit{Architecture}
\end{itemize}

\paragraph{Sponsor Committee (Alphabet Order)}
\begin{itemize}

    \item Junlan Feng, China Mobile Research Institute
 \item Yang Gao, Nanjing University
 \item Shizhu He, Institute of Automation, Chinese Academy of Sciences
 \item Zhoujun Li, Shenzhen Intelligent Strong Technology
 \item Tianyu Liu, M-A-P
 \item Fanyu Meng, China Mobile Research Institute
 \item Wenbo Su, Alibaba Group
 \item Yingshui Tan, Alibaba Group
 \item Zili Wang, Independent Researcher
 \item Jian Yang, Alibaba Group
 \item Wei Ye, Peking University
 \item Bo Zheng, Alibaba Group
 \item Wangchunshu Zhou, OPPO, M-A-P

\end{itemize}

\paragraph{Corresponding Authors}
\begin{itemize}
    \item Jiaheng Liu, Nanjing University, Institute of Automation, Chinese Academy of Sciences, M-A-P
    \item Dawei Zhu, Peking University
    \item Wenhao Huang, ByteDance, M-A-P
    \item Sujian Li, Peking University
    \item Zhaoxiang Zhang, Institute of Automation, Chinese Academy of Sciences

\end{itemize}

\bibliography{lclm}

\end{document}